\documentclass[lettersize,journal]{IEEEtran}

\usepackage{amsmath,amsfonts}
\usepackage{algorithmic}
\usepackage{algorithm}
\usepackage{array}
\usepackage[caption=false,font=normalsize,labelfont=sf,textfont=sf]{subfig}
\usepackage{textcomp}
\usepackage{stfloats}
\usepackage{url}
\usepackage{verbatim}
\usepackage{graphicx}
\usepackage{cite}

\usepackage{amssymb} 
\usepackage{pifont} 
\usepackage{tabularx} 
\usepackage{arydshln} 
\usepackage{xcolor} 
\usepackage{mathtools} 
\usepackage{adjustbox} 
\usepackage{multirow} 
\usepackage{enumitem}
\usepackage{CJKutf8} 
\usepackage{booktabs} 

\usepackage{hyperref}
\definecolor{citecolor}{HTML}{0071BC}
\definecolor{linkcolor}{HTML}{ED1C24}
\definecolor{darkblue}{rgb}{0, 0, 0.5}
\hypersetup{pagebackref=true, breaklinks=true, letterpaper=true,bookmarksnumbered=true, colorlinks=true, citecolor=darkblue, linkcolor=linkcolor, urlcolor=purple} 
\usepackage{bbding} 
\usepackage{wasysym}
\usepackage{tikz}
\usepackage[edges]{forest}
\usepackage{pgfplots}
\usepackage{nicefrac}
\usepackage{bm}
\usepackage{courier}

\usepackage{amsthm} 
\theoremstyle{definition}

\newif\ifprint
\printtrue

\newcommand{\cmark}{\ding{51}}
\newcommand{\xmark}{\ding{55}}

\newcommand{\Tref}[1]{Table~\ref{#1}}
\newcommand{\Eref}[1]{Equation~\eqref{#1}}
\newcommand{\Fref}[1]{Figure~\ref{#1}}
\newcommand{\Sref}[1]{Section~\ref{#1}}
\newcommand{\Aref}[1]{Algorithm~\ref{#1}}
\newcommand{\Apref}[1]{Appendix~\ref{#1}}

\newcommand{\fref}[1]{Fig.~\ref{#1}}
\newcommand{\sref}[1]{Sec.~\ref{#1}}

\usepackage{makecell}
\usepackage{tabulary}
\definecolor{demphcolor}{RGB}{144,144,144}
\newcommand{\demph}[1]{\textcolor{demphcolor}{#1}}
\definecolor{mygray}{gray}{0.4}

\newlength\savewidth\newcommand\shline{\noalign{\global\savewidth\arrayrulewidth
  \global\arrayrulewidth 1pt}\hline\noalign{\global\arrayrulewidth\savewidth}}

\newcommand{\tablestyle}[2]{\setlength{\tabcolsep}{#1}\renewcommand{\arraystretch}{#2}\centering} 

\newcommand{\vqa}[1]{VQAv2}

\newcommand{\mtname}[0]{{ManagerTower}}
\newcommand{\btname}[0]{{BridgeTower}}
\newcommand{\metername}[0]{{METER}}
\newcommand{\ovname}[0]{{LLaVA-OV}}

\newcommand{\ovmname}[0]{{LLaVA-OV-Manager}}

\newcommand{\basename}[0]{{Baseline}}
\newcommand{\basegname}[0]{{Baseline+Grid}}
\newcommand{\basemname}[0]{{Baseline+Manager}}
\newcommand{\basegmname}[0]{{Baseline+Grid+Manager}}
\newcommand{\vlmoname}[0]{\textsc{VLMo}}
\newcommand{\twotower}[0]{{Two-Tower}}

\newcommand{\modelbase}[0]{$_{\text{BASE}}$}
\newcommand{\modellarge}[0]{$_{\text{LARGE}}$}
\newcommand{\modelhuge}[0]{$_{\text{HUGE}}$}
\newcommand{\rmean}[0]{R$_\text{MEAN}$}

\newcommand{\mergedattention}[0]{\textit{merged-attention}}
\newcommand{\coattention}[0]{\textit{co-attention}}

\newcommand{\eg}[0]{\textit{e.g.},}
\newcommand{\ie}[0]{\textit{i.e.},}
\newcommand{\etc}[0]{\textit{etc}.}
\newcommand{\vs}[0]{\textit{vs}.}

\definecolor{gain}{HTML}{34a853}  %

\definecolor{lost}{HTML}{ea4335}  %

\definecolor{baselinecolor}{gray}{.9}

\begin{document}

\title{
	Manager: Aggregating Insights from Unimodal Experts \\ in Two-Tower VLMs and MLLMs
}

\author{
    Xiao Xu, Libo Qin, Wanxiang Che and Min-Yen Kan,~\IEEEmembership{Senior Member,~IEEE}

    \thanks{
        This work was supported by the National Natural Science Foundation of China (NSFC) via grant 62236004, 62441603 and 62476073. This work was done while Xiao Xu was visiting the National University of Singapore. \textit{(Corresponding authors: Wanxiang Che, Libo Qin.)}
        
        Xiao Xu and Wanxiang Che are with the Research Center for Social Computing and Information Retrieval, Harbin Institute of Technology, 150001, Harbin, Heilongjiang, China (e-mail: xxu@ir.hit.edu.cn, car@ir.hit.edu.cn).
        
        Libo Qin is with the School of Computer Science and Engineering, Central South University, 410083, Changsha, Hunan, China (e-mail: lbqin@csu.edu.cn).
        
        Min-Yen Kan is with the School of Computing, National University of Singapore, 117417, Singapore (e-mail: knmnyn@nus.edu.sg).

        This work is an extension of our conference paper~\cite{xu-etal-2023-managertower}, and has been accepted by TCSVT in \url{https://doi.org/10.1109/TCSVT.2025.3578266}.

        Copyright~\copyright~2025 IEEE. Personal use of this material is permitted. However, permission to use this material for any other purposes must be obtained from the IEEE by sending an email to pubs-permissions@ieee.org.
    }
}

\markboth{IEEE TRANSACTIONS ON CIRCUITS AND SYSTEMS FOR VIDEO TECHNOLOGY}%
{Xu \MakeLowercase{\textit{et al.}}: Manager: Aggregating Insights from Unimodal Experts in Two-Tower VLMs and MLLMs}

\maketitle

\begin{abstract}
    Two-Tower Vision--Language Models (VLMs) have demonstrated strong performance across various downstream VL tasks.
	While BridgeTower further enhances performance by building bridges between encoders, it \textit{(i)} suffers from ineffective layer-by-layer utilization of unimodal representations, \textit{(ii)} restricts the flexible exploitation of different levels of unimodal semantic knowledge, and \textit{(iii)} is limited to the evaluation on traditional low-resolution datasets only with the Two-Tower VLM architecture.
	In this work, we propose Manager, a lightweight, efficient and effective plugin that adaptively aggregates insights from different levels of pre-trained unimodal experts to facilitate more comprehensive VL alignment and fusion.
	First, under the Two-Tower VLM architecture, we introduce ManagerTower, a novel VLM that introduces the manager in each cross-modal layer.
	Whether with or without VL pre-training, ManagerTower outperforms previous strong baselines and achieves superior performance on 4 downstream VL tasks.
	Moreover, we extend our exploration to the latest Multimodal Large Language Model (MLLM) architecture.
	We demonstrate that LLaVA-OV-Manager significantly boosts the zero-shot performance of LLaVA-OV across different categories of capabilities, images, and resolutions on 20 downstream datasets, whether the multi-grid algorithm is enabled or not.
	In-depth analysis reveals that both our manager and the multi-grid algorithm can be viewed as a plugin that improves the visual representation by capturing more diverse visual details from two orthogonal perspectives (depth and width).
	Their synergy can mitigate the semantic ambiguity caused by the multi-grid algorithm and further improve performance.
	Code and models are available at \url{https://github.com/LooperXX/ManagerTower}.\looseness=-1
\end{abstract}

\begin{IEEEkeywords}
    Vision--Language Model, Multimodal Large Language Model, Representation Learning.
\end{IEEEkeywords}
\section{Introduction}
\label{sec:introduction}
\IEEEPARstart{R}{ecently}, the field of Vision--Language (VL) representation learning has gained significant attention, driven by advancements in Vision--Language Pre-training (VLP) techniques.
VLP aims to learn transferable multimodal knowledge from extensive image--text pairs into Vision--Language Models (VLMs), which can improve VL representation and thus further improve performance on various downstream tasks, such as visual question answering~\cite{balanced_vqa_v2}, visual entailment~\cite{xie2019visual}, visual reasoning~\cite{suhr2019corpus}, and image--text retrieval~\cite{young-etal-2014-image}.

\begin{figure}[t]
	\centering
	\includegraphics[width=0.35\textwidth]{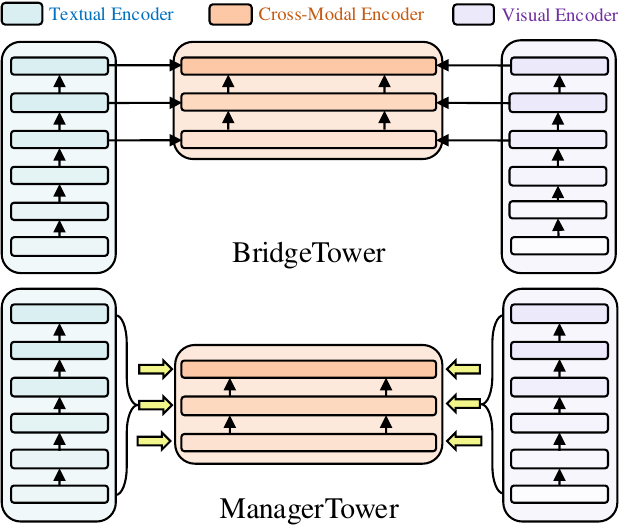}
	\caption{
		A brief overview of \btname{} and \mtname{}. 
		Hollow arrows represent the transmission of multi-layer unimodal representations in \mtname{}, in contrast to the layer-by-layer transmission in \btname{}.
	}
	\label{fig:overview}
\end{figure}

\twotower{} VLM is a general architecture that processes visual and textual modalities with corresponding unimodal encoders and then fuses them in a cross-modal encoder.
\metername{}~\cite{dou2021meter} and \btname{}~\cite{xu2023bridgetower} are two representative \twotower{} VLMs.
\metername{} uses CLIP-ViT~\cite{radford2021learning} and RoBERTa~\cite{liu2019roberta} as pre-trained unimodal encoders, but \textbf{overlooks} different levels of unimodal semantic knowledge contained in multi-layer unimodal representations.
It only feeds the last-layer representation from each unimodal encoder into the cross-modal encoder, which may limit the model capability.
To tackle this issue, as shown in~\fref{fig:overview}, \btname{} builds connections between multiple top unimodal layers and each cross-modal layer in a layer-by-layer manner, to exploit unimodal semantic knowledge at different levels.

In this work, we build upon the research of \btname{} and advance it in four aspects:
($a$) \textbf{Ineffective} layer-by-layer utilization of multi-layer unimodal representations.
Each cross-modal layer is limited to using a pre-defined unimodal layer representation,  which restricts the utilization of different levels of unimodal semantic knowledge and the model capability.
($b$) \textbf{Strictly bound} the number of cross-modal layers to the number of unimodal layer representations the model can use.
An increase in either side leads to a corresponding increase in the other side, leading to more parameters and computation cost, and poor scalability.
($c$) \textbf{Only} exploring the utilization of multi-layer unimodal representations in the \twotower{} VLM architecture.
Lack of exploration in other VLM architectures, \eg{} Multimodal Large Language Model (MLLM), limits the generality of the conclusions.
($d$) \textbf{Limited} post-fine-tuning evaluation on datasets with low-resolution natural images.
Constrained by the capability of traditional VLMs, the model cannot perform more challenging zero-shot evaluations on broader datasets, such as high-resolution document understanding.\looseness=-1

For the first two aspects, under the \twotower{} VLM architecture, we propose a novel VLM, \mtname{}, that introduces managers in each cross-modal layer to aggregate multi-layer unimodal representations, as shown in~\fref{fig:overview}.
Each manager takes multi-layer unimodal representations as \textbf{insights} from pre-trained unimodal \textbf{experts} at different levels (layers), and then aggregates them to facilitate more comprehensive vision--language alignment and fusion.
Inspired by the linear combination of layers method~\cite{wang-etal-2019-learning-deep}, we explore the feasibility of various designs of managers by evaluating and analyzing the performance on VQAv2 and Flickr30K datasets.
The best manager, Adaptive Aggregation Unimodal Manager (AAUM), can \textbf{adaptively} aggregate multi-layer unimodal representations for different tokens in different samples in each cross-modal layer.
Then, we pre-train \mtname{} with commonly used $4$M VLP data and evaluate it on $4$ downstream datasets.
With the same pre-training, fine-tuning and evaluation settings as previous strong \twotower{} VLMs such as \metername{} and \btname{},
\mtname{} achieves superior performances on all datasets, and outperforms not only many base-size models pre-trained on $4$M data but also some models pre-trained on more data and/or with larger size.
Moreover, in principle, managers are scalable and flexible enough to be used as a \textbf{plugin}, easily integrated into any cross-modal encoder, and works well with any unimodal encoder.

\begin{figure}[t]
	\centering
	\includegraphics[width=0.45\textwidth]{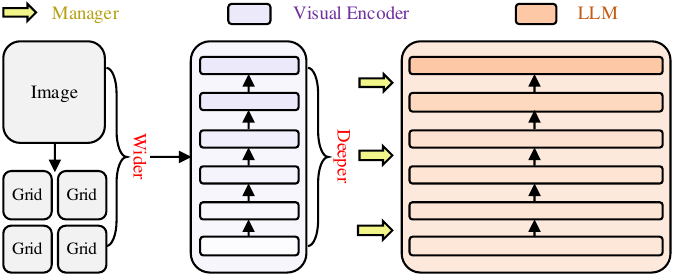}
	\caption{Brief illustrations of \ovmname{}.
        The base image and grids are encoded independently.
		Hollow arrows indicate the transmission of multi-layer visual representations aggregated by managers to the LLM at intervals.\looseness=-1
	}
	\label{fig:overview_mllm}
\end{figure}

For the last two aspects, we further extend the exploration of managers to the latest MLLM architecture, and introduce the manager to \ovname{}~\cite{li2024llava} to get \ovmname{}, as shown in~\fref{fig:overview_mllm}.
Benefiting from the strong LLM and the multi-grid algorithm~\cite{liu2024llavanext} capable of improving the supported image resolution in MLLMs, 
we can zero-shot evaluate the effectiveness of managers on \textbf{broader} downstream datasets, especially on high-resolution images.
We demonstrate that, whether with or without the multi-grid algorithm, managers can \textbf{significantly} improve the performance of MLLMs on $20$ downstream datasets across different categories of capabilities, images, and resolutions.
Further analysis reveals that both the manager and the multi-grid algorithm can be viewed as a \textbf{plugin} that improves the input visual representation.
The manager introduces different levels of semantic knowledge into MLLMs, which can increase the \textbf{diversity} of attention weights and attention heads, thus helping \textbf{guide} the attention of MLLMs that use the multi-grid algorithm.
Their synergy can capture more diverse visual details from two orthogonal perspectives (\textbf{depth} and \textbf{width}), mitigate the semantic ambiguity caused by the multi-grid algorithm and further improve performance.

\section{Preliminary}
\label{sec:preliminary}
We briefly introduce the basic components of Two-Tower VLMs used by \metername{}, \btname{}, and \mtname{}.

\subsection{Visual Encoder}
CLIP-ViT, the visual encoder of CLIP~\cite{radford2021learning}, has been widely used in VLMs~\cite{shen2021much,dou2021meter}.
Each input image is first transformed into a flattened sequence of patches, with a \texttt{[class]} token added at the beginning.
Following a linear projection, position embeddings are added to the sequence to obtain the visual input $\mathbf{V}_0$.
The $\ell^{\text{\,th}}$ visual layer representation is computed as: $\mathbf{V}_{\ell} \!=\! \operatorname{Encoder}^\mathrm{V}_\ell(\mathbf{V}_{\ell-1}), \ell \!=\! 1 \dots L_\mathrm{V}$, where $\ell$ is the layer
index and $L_\mathrm{V}$ is the number of layers in the visual encoder.

\begin{figure*}[t]
	\centering
	\includegraphics[width=0.9\textwidth]{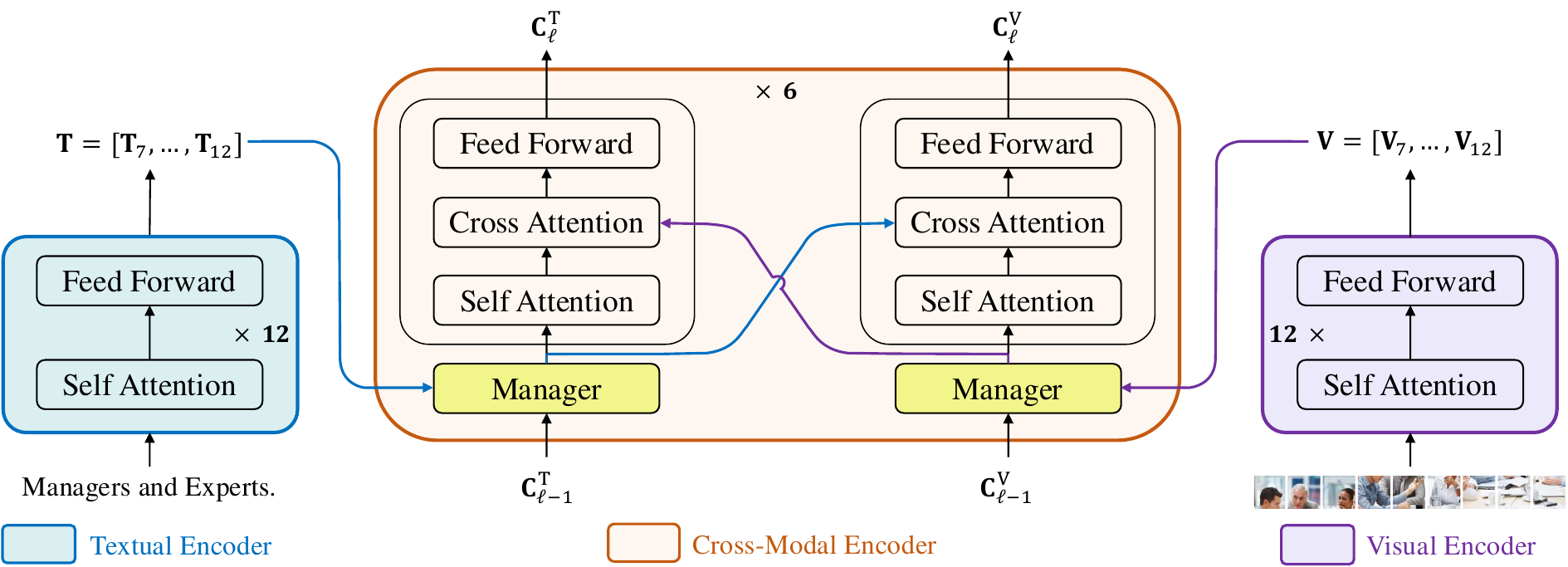}
	\caption{
		An illustration of \mtname{} shows that each cross-modal layer includes a textual manager and a visual manager.
		Top $\mathrm{N}\!=\!6$ unimodal layer representations $\mathbf{T},\mathbf{V}\!\in\!\mathbb{R}^{\mathrm{N} \times \mathrm{L} \times \mathrm{D}}$ along with the representations from the previous cross-modal layer ${\mathbf{C}}^\mathrm{T}_{\ell-1}, {\mathbf{C}}^\mathrm{V}_{\ell-1}, \ell \!=\! 1 \dots 6$ are input into the textual manager $\mathcal{M}_{\ell}^{\mathrm{T}}$ and the visual manager $\mathcal{M}_{\ell}^{\mathrm{V}}$, respectively.
		$\mathrm{N}$ refers to the number of pre-trained unimodal experts the model uses, 
		and $\mathrm{L}$ denotes the length of the input sequence.\looseness=-1
	}
	\label{fig:framework}
\end{figure*}

\subsection{Textual Encoder}
RoBERTa~\cite{liu2019roberta} is widely used in VLMs~\cite{dou2021meter,li2022unimo} due to its robust performance.
The input text is tokenized with the byte-level Byte-Pair Encoding (BPE)~\cite{sennrich-etal-2016-neural,radford2019language}.
$\texttt{[<s>]}$ and $\texttt{[</s>]}$ tokens are added to the start and end of the sequence, respectively.
Word embeddings and positional embeddings are then applied to the tokenized sequence to generate the visual input $\mathbf{T}_0$.
Similarly, the $\ell^{\text{\,th}}$ textual layer representation is computed as: $\mathbf{T}_{\ell} \!=\! \operatorname{Encoder}^\mathrm{T}_\ell(\mathbf{T}_{\ell-1}), \ell \!=\! 1 \dots L_\mathrm{T}$, where $L_\mathrm{T}$ denotes the number of layers in the textual encoder.

\subsection{Cross-Modal Encoder}
\label{sec:cross-modal-encoder}
We use the transformer encoder~\cite{vaswani2017attention} with a co-attention mechanism~\cite{lu2019vilbert} as the cross-modal encoder.
In each cross-modal layer, both modalities are equipped with a multi-head self-attention (MSA) block, a multi-head cross-attention (MCA) block, and a feed-forward (FFN) block.
The MCA block allows the visual part of the cross-modal encoder to attend to the textual part and vice versa.
$\operatorname{Encoder}^\mathrm{C}_\ell, \ell \!=\! 1 \dots L_\mathrm{C}$ denotes the $\ell^{\text{\,th}}$ cross-modal layer, where $L_\mathrm{C}$ is the number of cross-modal layers.
For brevity, it is computed as:
\begin{align}
	\tilde{\mathbf{C}}^\mathrm{V}_{\ell} & = {\mathbf{C}}^\mathrm{V}_{\ell-1}, \label{eq:prev1} \\
	\tilde{\mathbf{C}}^\mathrm{T}_{\ell} & = {\mathbf{C}}^\mathrm{T}_{\ell-1}, \label{eq:prev2} \\
	\mathbf{C}^\mathrm{V}_{\ell}, \mathbf{C}^\mathrm{T}_{\ell} & = \operatorname{Encoder}^\mathrm{C}_\ell(\tilde{\mathbf{C}}^\mathrm{V}_{\ell}, \tilde{\mathbf{C}}^\mathrm{T}_{\ell}), \label{eq:cross_modal_layer}
\end{align}
where $\mathbf{C}^\mathrm{V}_{\ell}, \mathbf{C}^\mathrm{T}_{\ell}$ are the visual and textual part of output representation of the $\ell^{\text{\,th}}$ layer, $\tilde{\mathbf{C}}^\mathrm{V}_{\ell}, \tilde{\mathbf{C}}^\mathrm{T}_{\ell}$ are inputs of each part. $\mathbf{C}^\mathrm{V}_{0}, \mathbf{C}^\mathrm{T}_{0}$ are initialized with the last-layer representations from unimodal encoders: $\mathbf{C}^\mathrm{V}_{0} \!=\! \mathbf{V}_{L_\mathrm{V}}\mathbf{W}_\mathrm{V}, \mathbf{C}^\mathrm{T}_{0} \!=\! \mathbf{T}_{L_\mathrm{T}}\mathbf{W}_\mathrm{T}$, where $\mathbf{W}_\mathrm{V},\mathbf{W}_\mathrm{T}$ are linear cross-modal projections.
In this work, we use the same default setting as \metername{} and \btname{} for a fair comparison: pre-trained unimodal encoders with $L_\mathrm{V}\!=\!L_\mathrm{T}\!=\!12$, randomly-initialized cross-modal encoder with $L_\mathrm{C}\!=\!6$, and only top $\mathrm{N}\!=\!6$ unimodal layer representations are used.\looseness=-1

\section{Manager Design}
\label{sec:manager-design}
\fref{fig:framework} illustrates the overall framework of \mtname{}.
It introduces managers in each cross-modal layer to aggregate insights from different levels of pre-trained unimodal experts.
Under the \twotower{} VLM architecture, we will elaborate on the detailed design schema for the three types of managers, and conclude with the cross-modal encoder with our managers.\footnote{
	Details about pre-training and fine-tuning are described 
	in~\Apref{appendix:implementation_details}.
}

\subsection{Static Aggregation Manager (SAM)}
\label{sec:sae}
The effectiveness of layer fusion in learning comprehensive representations has been well demonstrated~\cite{wang-etal-2018-multi-layer,wang-etal-2019-learning-deep,wei-etal-2020-multiscale}.
Inspired by this, we aim to apply this technique to VLMs. As a preliminary exploration, we adopt the linear combination of layers method~\cite{wang-etal-2019-learning-deep}, which is a simple yet effective way that aggregates the representations of previous layers using learned weights in each encoder layer.
We directly adapt it to aggregate both unimodal and cross-modal representations of all previous layers and call it the Static Aggregation Manager (SAM).
The calculation for the $\ell^{\text{\,th}}$ visual manager is given by:
\begin{equation}
	\begin{split}
		& \mathcal{M}_{\ell}^{\mathrm{V}}(\mathbf{V}_{7}, \dots, \mathbf{V}_{12}, \mathbf{C}^\mathrm{V}_{1}, \dots, \mathbf{C}^\mathrm{V}_{\ell-1}) = \\
		& \sum_{i=1}^{6}{\mathbf{W}_{i}^{\mathrm{V}, \ell} \odot \operatorname{LN}(\mathbf{V}_{i+6})} \!+\! \sum_{i=1}^{\ell-1}{\mathbf{W}_{i+6}^{\mathrm{V}, \ell} \odot \operatorname{LN}(\mathbf{C}^\mathrm{V}_{i})}, \label{eq:sae}
	\end{split}
\end{equation}
where $\mathcal{M}_{\ell}^{\mathrm{V}}$ represents the manager for the visual part of the $\ell^{\text{\,th}}$ cross-modal layer, and $\mathbf{W}^{\mathrm{V}, \ell} \!\in\! \mathbb{R}^{(6+\ell-1) \times \mathrm{D}}$ is a learnable parameter matrix.
$\odot$ denotes the element-wise product operation, and $\operatorname{LN}(\cdot)$ refers to Layer Normalization~\cite{ba2016layer}.
We then omit the superscript $^{\mathrm{V}, \ell}$ of $\mathbf{W}$ for brevity.
$\mathbf{W}$ can be seen as the learned aggregation weight and normalized by the $\operatorname{softmax}$ function with a learnable temperature.
We initialize $\mathbf{W}$ with $\frac{1}{6+\ell-1}$ on average to assign equal weights to the representations from all previous layers.

However, directly applying SAM to VLMs \textbf{does not} result in an expected performance improvement over \btname{}, and instead leads to a notable decrease in performance.
We hypothesize that this performance drop is due to the average initialization of $\mathbf{W}$.
It may not be suitable for both cross-modal and pre-trained unimodal layer representations as they have \textbf{different} numerical scales.
To test this hypothesis, we propose dividing the parameter matrix $\mathbf{W}$ into unimodal and cross-modal parts, and initializing them with $\frac{1}{6}$ and $\frac{1}{\ell-1}$, respectively,\footnote{
	We also experimented with other initialization methods: one, progressive, exponential moving average, \btname{}-like one-hot, \etc{}, but the results were similar to or worse than this initialization.
} and also learn the $\operatorname{softmax}$ temperature separately.
The experimental result shows a \textbf{significant} improvement over the direct application of SAM, 
though the improvement is still somewhat \textbf{limited} compared to \btname{}.
These observations provide a compelling argument for re-examining how to aggregate multi-layer pre-trained unimodal representations.\looseness=-1

\begin{figure}[t]
	\centering
	\includegraphics[width=0.45\textwidth]{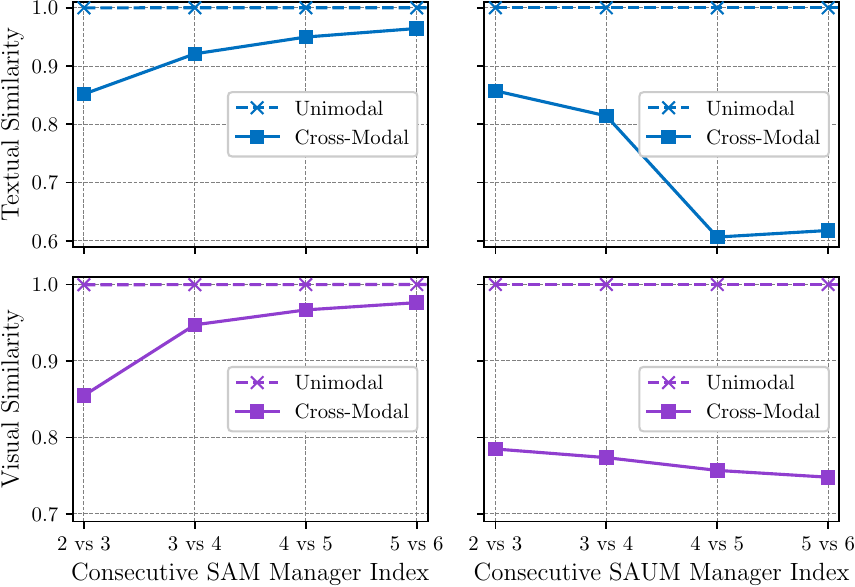}
	\caption{
		Cosine similarity between the aggregated unimodal/cross-modal representations of each pair of consecutive textual/visual managers.
		The aggregated representations derived from~\Eref{eq:sae} can be divided into unimodal and cross-modal parts.
		For each part, we analyse the representations similarity between every two consecutive managers, modality-wise.
	}
	\label{fig:cosine_similarity_cross_uni}
\end{figure}

\subsection{Static Aggregation Unimodal Manager (SAUM)}
\label{sec:saue}
Since the aggregated representations derived from~\Eref{eq:sae} consist of an unimodal part and a cross-modal part, 
we calculate the cosine similarity between aggregated unimodal/cross-modal representations of each pair of consecutive textual/visual managers.
This can help further analyse insights aggregated in different SAMs, \ie{} inputs to different cross-modal layers.
As shown in~\fref{fig:cosine_similarity_cross_uni}, for SAMs, the unimodal similarity \textbf{remains} close to $1$, while the cross-modal similarity \textbf{increases} with depth and tends toward $1$. 
This suggests that, the unimodal representations aggregated by different SAMs are \textbf{nearly identical}, and the aggregated cross-modal representations get more similar with depth.

We hypothesize that, since different SAMs provide \textbf{similar} aggregated unimodal representations for each cross-modal layer, the representations from more preceding cross-modal layers may bring \textbf{redundant} information to \textbf{confuse} the managers. 
This leads to aggregated cross-modal representations converging to indistinguishable vectors as the depth increases.

To address this, we propose focusing on aggregating insights from pre-trained unimodal experts and retaining \textbf{only} the representation from the previous cross-modal layer.
We refer to it as the Static Aggregation Unimodal Manager (SAUM).
The calculation of the $\ell^{\text{\,th}}$ visual manager computes as:
\begin{equation}
	\begin{split}
		& \mathcal{M}_{\ell}^{\mathrm{V}}(\mathbf{V}_{7}, \dots, \mathbf{V}_{12}, \mathbf{C}^\mathrm{V}_{\ell-1}) = \\
		& \sum_{i=1}^{6}{\mathbf{W}_{i} \odot \operatorname{LN}\left(\mathbf{V}_{i+6}\right)} \!+\! {\mathbf{W}_{\mathrm{C}} \odot \operatorname{LN}(\mathbf{C}^\mathrm{V}_{\ell-1})}, \label{eq:saue}
	\end{split}
\end{equation}
where $\mathbf{W}\!\in\!\mathbb{R}^{6 \times \mathrm{D}}$ and $\mathbf{W}_{\mathrm{C}}\!\in\!\mathbb{R}^{1 \times \mathrm{D}}$ are learnable parameter matrices, initialized with $\frac{1}{6}$ and $1$ on average, respectively.
The $\operatorname{softmax}$ with a learnable temperature only normalizes $\mathbf{W}$.\looseness=-1

The substantial improvement observed compared to \btname{} provides empirical support for our hypothesis.
Furthermore, as shown in~\fref{fig:cosine_similarity_cross_uni}, the cross-modal similarity of SAUM \textbf{decreases} with the depth, suggesting that more comprehensive and distinguishable cross-modal representations are aggregated as the depth increases.

\subsection{Adaptive Aggregation Unimodal Manager (AAUM)}
Despite the significant performance gains achieved by SAUM, it still faces two key limitations:
($i$) $\mathbf{W}$, the learned aggregation weight for unimodal representations is nearly \textbf{identical} across managers in different cross-modal layers, as demonstrated in~\fref{fig:cosine_similarity_cross_uni},
which contradicts the intuition that the requirement for unimodal semantic knowledge should \textbf{vary} among cross-modal layers; 
($ii$) during inference, for each manager, the \textbf{same} aggregation weight $\mathbf{W}$ learned during training is applied to all tokens in different samples, which does not align with the intuition that the need for unimodal semantic knowledge should \textbf{vary} among tokens and samples.

\begin{figure*}[t]
	\centering
	\includegraphics[width=0.9\textwidth]{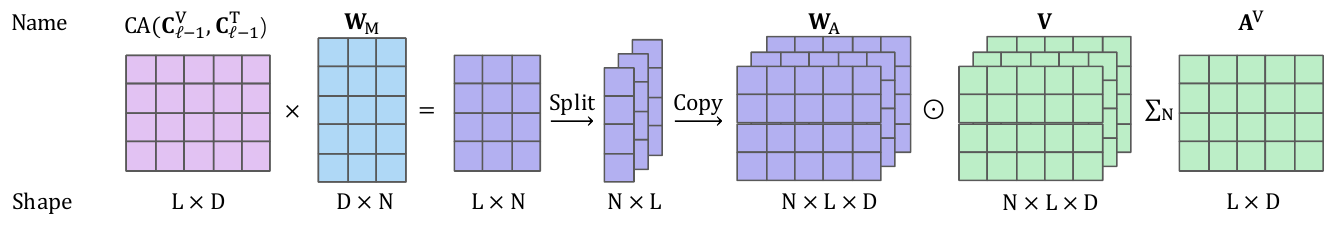}
	\caption{
	An illustration of how the aggregated unimodal representations $\mathbf{A}^\mathrm{V}\!\in\!\mathbb{R}^{\mathrm{L} \times \mathrm{D}}$ are calculated in the visual AAUM.
	CA refers to the cross-attention mechanism.
	$\mathrm{N}\!=\!6$.
	For simplicity, we omit $\operatorname{LN}$ and $\operatorname{softmax}$ function.
	}
	\label{fig:details}
\end{figure*}

To address the above limitations, we propose the Adaptive Aggregation Unimodal Manager (AAUM).
During training and inference, AAUM can \textbf{adaptively} utilize different levels of unimodal semantic knowledge from pre-trained unimodal experts for different tokens across different samples.
Take the visual AAUM for example, the $\ell^{\text{\,th}}$ visual manager computes as:\looseness=-1
\begin{equation}
	\begin{split}
		& \mathcal{M}_{\ell}^{\mathrm{V}}(\mathbf{V}_{7}, \dots, \mathbf{V}_{12}, \mathbf{C}^\mathrm{V}_{\ell-1}) = \\
		& \sum_{i=1}^{6}{\mathbf{W}_{\mathrm{A}, i} \odot \operatorname{LN}\left(\mathbf{V}_{i+6}\right)} \!+\! {\mathbf{W}_{\mathrm{C}}\!\odot\!\operatorname{LN}(\mathbf{C}^\mathrm{V}_{\ell-1})}, \label{eq:aaue}
	\end{split}
\end{equation}
\begin{equation}
	\mathbf{W}_\mathrm{A} = \operatorname{softmax}(\operatorname{LN}(\mathbf{C}^\mathrm{V}_{\ell-\!1}) \times \mathbf{W}_{\mathrm{M}} + \epsilon), \label{eq:moe}
\end{equation}
where $\mathbf{W}_{\mathrm{M}}\!\in\!\mathbb{R}^{\mathrm{D} \times 6}$ denotes a linear projection layer.
The generated aggregation weights $\mathbf{W}_\mathrm{A}\!\in\!\mathbb{R}^{6 \times \mathrm{L} \times \mathrm{D}}$ can adaptively aggregate unimodal representations from different levels of pre-trained unimodal experts for each token.
The $\operatorname{softmax}$ function features a learnable temperature and $\epsilon \!\sim\! \mathcal N(0, \frac{1}{6^2})$ denotes a Gaussian noise for exploration of aggregation~\cite{fedus2022switch}.

Furthermore, to help managers better exploit unimodal semantic knowledge, 
we propose replacing the visual query $\mathbf{C}^\mathrm{V}_{\ell-1}$ in~\Eref{eq:moe} with the cross-modal fused query $\operatorname{CA}(\mathbf{C}^\mathrm{V}_{\ell-1}, \mathbf{C}^\mathrm{T}_{\ell-1})$ to further improve performance, 
where $\operatorname{CA}$ is a cross-attention mechanism.

\subsection{Cross-Modal Encoder with Managers}

Since the $1^{\text{st}}$ cross-modal layer lacks the representation of the previous cross-modal layer as the query,
we introduce SAUM in the $1^{\text{st}}$ cross-modal layer and AAUMs in the subsequent layers.
Therefore,~\Eref{eq:prev1}~\&~(\ref{eq:prev2}) for the $1^{\text{st}}$ cross-modal layer with SAUMs is computed as:
\begin{gather}
	\tilde{\mathbf{C}}^\mathrm{V}_{1} = \mathcal{M}_{1}^{\mathrm{V}}(\mathbf{V}_{7}, \dots, \mathbf{V}_{12}), \\
	\tilde{\mathbf{C}}^\mathrm{T}_{1} = \mathcal{M}_{1}^{\mathrm{T}}(\mathbf{T}_{7}, \dots, \mathbf{T}_{12}),
\end{gather}
For the $2^{\text{nd}}$ and subsequent cross-modal layers with AAUMs:
\begin{gather}
	\tilde{\mathbf{C}}^\mathrm{V}_{\ell} = \mathcal{M}_{\ell}^{\mathrm{V}}(\mathbf{V}_{7}, \dots, \mathbf{V}_{12}, \mathbf{C}^\mathrm{V}_{\ell-1}, \mathbf{C}^\mathrm{T}_{\ell-1}), \\
	\tilde{\mathbf{C}}^\mathrm{T}_{\ell} = \mathcal{M}_{\ell}^{\mathrm{T}}(\mathbf{T}_{7}, \dots, \mathbf{T}_{12}, \mathbf{C}^\mathrm{T}_{\ell-1}, \mathbf{C}^\mathrm{V}_{\ell-1}),
\end{gather}
where we omit the modality type and layer index embeddings added to unimodal layer representations $\mathbf{V}, \mathbf{T}$ in the above equations for simplicity.

\fref{fig:details} shows the adaptive aggregation of insights from pre-trained visual experts in AAUMs, which corresponds to the unimodal (right) part of Equation~(\ref{eq:aaue}).
As for SAUMs, the learned weights $\mathbf{W}\!\in\!\mathbb{R}^{\mathrm{6} \times \mathrm{D}}$ are directly broadcast to $\mathbf{W}_\mathrm{A}$, and then they aggregate insights similarly to AAUMs.

\section{Exploration on \twotower{} VLM}
\label{sec:experiments}

\subsection{Implementation Details}
\mtname{} comprises a pre-trained textual encoder, RoBERTa\modelbase{} with $124$M parameters, a pre-trained visual encoder, CLIP-ViT B-224/16 with $86$M parameters, and a randomly initialized $6$-layer cross-modal encoder with managers, totaling $113$M$+12$M parameters.
The detailed setting of the cross-modal encoder is the same as \btname{}.
The maximum length of the text sequence is set to $50$, and the image patch size is $16 \times 16$.
For a fair comparison with \btname{}, we use an image resolution of $384 \times 384$ for Flickr30K and $576 \times 576$ for VQAv2.
AdamW~\cite{loshchilov2018decoupled} optimizer with a base learning rate of $2e^{-5}$ and warmup ratio of $0.1$ is used.\looseness=-1

\subsection{Investigation and Analysis}
\label{sec:investigation-and-analysis}
In this section, we investigate various designs of managers and evaluate the performance by directly fine-tuning on VQAv2 and Flickr30K without VLP.
Experimental settings are the same as \btname{} for a fair comparison.
Note that unimodal encoders are initialized with their pre-trained weights.\looseness=-1

\subsubsection{Type of Manager}

\begin{table}[t]
	\tablestyle{3pt}{1.1}
	\caption{
		Performance of different types of managers and queries on VQAv2 and Flickr30K.
		\rmean{} indicates the mean recall metrics for image--text retrieval.
		BT denotes \btname{}.
	}
	\label{tab:type-and-query}
	\adjustbox{width=0.95\linewidth}{
		\begin{tabular}{l|c|c|c|c}
			Type                  & Visual Query                                                                & Weight                                           & Test-Dev (\%)       & \rmean{} (\%)   \\
			\shline
			\demph{BT}  & \demph{-}                                                                           & \demph{$\mathrm{N} \times 1$}                            & \demph{75.91}          & \demph{93.33}      \\
			\hline
			\multirow{2}{*}{SAM}  & -                                                                           & $\mathrm{N} \times 1$                            & 76.19          & 93.57      \\
			                      & -                                                                           & $\mathrm{N} \times \mathrm{D}$                   & 76.18          & 93.73      \\
			\hline
			\multirow{2}{*}{SAUM} & -                                                                           & $\mathrm{N} \times 1$                            & 76.38          & 93.75      \\
			\textbf{}             & -                                                                           & $\mathrm{N} \times \mathrm{D}$                   & 76.55          & 93.82      \\
			\hline
			\multirow{2}{*}{AAUM} & $\mathbf{C}^\mathrm{V}_{\ell-1}$                                            & $\mathrm{N} \times \mathrm{L}$                   & 76.52          & 93.84      \\
			                      & $\mathbf{C}^\mathrm{V}_{\ell-1}, \mathbf{C}^\mathrm{T}_{\ell-1}$            & $\mathrm{N} \times \mathrm{L}$                   & \textbf{76.65} & \bf{93.97} \\
			\hline
			Concat-               & $\mathbf{V}, \mathbf{C}^\mathrm{V}_{\ell-1}$                                & $\mathrm{N} \times \mathrm{L} \times \mathrm{D}$ & 76.38          & 93.78      \\
			Attention             & $\mathbf{V},\mathbf{C}^\mathrm{V}_{\ell-1}, \mathbf{C}^\mathrm{T}_{\ell-1}$ & $\mathrm{N} \times \mathrm{L} \times \mathrm{D}$ & 76.43          & 93.83      \\
			\hline
			Cross-                & $\mathbf{C}^\mathrm{V}_{\ell-1}$                                            & $\mathrm{N} \times \mathrm{L}$                   & 76.41          & 92.15      \\
			Attention             & $\mathbf{C}^\mathrm{V}_{\ell-1}, \mathbf{C}^\mathrm{T}_{\ell-1}$            & $\mathrm{N} \times \mathrm{L}$                   & 76.45          & 92.61
		\end{tabular}
	}
\end{table}

We first explore the performance of different types of managers and queries.
Take the visual manager for example, based on the top $\mathrm{N}\!=\!6$ visual layer representations $\mathbf{V}\!\in\!\mathbb{R}^{\mathrm{N} \times \mathrm{L} \times \mathrm{D}}$ from CLIP-ViT,
different managers provide the aggregation weights that can be broadcast to $\mathbf{W}_\mathrm{A}$ for aggregating insights from pre-trained visual experts.

From the perspective of aggregation weights $\mathbf{W}_\mathrm{A}$, SAM and SAUM are \textbf{static} sentence-level managers that share the same aggregation weights for all tokens across different samples.
In contrast, AAUM is an \textbf{adaptive} token-level manager that adaptively \textbf{generates} different aggregation weights for different tokens across different samples.
Besides, we also implement \Eref{eq:moe} with common cross- and concat-attention mechanisms for comparison, detailed in~\Aref{appendix:other-managers}.\looseness=-1

The results are summarized in~\Tref{tab:type-and-query}.
By focusing on aggregating insights from pre-trained unimodal experts, SAUM demonstrates \textbf{superior} performance over SAM on both datasets.
Furthermore, with the help of the cross-modal fused query, AAUM \textbf{significantly} outperforms the other managers.
This highlights that \textbf{adaptive} token-level aggregation with a cross-modal fused query outperforms \textbf{static}, sentence-level aggregation.
Notably, the cross-modal fused query incorporates both visual and textual parts of the previous cross-modal layer representation, which can better help managers correctly aggregate unimodal semantic knowledge required by the current cross-modal layer.\footnote{
	Further elaboration of the relationship between different types of managers can be found in~\Apref{appendix:managers-equations}\&\ref{appendix:managers}.
}

\subsubsection{Number of Cross-Modal Layers}
We conduct a comparison between \mtname{} and \btname{} with different numbers of cross-modal layers in~\Tref{tab:number-of-cross-modal-layers}, to further assess the effectiveness of \mtname{}.
Regardless of the number of cross-modal layers, \mtname{} \textbf{consistently} and \textbf{significantly} outperforms \btname{} on both datasets.
More interestingly, the performance of \mtname{} with $L_\mathrm{C}\!=\!3$ is even better than that of \btname{} with $L_\mathrm{C}\!=\!6$ ($76.04\%\!>\!75.91\%,\  93.41\%\!>\!93.33\%$).

In contrast to \btname{}, $\mathrm{N}$, the number of top unimodal layer representations used by \mtname{}, is not bound to the number of cross-modal layers $L_\mathrm{C}$ and can be flexibly adjusted.
The default setting is $\mathrm{N}\!=\!6$.
Therefore, \mtname{} actually utilizes the same number of unimodal layer representations as \btname{}, but achieves \textbf{superior} performance with \textbf{only half} the number of cross-modal layers.
This further highlights the \textbf{flexibility} and \textbf{effectiveness} of \mtname{} in adaptive aggregation of  unimodal semantic knowledge, in contrast to layer-by-layer exploitation in \btname{}.

\begin{table}[t]
	\tablestyle{5pt}{1.1}
	\caption{
		Performance of \btname{} (BT) and \mtname{} (Ours) with different numbers of cross-modal layers.
	}
	\label{tab:number-of-cross-modal-layers}
	\adjustbox{width=0.8\linewidth}{
		\begin{tabular}{c|cc|cc}
			\multirow{2}{*}{$L_\mathrm{C}$} & \multicolumn{2}{c|}{VQAv2 Test-Dev (\%)} & \multicolumn{2}{c}{Flickr30K \rmean{}(\%)}                                             \\
			                                & BT                                  & \multicolumn{1}{c|}{Ours}              & BT    & \multicolumn{1}{c}{Ours}          \\
			\shline
			2                               & 74.86                               & 75.47 ($\uparrow$\,0.61)               & 92.45 & 93.31 ($\uparrow$\,0.86)          \\
			3                               & 75.33                               & 76.04 ($\uparrow$\,0.71)               & 92.50 & 93.41 ($\uparrow$\,0.91)          \\
			4                               & 75.74                               & 76.26 ($\uparrow$\,0.52)               & 92.76 & 93.59 ($\uparrow$\,0.83)          \\
			6                               & 75.91                               & \textbf{76.65} ($\uparrow$\,0.74)      & 93.33 & \textbf{93.97} ($\uparrow$\,0.64) \\
			8                               & 75.89                               & 76.47 ($\uparrow$\,0.58)               & 93.03 & 93.65 ($\uparrow$\,0.62)          \\
		\end{tabular}
	}
\end{table}

\begin{figure}[t]
	\centering
	\includegraphics[width=0.43\textwidth]{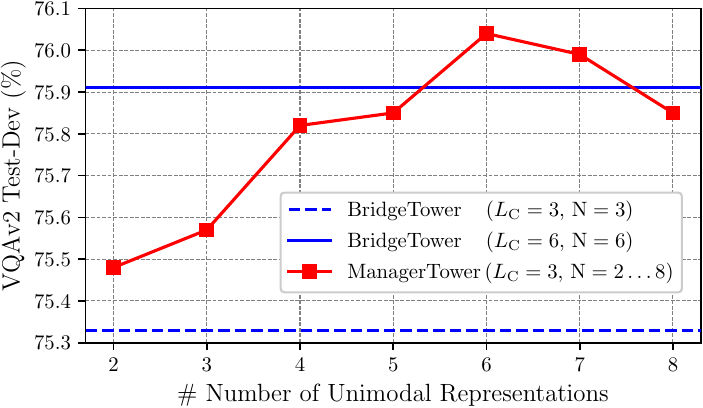}
	\caption{
		VQAv2 Test-Dev Performance using different numbers of unimodal representations in \mtname{} ($L_\mathrm{C}\!=\!3, \mathrm{N}=2 \dots 8$), 
		where $L_\mathrm{C}$ is the number of cross-modal layers, and $\mathrm{N}$ is the number of top unimodal layer representations used in each bridge or manager.
	}
	\label{fig:number-of-unimodal-representations}
\end{figure}

\begin{table*}[ht]
	\tablestyle{5pt}{1.1} 
	\caption{
		Comparisons with previous models on $4$ downstream datasets after VLP.
		The best score is bolded. 
		$\ast$ indicates that the model also uses VG-QA data to fine-tune on VQAv2.
	}
	\label{tab:main-results}
	\adjustbox{width=0.85\linewidth}{
		\begin{tabular}{lr|cc|cc|cc|cc}
			\multirow{2}{*}{Model}                             & {\#~Pre-train} & \multicolumn{2}{c|}{VQAv2 (\%)} & \multicolumn{2}{c|}{SNLI-VE (\%)} & \multicolumn{2}{c|}{NLVR$^2$ (\%)} & \multicolumn{2}{c}{Flickr30K (\%)}                                                     \\
			                                                   & Images~        & Test-Dev                   & Test-Std                     & Dev                           & Test                          & Dev        & Test-P     & IR@1       & TR@1       \\
			\shline
			\multicolumn{10}{l}{ { \it{Base-size models pre-trained on 4M public data} } }                                                                                                                                                                      \\
			\hline
			ViLT\modelbase{}~\cite{kim2021vilt}               & 4M             & 71.26                      & -                            & -                             & -                             & 75.70      & 76.13      & 64.4       & 83.5       \\
			UNITER\modelbase{}~\cite{chen2020uniter}\,$\ast$  & 4M             & 72.70                      & 72.91                        & 78.59                         & 78.28                         & 77.18      & 77.85      & 72.52      & 85.90      \\
			UNIMO\modelbase{}~\cite{li2021unimo}              & 4M             & 73.79                      & 74.02                        & 80.00                         & 79.10                         & -          & -          & 74.66      & 89.70      \\
			ALBEF\modelbase{}~\cite{li2021align}\,$\ast$      & 4M             & 74.54                      & 74.70                        & 80.14                         & 80.30                         & 80.24      & 80.50      & 82.8       & 94.3       \\
			\metername{}-Swin\modelbase{}\cite{dou2021meter}  & 4M             & 76.43                      & 76.42                        & 80.61                         & 80.45                         & 82.23      & 82.47      & 79.02      & 92.40      \\
			\vlmoname{}\modelbase{}~\cite{wang2021vlmo}       & 4M             & 76.64                      & 76.89                        & -                             & -                             & 82.77      & 83.34      & 79.3       & 92.3       \\
			\metername{}-CLIP\modelbase{}~\cite{dou2021meter} & 4M             & 77.68                      & 77.64                        & 80.86                         & 81.19                         & 82.33      & 83.05      & 82.22      & 94.30      \\
			\btname{}\modelbase{}~\cite{xu2023bridgetower}         & 4M             & 78.66                      & 78.73                        & 81.11                         & 81.19                         & 81.85      & 83.09      & 85.83      & 94.73      \\
			\mtname{}\modelbase{}~(\bf{Ours})                  & 4M             & \bf{79.39}                 & \bf{79.15}                   & \bf{81.26}                    & \bf{81.44}                    & \bf{82.81} & \bf{83.34} & \bf{86.56} & \bf{95.64} \\
			\hline
			\multicolumn{10}{l}{ { \it{Models pre-trained on more data and/or with larger size}}}                                                                                                                                                               \\
			\hline
			UNITER\modellarge{}~\cite{chen2020uniter}\,$\ast$ & 4M             & 73.82                      & 74.02                        & 79.39                         & 79.38                         & 79.12      & 79.98      & 75.56      & 87.30      \\
			UNIMO\modellarge{}~\cite{li2021unimo}             & 4M             & 75.06                      & 75.27                        & 81.11                         & 80.63                         & -          & -          & 78.04      & 89.40      \\
			ALBEF\modelbase{}~\cite{li2021align}\,$\ast$      & 14M            & 75.84                      & 76.04                        & 80.80                         & 80.91                         & 82.55      & 83.14      & 85.6       & 95.9       \\
			SimVLM\modelbase{}~\cite{wang2021simvlm}          & 1.8B           & 77.87                      & 78.14                        & 84.20                         & 84.15                         & 81.72      & 81.77      & -          & -          \\
			BLIP\modelbase{}~\cite{li2022blip}\,$\ast$        & 129M           & 78.24                      & 78.17                        & -                             & -                             & 82.48      & 83.08      & 87.3       & 97.3       \\
			SimVLM\modellarge{}~\cite{wang2021simvlm}         & 1.8B           & 79.32                      & 79.56                        & 85.68                         & 85.62                         & 84.13      & 84.84      & -          & -          \\
		\end{tabular}
	}
\end{table*}

\subsubsection{Number of Unimodal Experts}
\label{sec:number-of-unimodal-experts}
We further explore the impact of varying $\mathrm{N}$ in \mtname{} with $L_\mathrm{C}\!=\!3$.
As shown in~\fref{fig:number-of-unimodal-representations}, there exist two interesting observations:
($i$) \mtname{} ($L_\mathrm{C}\!=\!3, \mathrm{N}\!=\!3$) outperforms \btname{} ($L_\mathrm{C}\!=\!3, \mathrm{N}\!=\!3$), suggesting that when the same number of unimodal layer representations are introduced, \mtname{} allows more \textbf{effective} aggregation of unimodal semantic knowledge, thus facilitating vision--language alignment and fusion in each cross-modal layer; 
($ii$) the performance of \mtname{} initially improves gradually, but decreases after $\mathrm{N}\!>\!6$.
We assume that lower-layer unimodal representations \textbf{may not} help \mtname{} learn vision--language alignment and fusion, and may also increase the computational cost. This is also consistent with \btname{}'s observations.

\subsection{Comparison with Previous Arts}
\subsubsection{Pre-train Settings}
We pre-train \mtname{} with two standard VLP objectives, masked language modeling (MLM) and image--text matching (ITM), on the widely-used $4$M public data: Conceptual Captions~\cite{sharma-etal-2018-conceptual}, SBU Captions~\cite{NIPS2011_5dd9db5e}, MSCOCO Captions~\cite{chen2015microsoft}, and Visual Genome (VG)~\cite{krishna2017visual}.
The pre-train settings are the same as \btname{} and \metername{} for a fair comparison.
\mtname{} is pre-trained for $100$k steps with a batch size of $4096$ and a learning rate of $1e^{-5}$. The image resolution for VLP is $288 \times 288$ and only center-crop~\cite{radford2021learning} is used without any data augmentation.

\begin{figure*}[t]
	\centering
	\includegraphics[width=\textwidth]{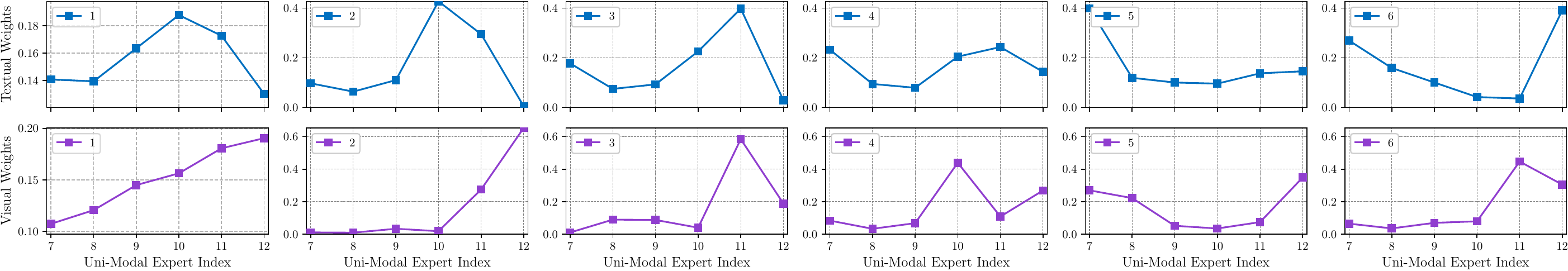}
	\caption{
		A visualization of aggregation weights of textual and visual AAUMs in each cross-modal layer after VLP.
		The X-axis shows the index of the unimodal expert, and the legend shows the index of the cross-modal layer. 
	}
	\label{fig:aggregation-weights-pre-trained-aaum}
\end{figure*}

\subsubsection{Main Results}
\Tref{tab:main-results} shows the performance of \mtname{} compared with other previous works on $4$ downstream datasets.
With only $4$M VLP data, \mtname{} achieves \textbf{superior} performances on these datasets.
Based on the same pre-training and fine-tuning settings and unimodal backbones as previous strong \twotower{} VLMs, \ie{} \metername{} and \btname{},
\mtname{} achieves \textbf{significant} improvements on all datasets, especially $79.15\%$ accuracy on VQAv2 Test-Std, $86.56\%$ IR@1 and $95.64\%$ TR@1 on Flickr30K.
This further demonstrates that with all other factors fixed, compared to \btname{} that introduces bridges to \metername{}, managers in \mtname{} allow \textbf{effective} aggregation of multi-layer unimodal representations via well-designed managers.
Managers can \textbf{adaptively} aggregate more required unimodal semantic knowledge to facilitate comprehensive vision--language alignment and fusion in each cross-modal layer.
Notably, \mtname{} not only outperforms many base-size models pre-trained on $4$M data, but also surpasses some models pre-trained on more data and/or with larger size.\footnote{
	Comparison of computational budget can be found in~\Apref{appendix:computational_budget}.
}

\subsection{Visualization of Aggregation Weights}
\label{sec:visualization}
We delve into managers by visualizing the average aggregation weights $\mathbf{W}_\mathrm{A}$ they generate across all samples in VQAv2 validation set in each cross-modal layer in~\fref{fig:aggregation-weights-pre-trained-aaum}.
For each row, the first column displays the learned aggregation weights of SAUMs, while the remaining five columns show the aggregation weights generated by AAUMs and share the Y-axis to provide easy horizontal comparison.

Interestingly, the aggregation weight distributions from managers are \textbf{completely different} from the one-hot distributions manually specified in \btname{}, and there are two distinct trends:
($i$) For SAUMs in the $1^{\text{st}}$ cross-modal layer, vertically, textual manager exhibits increasing and then decreasing weights, most favoring $\mathbf{T}_{10}$, unlike $\mathbf{T}_{12}$ and $\mathbf{T}_{7}$ used in \metername{} and \btname{}, respectively;
visual manager exhibits increasing weights, most favoring $\mathbf{V}_{12}$, similar to \metername{} and \btname{}.
($ii$) For AAUMs in the $2^{\text{nd}}$ to $6^{\text{th}}$ cross-modal layers, horizontally, whether textual or visual managers, they exhibit \textbf{diverse} aggregation weight distributions in different layers.

Overall, by comparing the aggregation weight distributions horizontally and vertically, we observe that \mtname{} learns \textbf{diverse} distributions in different cross-modal layers.
This provides strong evidence that the introduced managers can \textbf{adaptively} aggregate unimodal semantic knowledge for more comprehensive vision--language representation learning.
\section{Exploration on MLLM}
\label{sec:experiments_mllm}

\subsection{Motivation}
As stated in~\sref{sec:introduction}, in principle, the manager is a lightweight and flexible plugin that can be easily integrated into various VLMs.
Naturally, we can take the manager as a plugin and further explore its effectiveness in the latest MLLM architecture, which typically consists of a visual encoder and an LLM.

Moreover, traditional \twotower{} VLMs and MLLMs both use ViTs as their visual encoder, which have to resize the input image to a fixed resolution.
This greatly \textbf{limits} their effectiveness in handling high-resolution images due to the \textbf{loss} of \textbf{visual details}.
Recent multi-grid MLLMs~\cite{liu2024improved,li2024monkey,li2024llava} overcome this limitation by training with the multi-grid algorithm.\footnote{
	An illustration of the multi-grid algorithm can be found 
	in~\Apref{appendix:multi-grid}.
}
During training and inference, they divide the padded input image into multiple image grids, and encode both the resized base image and multiple image grids with the visual encoder independently.
Then, they combine the encoded features to obtain a longer input visual representation with more visual details.

Compared the manager with the multi-grid algorithm, they both can be seen as a \textbf{plugin} that improves the input visual representation and thus improves the VL representation.
They are two \textbf{orthogonal} directions to supplement visual details, either by ($i$) \textbf{deeper}: introducing aggregation of insights from pre-trained visual experts at different levels/depths; or ($ii$) \textbf{wider}: directly improving image resolution by encoding multiple image grids, \ie{} a wider receptive field.
Hence, we are motivated to explore the effectiveness of managers not only in MLLMs, but also in multi-grid MLLMs, to investigate the \textbf{synergy} between the manager and the multi-grid algorithm.

Besides, with the help of the MLLM architecture and the multi-grid algorithm, 
we can further \textbf{extend} downstream datasets, not only limited to traditional general datasets with low-resolution natural images, \eg{} VQAv2 and Flickr30K used in \sref{sec:experiments}, 
but also text-rich datasets with high-resolution abstract images (documents, charts, \etc{}), \eg{} DocVQA~\cite{mathew2021docvqa} and OCRBench~\cite{liu2024ocrbenchhiddenmysteryocr}, and real-world multimodal datasets.
Without fine-tuning on specific datasets, we can provide more \textbf{comprehensive} and \textbf{challenging} zero-shot evaluations of the effectiveness of managers.

Overall, we aim to explore the effectiveness of managers in more diverse downstream datasets, to answer the questions: (\textbf{RQ1}) Can the manager be used as a plugin to help MLLMs and multi-grid MLLMs? (\textbf{RQ2}) When and why can managers improve performance, especially for multi-grid MLLMs?

\subsection{Experimental Settings}

\subsubsection{Baseline}
We take LLaVA-OneVision-0.5B-SI~\cite{li2024llava} as our baseline (\ovname{} for short), which is a widely used open-source multi-grid MLLM.
It consists of a pre-trained $27$-layer visual encoder SigLIP~\cite{zhai2023sigmoid} with $0.4$B parameters, a pre-trained $24$-layer LLM Qwen2-0.5B-Instruct~\cite{yang2024qwen2} with $0.5$B parameters and a $2$-layer MLP with $1.8$M parameters.
It releases most of the training data, which helps us reproduce not only the multi-grid version (\basegname{}), but also the plain version (\basename{}).
We follow the same training settings as the original \ovname{} and use about $8$M data samples for multi-stage training of the autoreregressive objective for answer tokens.
The maximum length of the input token sequence is set to $16384$, and the image patch size is $14\!\times\!14$.
The last layer of the visual encoder is removed, and the visual representation of the penultimate layer is projected into the LLM word embedding space as the visual part of the input tokens of the LLM.
More details can be found in~\Apref{appendix:implementation_details_mllm}.\looseness=-1

\subsubsection{Adapt Manager to MLLM}
\label{sec:manager-design-mllm}
Since the LLM in MLLM acts as both a textual module and a cross-modal module, as shown in~\fref{fig:overview_mllm}, we \textbf{directly} introduce visual managers in \ovname{}, to aggregate multi-layer visual representations and inject them into the LLM at \textbf{equal} intervals, thus obtaining \ovmname{}.
Similar to \ovname{}, we train two versions of \ovmname{} and name them as \basemname{} and \basegmname{}, respectively.
Managers aggregate insights from the top half of the visual encoder to improve the visual representations of both the base image and image grids independently.
We inject $6$ visual managers into the LLM with the interval of $4$ as the default setting.\footnote{
	Ablation study for the default setting can be found in~\Sref{sec:manager-injection-times}.
}
Since AAUM achieves similar performance compared to SAUM in \ovmname{}, we directly use SAUM for better efficiency in the following experiments.\footnote{
	Discussions about managers in the MLLM can be found 
	in~\Apref{appendix:spirit-of-manager}.
}
For brevity, the $\ell^{\text{\,th}}$ LLM layer with SAUM computes as:
\begin{align}
	\tilde{\mathbf{C}}^\mathrm{V}_{\ell} & = \mathcal{M}_{\ell}^{\mathrm{V}}(\mathbf{V}_{14}, \dots, \mathbf{V}_{26}) \odot \epsilon + \mathbf{C}^\mathrm{V}_{\ell-1}, \label{eq:prev1_mllm} \\
	\mathbf{C}^\mathrm{V}_{\ell}, \mathbf{C}^\mathrm{T}_{\ell} & = \operatorname{Encoder}^\mathrm{C}_\ell(\tilde{\mathbf{C}}^\mathrm{V}_{\ell}, \mathbf{C}^\mathrm{T}_{\ell-1}), \label{eq:layer_mllm} \\
	\mathcal{M}_{\ell}^{\mathrm{V}}(\mathbf{V}_{14}, & \dots, \mathbf{V}_{26}, \mathbf{C}^\mathrm{V}_{\ell-1}) = \sum_{i=1}^{13}{\mathbf{W}_{i} \odot \mathbf{V}_{i+13}}. \label{eq:saue_mllm}
\end{align}
\Eref{eq:saue_mllm} is an optimized version of SAUM for MLLM.
The original version does not work well in our preliminary experiments, as the LLM in MLLM has been well pre-trained, rather than the random-initialized cross-modal module in \mtname{}.
Hence, we remove the $\mathbf{W}_{\mathrm{C}}$, $\operatorname{LN}$, and $\operatorname{softmax}$ in~\Eref{eq:saue}, and initialize $\mathbf{W}$ to zero, to \textbf{reduce} the interference with the pre-trained LLM in the early training stage~\cite{li2022exploring,zhang2023llama}, which helps SAUM work well in MLLM.
$\epsilon \!\sim\! \mathcal U(0.98, 1.02)$ is a multiplicative jitter noise uniformly sampled for exploration across experts during training~\cite{fedus2022switch}.

\subsubsection{Evaluation}
\label{sec:evaluation-mllm}
We follow the same evaluation settings as the original \ovname{}, to evaluate the zero-shot performance of our four baselines on $20$ datasets via their official evaluation tool, lmms-eval.\footnote{\url{https://github.com/EvolvingLMMs-Lab/lmms-eval}}
From the perspective of \textbf{capability categories}, we can divide them into the following four categories: 
\begin{itemize}[topsep=3pt,parsep=2pt,partopsep=2pt]
    \small
    \item General: VQAv2~\cite{balanced_vqa_v2}, OKVQA~\cite{okvqa}, GQA~\cite{hudson2019gqa}, MMVet~\cite{yu2024mmvet}, SEED-Bench~\cite{li2023seed}, RealWorldQA~\cite{xai2024realworldqa}.
    \item Text-rich: TextVQA~\cite{singh2019towards}, ChartQA~\cite{masry2022chartqa}, DocVQA~\cite{mathew2021docvqa}, InfoVQA~\cite{mathew2022infographicvqa}, OCRBench~\cite{liu2024ocrbenchhiddenmysteryocr}.
    \item Knowledge: AI2D~\cite{kembhavi2016diagram}, ScienceQA~\cite{lu2022learn}, MMMU~\cite{yue2023mmmu}, MathVista~\cite{lu2024mathvista}.
    \item Real-world: ImageDC~\cite{li2024llavanext-ablations}, MM-LiveBench (07, 09)~\cite{zhang2024lmms}, LLaVA-Wild~\cite{liu2024visual}, LLaVA-Wilder~\cite{liu2024improved}.
\end{itemize}
For simplicity, we use the average score of the corresponding metric score (normalize to $[0, 100]$) as the overall performance of baselines.
We also calculate the average score of each capability category for in-depth analysis.
Furthermore, since these datasets contain not only low-resolution natural images, but also high-resolution abstract images, we can also analyse and divide these datasets from the perspective of \textbf{image categories} ``Natural, Abstract, Hybrid'' and \textbf{resolutions} ``Low, High''.\footnote{
	More evaluation details can be found 
	in~\Apref{appendix:evaluation_details_mllm}.
}\looseness=-1
\begin{figure}[t]
	\centering
	\includegraphics[width=0.48\textwidth]{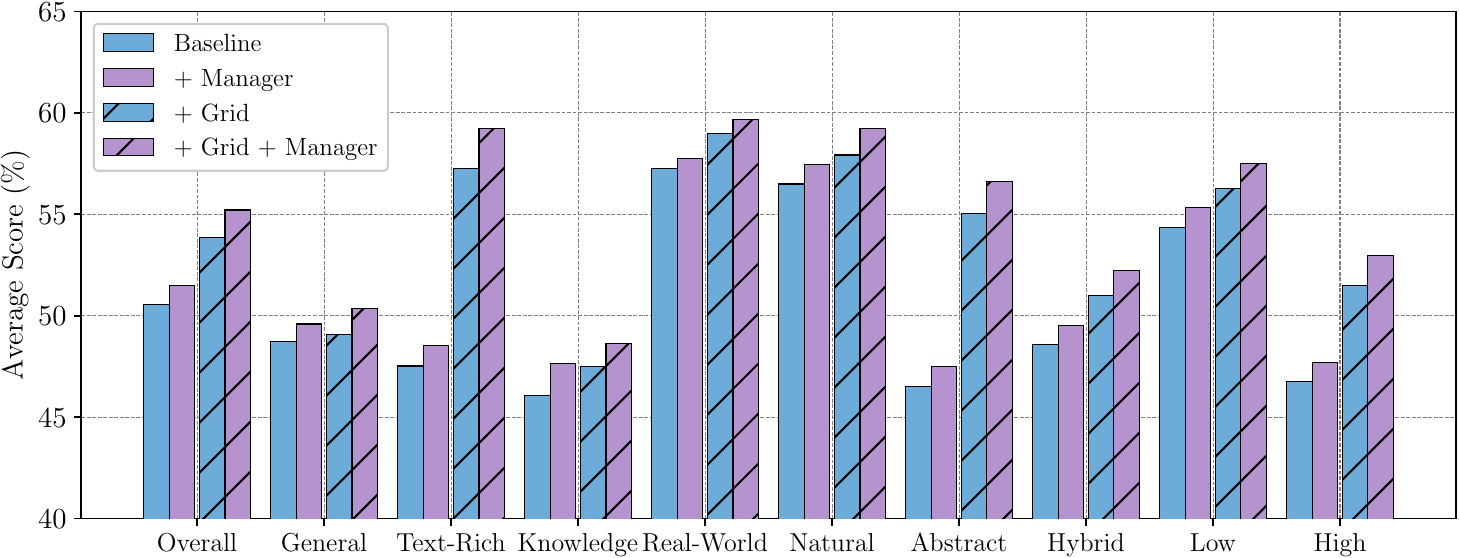}
	\caption{
        Zero-shot performance of four baselines on $20$ datasets.
        The overall average score and the average score of each capability category are shown.
	}
	\label{fig:main-results-mllm}
\end{figure}

\subsection{Results and Computational Budget}

\fref{fig:main-results-mllm} shows the zero-shot performance of four baselines on $20$ datasets after training with about $8$M data samples following the original \ovname{}.\footnote{
	Detailed results of each dataset can be found
	in~\Apref{appendix:detailed-results-mllm}.
}
The difference between baselines is with or without the multi-grid algorithm and managers.
Similar to existing multi-grid MLLMs, we can observe that the multi-grid algorithm greatly helps \basename{} and \basemname{}, especially on text-rich datasets, abstract images, and high-resolution images.
When introducing managers, whether the multi-grid algorithm is enabled or not, the performance of \basemname{} and \basegmname{} is \textbf{significantly} improved over the corresponding \basename{} and \basegname{} on different categories of capabilities, images, and resolutions.
Especially on datasets with capability category of ``General, Knowledge'', \basemname{} even achieves better performance than \basegname{} with significantly lower computational cost.

\Tref{tab:computational_budget_mllm} shows the computational budget and average overall performance of four MLLM baselines.
We measure the average training time based on two $8\times$NVIDIA A100 GPU servers, and the average inference time on VQAv2 validation set with a single A100 GPU.
Compare to \basename{}, the multi-grid algorithm significantly increases
FLOPs ($\times 1.78$), training time ($\times 4.39$), inference time ($\times 1.68$) and performance ($\uparrow\!3.26\%$).
Whether with or without the multi-grid algorithm, managers only brings \textbf{negligible} parameter overhead ($0.08$M), FLOPs ($\times 1.02$), and computational cost ($\times 1.04$), but \textbf{significantly} improves performance ($\uparrow\!1.06\%$ and $\uparrow\!1.44\%$) on $20$ datasets.\footnote{
    $1504.12/1469.34\approx1.02$, $54.17/51.95\approx1.04$, $24.45/23.47\approx1.04$ and $55.21-51.67=1.44$.
}

In summary, for our \textbf{RQ1}, 
\Fref{fig:main-results-mllm} and \Tref{tab:computational_budget_mllm} demonstrate that the manager is a \textbf{lightweight}, \textbf{efficient} and \textbf{effective} plugin that helps MLLMs and multi-grid MLLMs achieve \textbf{better} performance in different capability categories, image categories and resolutions, with \textbf{acceptable} computational cost.
More interestingly, the collaboration between managers and the multi-grid algorithm not only supplements \textbf{visual details} from the \textbf{depth} and \textbf{width} directions, respectively, to improve performance, but also further boosts performance by their synergy ($1.44\%\!>\!1.06\%$).

\subsection{Ablation Study on Adaptation of Managers in MLLMs}

In this section, we further explore the adaptation of managers in MLLMs.
We use $\frac{1}{4}$ of the training data ($2$M samples) and evaluate on $9$ datasets for efficiency and robustness.

\begin{figure}[t]
	\centering
	\includegraphics[width=0.48\textwidth]{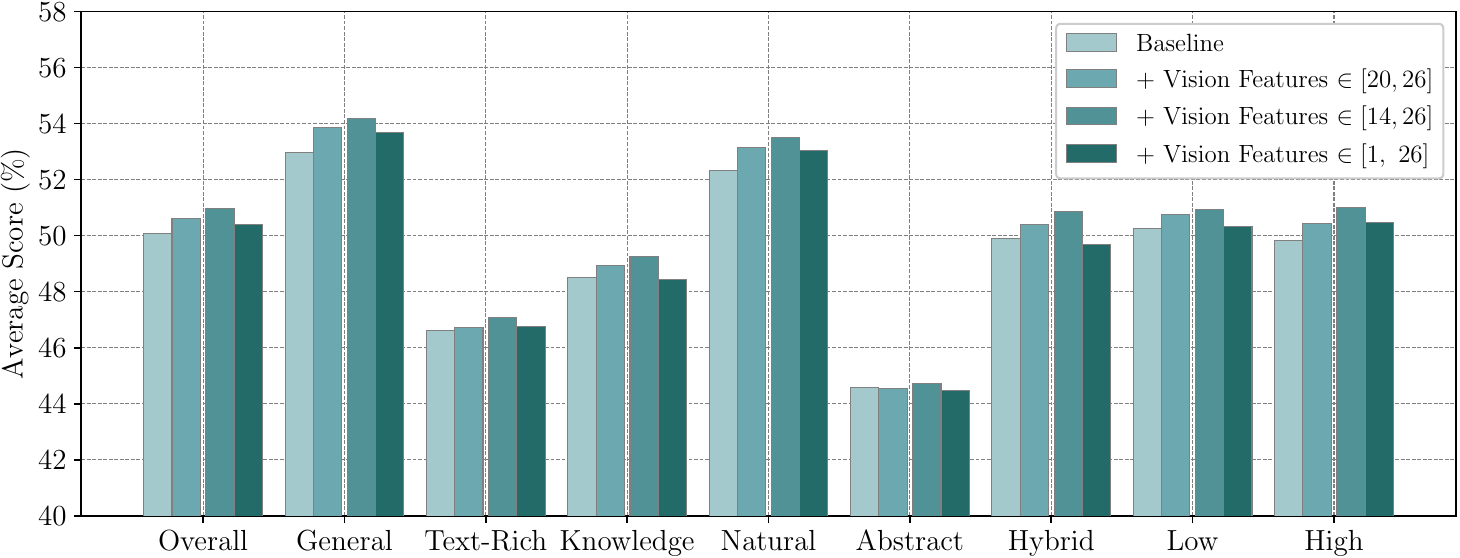}
	\caption{
        Ablation study of visual representation selection on $9$ datasets.
	}
	\label{fig:vision-selection}
\end{figure}

\begin{table}[t]
    \tablestyle{3pt}{1.1}
    \caption{
        Computational budget and average overall performance of four baselines on $20$ datasets.
        The numbers in parentheses denote the relative change compared to \basename{}.
    }
    \label{tab:computational_budget_mllm}
	\adjustbox{width=\linewidth}{
        \begin{tabular}{l|c|l|l|l|l}
            Model & \# Params & \multicolumn{1}{c|}{\# FLOPs} & \multicolumn{1}{c|}{Training Time} & \multicolumn{1}{c|}{Inference Time} & \multicolumn{1}{c}{Performance} \\
            & (M) & \multicolumn{1}{c|}{(G)} & \multicolumn{1}{c|}{(ms/sample)} & \multicolumn{1}{c|}{(ms/sample)} & \multicolumn{1}{c}{Overall (\%)} \\
            \shline
            Baseline         & 893.62 & { }\,827.29 & 11.84         & 13.97         & 50.61         \\
            \hdashline
            + Manager        & 893.70 & { }\,844.68 ($\times$1.02) & 12.22 ($\times$1.03) & 14.54 ($\times$1.04) & 51.67 ($\uparrow$1.06) \\
            + Grid           & 893.62 & 1469.34 ($\times$1.78) & 51.95 ($\times$4.39) & 23.47 ($\times$1.68) & 53.87 ($\uparrow$3.26) \\
            + Grid + Manager & 893.70 & 1504.12 ($\times$1.82) & 54.17 ($\times$4.58)  & 24.45 ($\times$1.75) & \textbf{55.21} ($\uparrow$4.60)
            \end{tabular}
        }
\end{table}

\begin{figure}[t]
	\centering
	\includegraphics[width=0.45\textwidth]{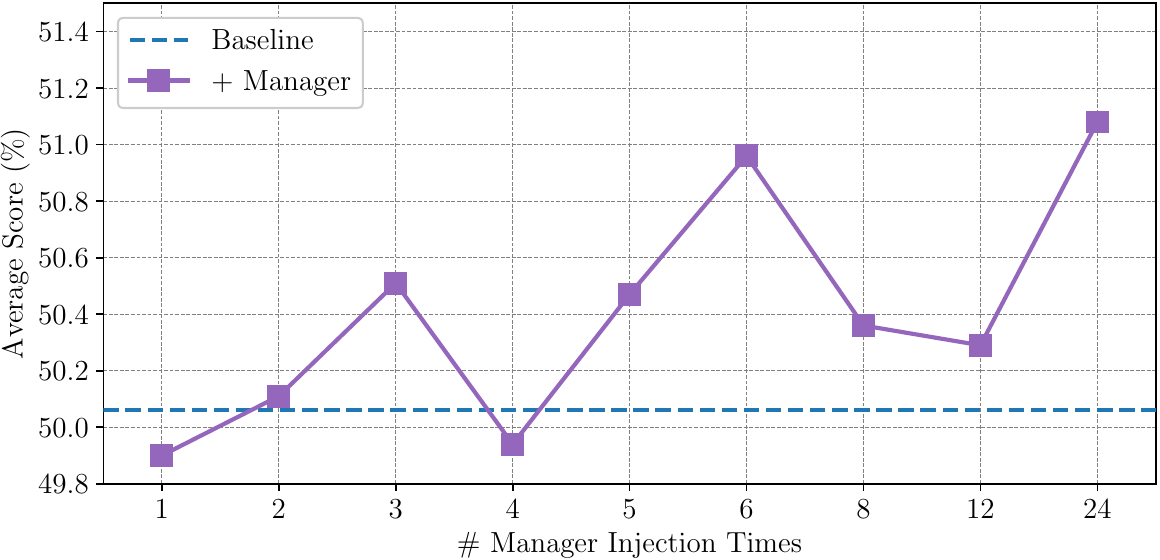}
	\caption{
        Ablation study of manager injection times on $9$ datasets.
	}
	\label{fig:manager-injection}
\end{figure}

\begin{figure}[t]
	\centering
	\includegraphics[width=0.45\textwidth]{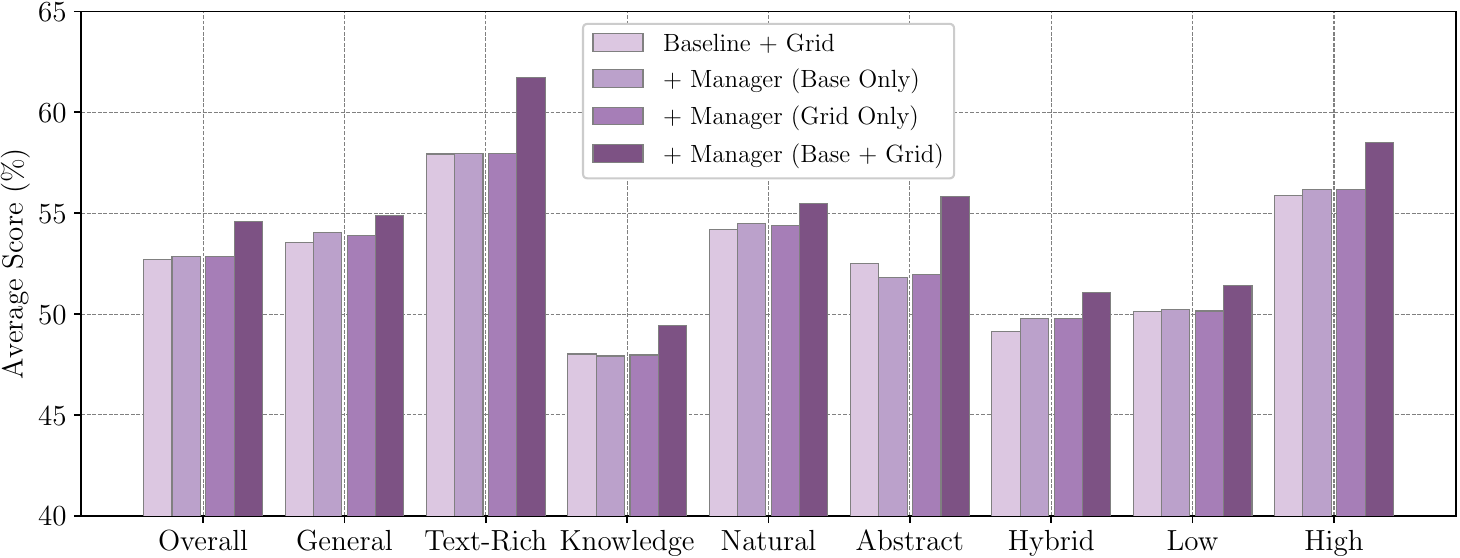}
	\caption{
        Ablation study of how manager works with multi-grid on $9$ datasets.
	}
	\label{fig:manager-grid-type}
\end{figure}

\subsubsection{Visual Representation Selection}
\label{sec:visual-representation-selection}

As shown in~\fref{fig:vision-selection}, overall, no matter what visual representations are selected, managers \textbf{consistently} improve the performance of \basename{}.
Similar to the observations in both \btname{} and \mtname{}, visual representations from the \textbf{top} half of the visual encoder bring the best performance, and using visual representations from all layers leads to the lowest performance improvement.
We attribute this to the fact that the average attention distance of the visual encoder increases with the layer depth, especially in the top half of the visual encoder, where most attention heads attend \textbf{widely} across tokens~\cite{dosovitskiy2020image} and capture global visual features.\footnote{
    Detailed explanations and visualizations are provided 
	in~\Apref{appendix:attention_distance}
}

\subsubsection{Manager Injection Times}
\label{sec:manager-injection-times}

We uniformly inject managers into the LLM from the first layer at a fixed layer interval.
Specifically, for the LLM with $L_\mathrm{C}\!=\!24$, we can inject $6$ managers with the interval of $4$.
As shown in~\fref{fig:manager-injection}, the injection times of managers will affect the performance, and the overall trend is that performance improves with increasing injection frequency, but with some fluctuations.
\basemname{} can achieve \textbf{better} performance than \basename{} most of the time.
Compared to the injection times of $6$, although injecting managers into each LLM layer slightly increases the average performance from $50.96\%$ to $51.08\%$, it also \textbf{increases} the computational cost by about $7\%$ in both training and inference.
Hence, we choose the injection times of $6$ to achieve a good \textbf{balance} between performance and computational cost.

\begin{figure}[t]
	\centering
	\includegraphics[width=0.48\textwidth]{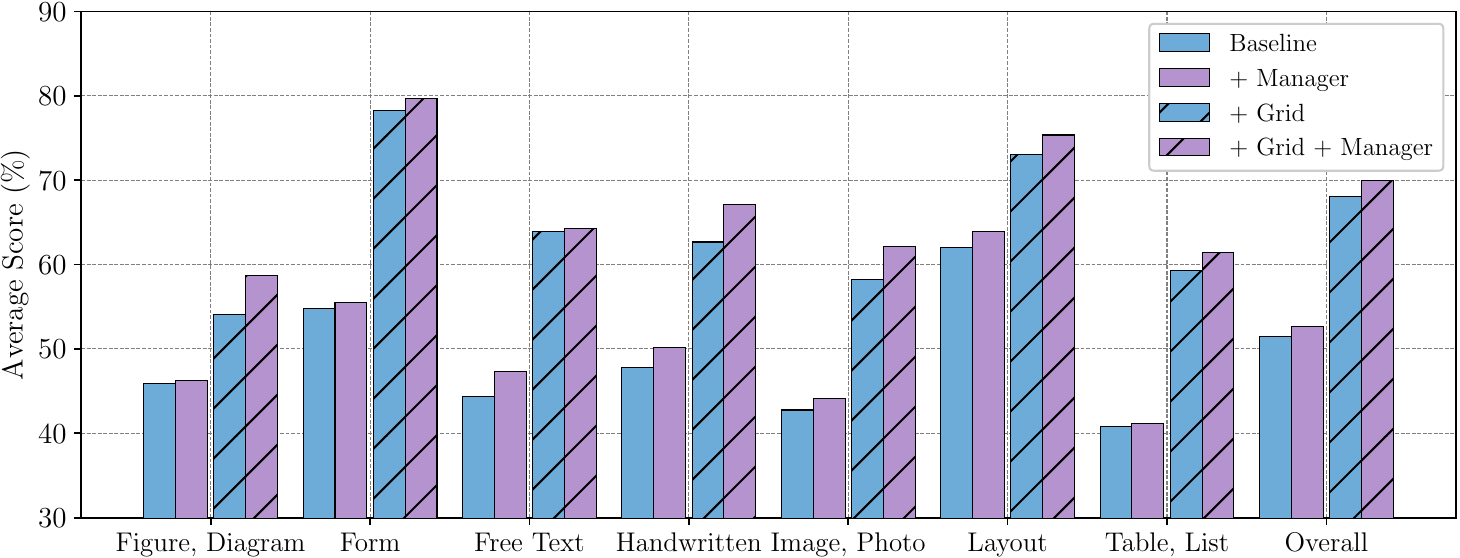}
	\caption{
        Zero-shot performance of four baselines on DocVQA validation set.
	}
	\label{fig:category_docvqa}
\end{figure}

\begin{figure}[!h]
	\centering
	\includegraphics[width=0.45\textwidth]{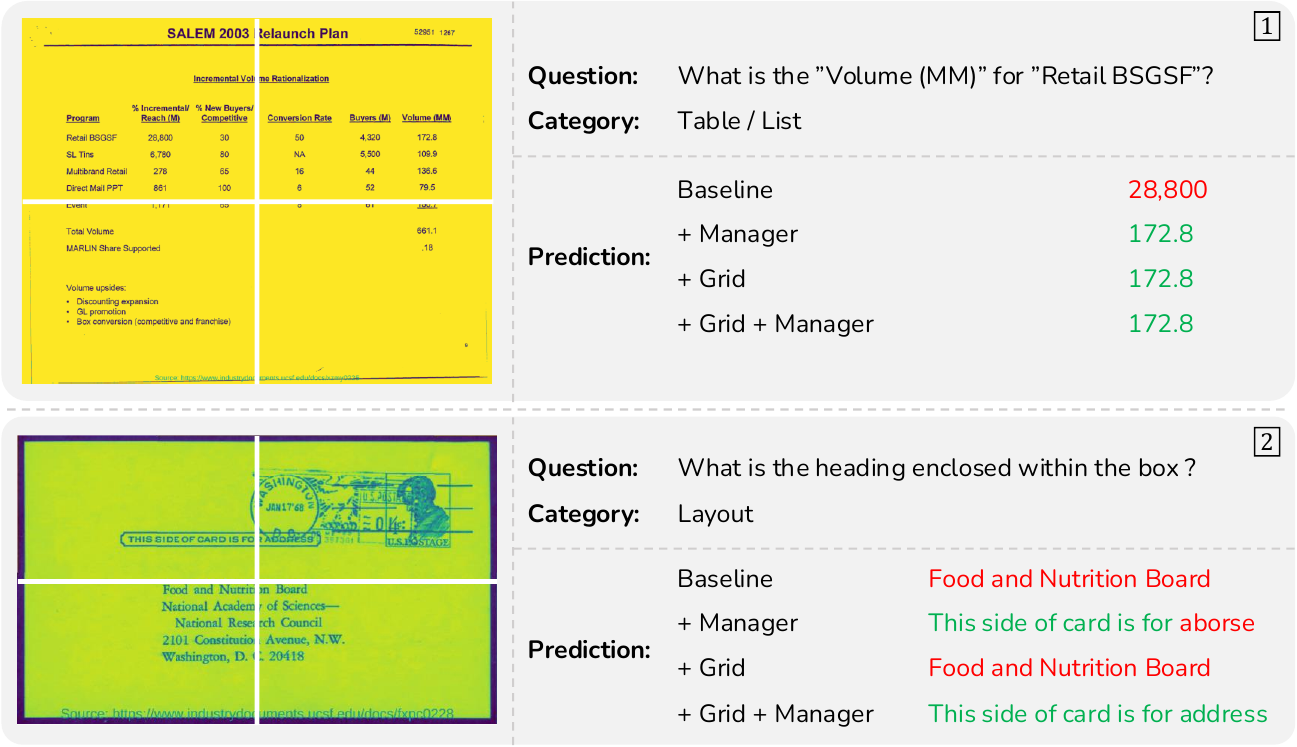}
	\caption{
        Case studies of four baselines on DocVQA validation set.
        Red and green fonts represent incorrect and correct predictions, respectively.
        White lines indicate the boundaries of the image grids.
	}
	\label{fig:case_docvqa}
\end{figure}

\subsubsection{Manager Meets Multi-Grid}

Both the manager and the multi-grid algorithm are plugins that can be easily combined and integrated into MLLMs.
Their direct combination means that managers aggregate insights from pre-trained visual experts at different levels to improve the visual representations of the base image and multiple image grids, respectively.
As shown in~\fref{fig:manager-grid-type}, managers \textbf{greatly} improve the performance of \basegname{}, especially on text-rich datasets, abstract images, and high-resolution images, which are exactly what the multi-grid algorithm excels at.
This indicates that the manager and the multi-grid algorithm are orthogonal (depth and width) and \textbf{complementary} in complementing visual details, and their synergy can further improve performance.
More interestingly, when managers only manage the base image or image grids, the performance is not obviously improved.
We speculate that the change in part of the visual representation by managers may be considered as \textbf{noise} due to the numerical difference between the changed and unchanged parts.

\begin{figure}[t]
    \centering
    \includegraphics[width=0.48\textwidth]{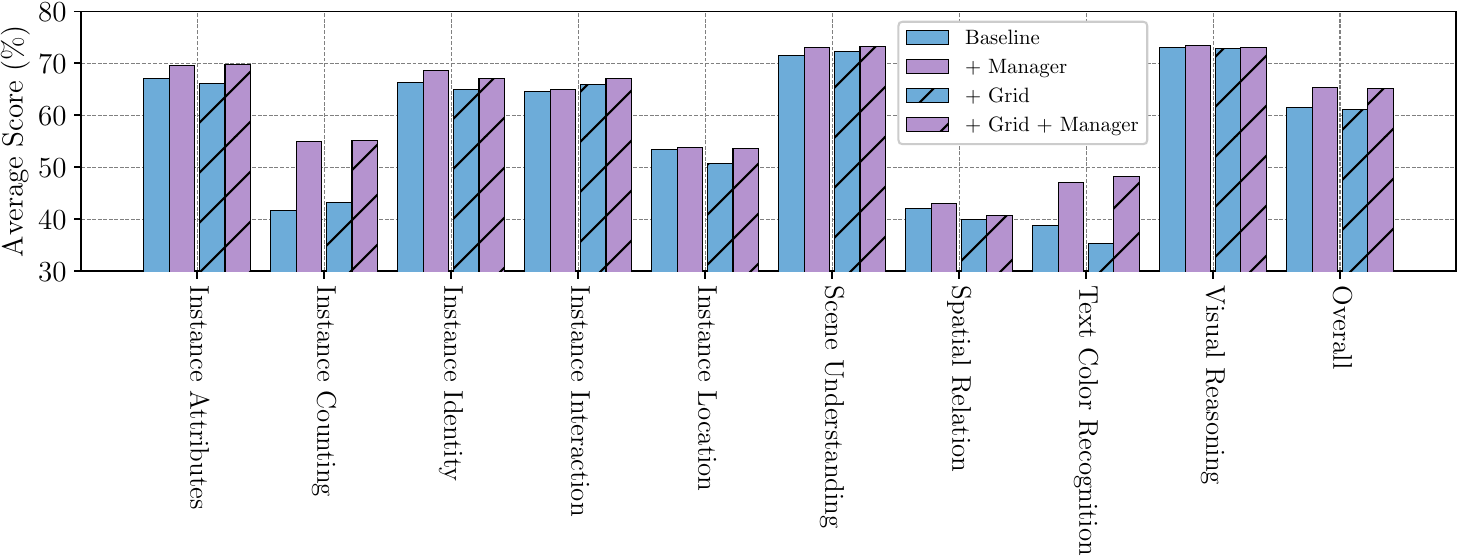}
    \caption{
        Zero-shot performance of four baselines on SEED-Bench.
    }
    \label{fig:category_seed_bench}
\end{figure}

\begin{figure}[t]
	\centering
	\includegraphics[width=0.48\textwidth]{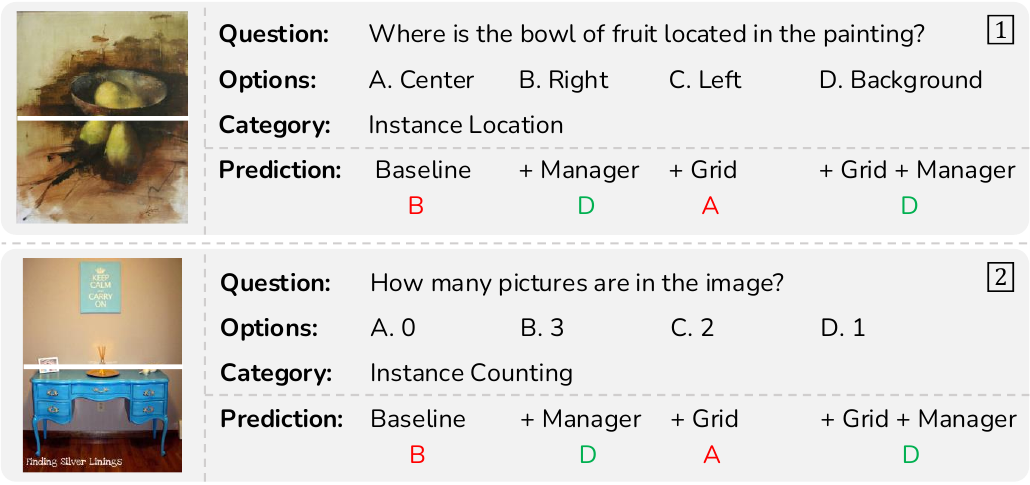}
	\caption{
        Case studies of four baselines on SEED-Bench.
	}
	\label{fig:case_seed_bench}
\end{figure}

\begin{figure}[t]
    \centering
    \includegraphics[width=0.48\textwidth]{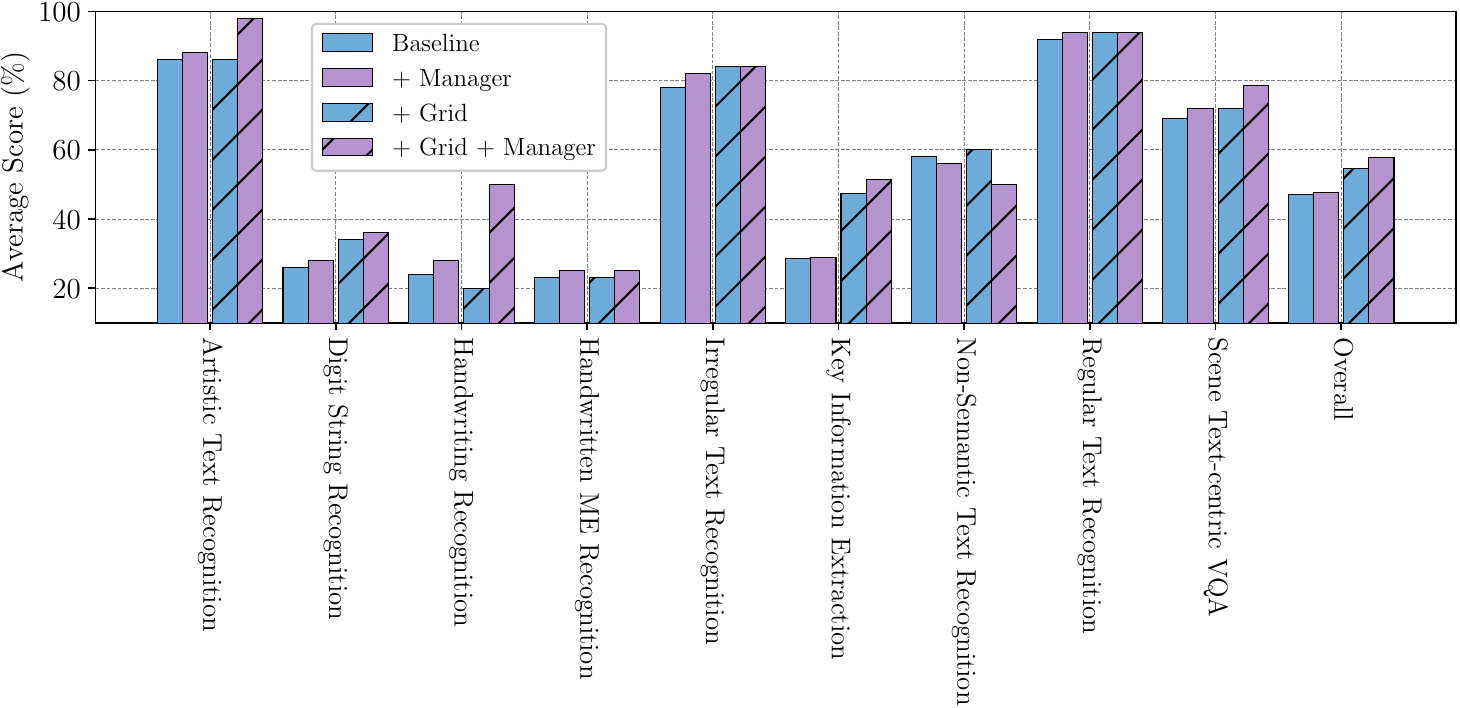}
    \caption{
        Zero-shot performance of four baselines on OCRBench.
        ``ME'' in ``Handwritten ME Recognition'' is short for ``Mathematical Expression''.
    }
    \label{fig:category_ocrbench}
\end{figure}

\begin{figure}[t]
	\centering
	\includegraphics[width=0.48\textwidth]{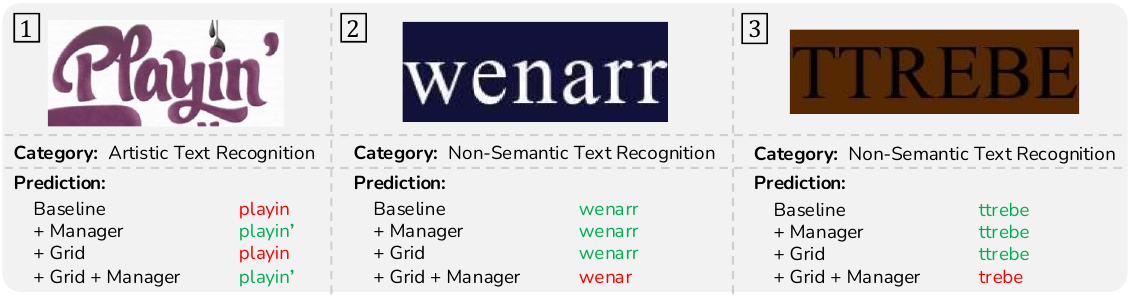}
	\caption{
        Case studies of four baselines on OCRBench.
	}
	\label{fig:case_ocrbench}
\end{figure}

\subsection{Detailed Analysis and Case Study}
\label{sec:detailed-analysis-mllm}
To intuitively analyse the effectiveness of managers and answer our \textbf{RQ2}, we conduct a detailed analysis on different dimensions of specific datasets, including DocVQA, SEED-Bench, and OCRBench, and provide case studies.\footnote{
	More detailed analysis and case studies on ScienceQA and OK-VQA can be found 
	in~\Apref{appendix:detailed-analysis-mllm}.
}

\subsubsection{DocVQA}
Based on the three dataset classification criterion we used in~\Sref{sec:evaluation-mllm}, DocVQA is a text-rich dataset with high-resolution abstract images.
As shown in~\fref{fig:category_docvqa}, the multi-grid algorithm helps \basename{} on different types of abstract images in DocVQA.
Furthermore, managers can further improve the performance of \basename{} and \basegname{} on different dimensions.
Take the case {\scriptsize $\boxed{1}$} in ~\fref{fig:case_docvqa} as an example, both managers and the multi-grid algorithm can help \basename{} \textbf{capture} visual details for accurate table understanding.
Interestingly, in the case {\scriptsize $\boxed{2}$}, both \basename{} and \basegname{} fail to find the heading enclosed within the box, and take the first line of text below the box as the heading.
The multi-grid algorithm also \textbf{cuts off} the boxed heading, may make it more difficult to find the heading.
\basemname{} can \textbf{correctly} find it based on the visual details provided by different levels of semantic knowledge, but fails to recognize all characters.
With the \textbf{collaboration} between the manager and the multi-grid algorithm, \basegmname{} can correctly find it and recognize all characters.

\subsubsection{SEED-Bench}
This is a general dataset with high-resolution natural images.
Surprisingly, as shown in~\fref{fig:category_seed_bench}, the multi-grid algorithm does not improve the performance much and even leads to performance \textbf{degradation} on some dimensions, \ie{} ``Instance Identity, Instance Location, Spatial Relation, Text Color Recognition''.
They inspect the category, spatial and color information about instances in the image.
Take~\fref{fig:case_seed_bench} as an example, the multi-grid algorithm \textbf{cuts off} objects and connected regions, leading to higher understanding difficulty and bringing \textbf{semantic ambiguity}~\cite{huang2024mini}.
This \textbf{hinder} MLLMs from perceiving the spatial relationship between objects as well as the category and number of objects.
Moreover, managers \textbf{consistently} brings performance improvements to \basename{} and also help \textbf{overcome} the semantic ambiguity caused by the multi-grid algorithm by incorporating aggregation of \textbf{insights} from pre-trained visual experts at different levels, especially on ``Instance Counting, Text Color Recognition''.\looseness=-1

\subsubsection{OCRBench}
This is a text-rich dataset with low-resolution hybrid images.
As shown in~\fref{fig:category_ocrbench}, for ``Artistic Text Recognition, Handwriting Recognition'' dimensions, both the manager and the multi-grid algorithm can only bring slight performance improvements or even performance degradation to \basename{}.
However, the collaboration between them can bring \textbf{significant} performance improvements on \basegmname{}.
This further demonstrates that their \textbf{synergy} can complement \textbf{visual details} from the depth and width directions and mitigate the semantic ambiguity caused by the multi-grid algorithm.
\textbf{Unexpectedly}, for ``Non-Semantic Text Recognition'' dimension, which focuses on character combinations that are meaningless or lack semantics, the manager brings performance degradation to both baselines.
Take the cases in~\fref{fig:case_ocrbench} as an example, although managers can help capture visual details, \eg{} a single quote at the end of the word, 
\basegmname{} \textbf{incorrectly} identifies the \textbf{non-semantic} text ``wenar'' and ``ttrebe'' as semantic text ``wenar'' and ``trebe'', respectively, where ``wenar'' is a surname of a person and ``trebe'' is a German noun for a runaway.
Different levels of semantic knowledge brought by managers instead cause more interference, leading to performance degradation when work with the multi-grid algorithm in ``Non-Semantic Text Recognition''.\looseness=-1

In summary, for our \textbf{RQ2}, the manager can not only \textbf{improve} the performance of MLLMs, but also help \textbf{alleviate} the semantic ambiguity caused by the multi-grid algorithm.
Hence, their \textbf{synergy} can further improve performance, especially on the perception of category, spatial, color and number information of instances, and artistic, handwriting text recognition.

\begin{figure}[t]
	\centering
	\includegraphics[width=0.48\textwidth]{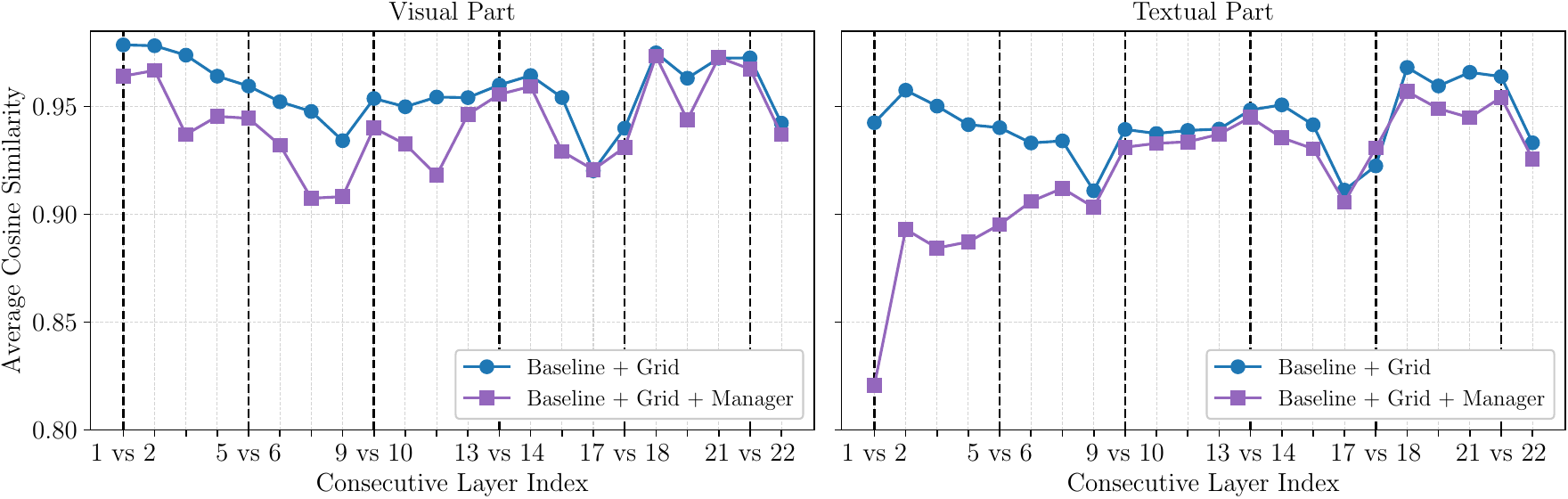}
	\caption{
        Cosine similarity between output representations of consecutive layers.
        The dotted vertical lines indicate the layers where managers are injected, \ie{} \# Layer Index$=[1,5,9,13,17,21]$.
	}
	\label{fig:visualization_cos_sim}
\end{figure}

\begin{figure}[t]
	\centering
	\includegraphics[width=0.48\textwidth]{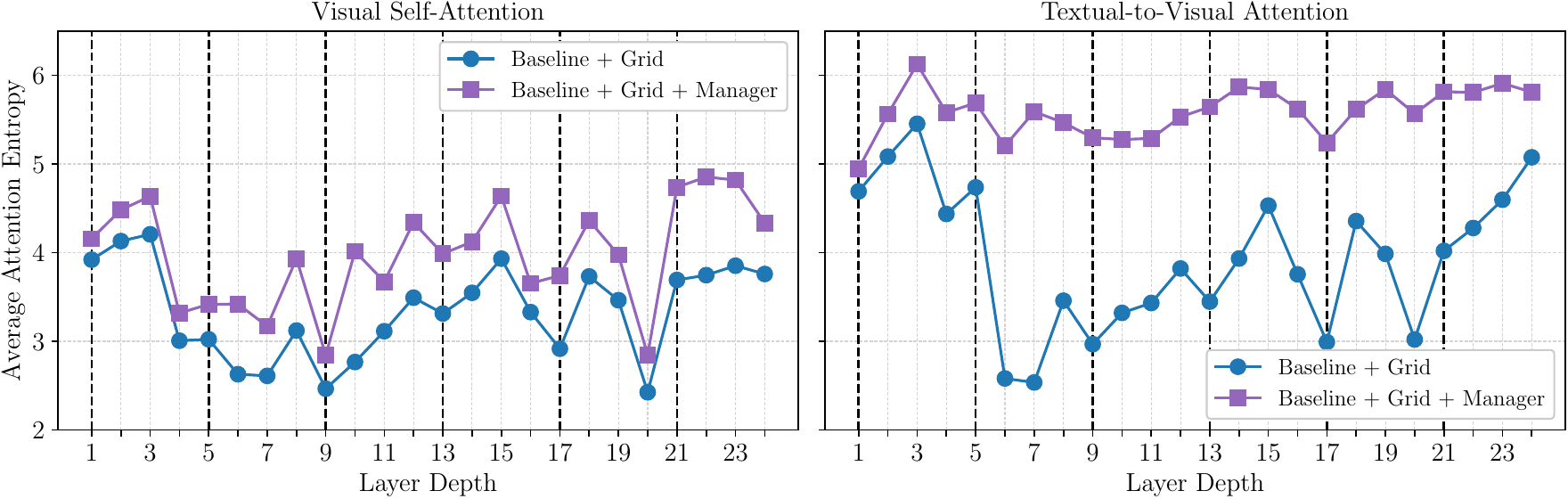}
	\caption{
        Average entropy of attention weight distributions in each layer.
	}
	\label{fig:visualization_entropy}
\end{figure}

\begin{figure}[t]
	\centering
	\includegraphics[width=0.48\textwidth]{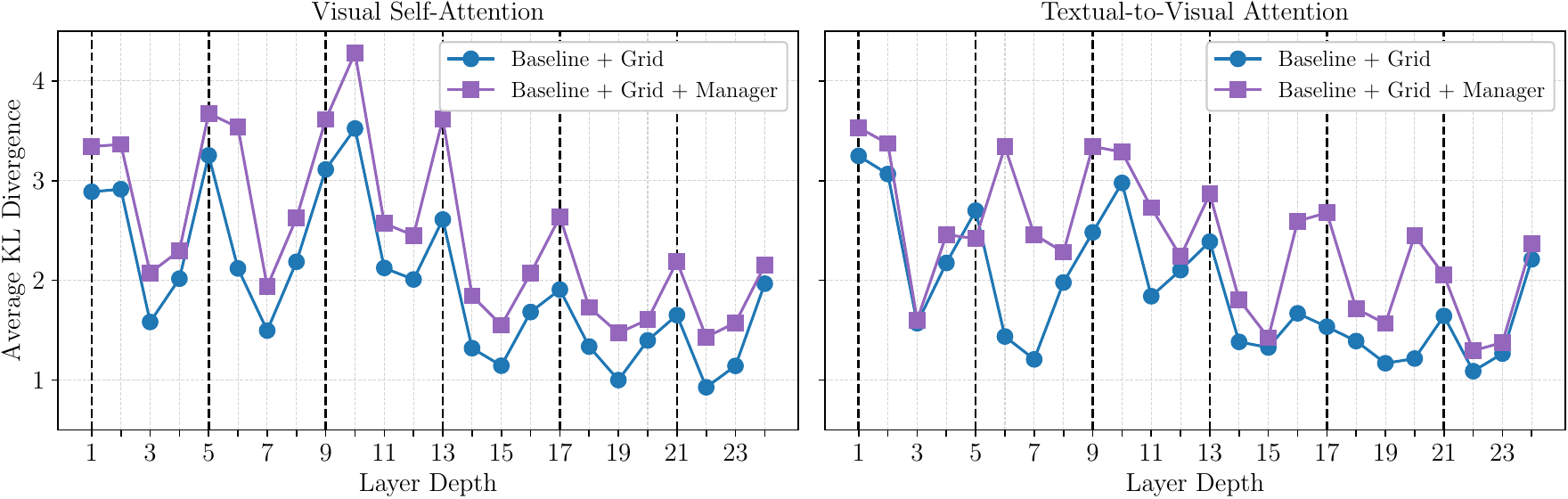}
	\caption{
        Average KL divergence between attention weight distributions of attention heads in each layer.
	}
	\label{fig:visualization_kl}
\end{figure}

\subsection{Visualization Analysis}
\label{sec:visualization-analysis-mllm}

To analyse the underling reasons for the \textbf{collaboration} improvement between the manager and the multi-grid algorithm in MLLMs and further answer our \textbf{RQ2}, we conduct analyses from the perspective of consecutive layer representation similarity and attention weight distribution of each layer.

\subsubsection{Consecutive Layer Representation Analysis}
In~\Eref{eq:layer_mllm}, the output representation of each LLM layer consists of a visual part and a textual part.
For each part, we calculate the cosine similarity between output representations of consecutive layers in \basegname{} and \basegmname{}.
As shown in~\fref{fig:visualization_cos_sim}, managers \textbf{reduce} the similarity between representations of consecutive layers, especially for the \textbf{bottom} layers of MLLMs.
Compare to \basegname{}, changes in the similarity become more frequent and drastic in the layers between manager \textbf{injections}.
This indicates that the aggregation of different levels of semantic knowledge introduced by managers can \textbf{supplement} more insights and visual details, and facilitate more \textbf{diverse} vision--language representation learning in subsequent layers.
It is worth noting that although we do not have textual managers, the textual part of the output representation is \textbf{causally} influenced by the visual part in its front, resulting in a similar phenomenon.

\subsubsection{Attention Weight Distribution Analysis}
\label{sec:attention-weight-distribution-analysis}
The attention mechanism~\cite{bahdanau2014neural} is a key component in deep neural networks, where attention weight distributions reflect how much attention each token pays to the other tokens.
Following~\cite{xie2022revealing}, we delve into attention weight distributions from the following two angles to provide an intuitive and interpretable analysis.
Besides, for the attention weight distribution of each layer, we focus on the self-attention of the visual part, and the attention from the textual part at the back to the visual part at the front.\footnote{
	Attention weight distribution analysis of \basename{} and \basemname{} can be found
	in~\Apref{appendix:attention-weight-distribution-analysis}.
}

\paragraph{Attention Entropy}
The average entropy of attention weight distributions reflects the \textbf{diversity} of attention weights in each layer.
Higher/lower attention entropy means that the attention weights are concentrated on \textbf{more/few} tokens.
As shown in~\fref{fig:visualization_entropy}, compared to \basegname{}, managers \textbf{increase} the attention entropy in each layer.
Such \textbf{broad} attention can help \basegmname{} handle more complex and varied input, leading to greater diversity and flexibility, and thereby preventing focusing too narrowly on certain aspects of the input.
Besides, interestingly, the entropy of textual-to-visual attention becomes more stable and significantly larger than the entropy of visual self-attention when managers manage the visual part of the input.

\paragraph{KL Divergence}
The average Kullback–Leibler (KL) divergence~\cite{kullback1951information} between attention weight distributions of different attention heads reflects the \textbf{diversity} of attention heads in each layer.
Higher/lower KL divergence means that different attention heads pay attention to \textbf{different/similar} tokens.
As shown in~\fref{fig:visualization_kl}, compared to \basegname{}, managers \textbf{increase} the KL divergence between attention heads in most layers.
Intuitively, low diversity across different attention heads may limit the model's ability to capture diverse features.
Managers can help \basegmname{} focus on different aspects of the sequence to capture more \textbf{diverse} features, and prevent excessive focus on similar or redundant information.

In summary, for our \textbf{RQ2}, the manager introduces the aggregation of \textbf{insights} from visual experts at different levels into multi-grid MLLMs, which can \textbf{increase} the \textbf{diversity} of attention weights and attention heads. 
This can help \textbf{guide} the attention of multi-grid MLLMs, thus capturing more diverse visual details from both the manager (\textbf{depth}) and the multi-grid algorithm (\textbf{width}) directions, and also alleviating the semantic ambiguity caused by the multi-grid algorithm.

\section{Related Work}

\subsection{Vision--Language Models}
Although VLMs differ in model architecture, most of them use unimodal encoders to extract visual and textual representations, and then fuse them in a cross-modal module, which can be unified into the \twotower{} architecture~\cite{lu2019vilbert,su2019vl,chen2020uniter,li2020unicoder,li2020oscar,kim2021vilt,radford2021learning,li2021align,li2021unimo,li2022blip,dou2021meter,wang2021vlmo,wang2021simvlm,wang2022OFA,wang2022image,yu2022coca,10659916,10505301}.\footnote{
    Detailed discussion of the related work for multimodal fusion from the perspective of architecture can be found in~\Apref{appendix:related-work-multimodal-fusion}
}
As a representative model, \metername{}~\cite{dou2021meter} adopts pre-trained unimodal encoders and feeds their last-layer representations into the cross-modal encoder with the co-attention mechanism.
\btname{}~\cite{xu2023bridgetower} proposes building layer-by-layer connections between the top unimodal layers and each cross-modal layer to leverage multi-layer unimodal representations.
However, they still cannot utilize adaptive and effective aggregation of multi-layer pre-trained unimodal representations in each cross-modal layer.

\subsection{Utilization of Multi-Layer Unimodal Representations}
Different layers of pre-trained unimodal encoders encoding different levels of semantic knowledge are well demonstrated in vision~\cite{dosovitskiy2020image,raghu2021vision,naseer2021intriguing} and language~\cite{peters-etal-2018-dissecting,liu2019linguistic,jawahar-etal-2019-bert}.
As shown in prior work~\cite{dosovitskiy2020image,raghu2021vision}, lower layers of ViTs tend to attend both locally and globally, while higher layers primarily focus on global features.
Similarly, previous work~\cite{jawahar-etal-2019-bert} found that the intermediate layers of BERT~\cite{devlin-etal-2019-bert} encode a hierarchy of linguistic knowledge, with surface features at the bottom, syntactic features in the middle, and semantic features at the top.\looseness=-1

Furthermore, the effectiveness of multi-layer representation aggregation in learning comprehensive representations has been well demonstrated in vision~\cite{lin2017feature,huang2017densely,yu2018deep,xie2021segformer,10098834,10485473,10753025} and language~\cite{peters-etal-2018-deep, wang-etal-2018-multi-layer,wang-etal-2019-learning-deep,wei-etal-2020-multiscale}.
Hence, some \twotower{} VLMs and MLLMs have explored the utilization of pre-trained multi-layer unimodal representations for better vision--language representation learning~\cite{dou2021meter,xu2023bridgetower,dou2022fiber,yao2024dense,li2024tokenpacker}.
They simply feed the weighted sum or fusion of multi-layer unimodal representations into the first cross-modal layer, or exploit multiple top unimodal layer representations layer by layer in each cross-modal layer, which is not only ineffective but also lack scalability.
In this work, we take each layer of the pre-trained unimodal encoder as an unimodal \textbf{expert}, and the output representation of each layer as the \textbf{insight} of the unimodal expert into the current input.
We propose managers to \textbf{adaptively} aggregate insights from unimodal experts at different levels for each cross-modal layer.\looseness=-1

\subsection{Multimodal Large Language Models}
With the rapid development of Large Language Models (LLMs)~\cite{brown2020language,touvron2023llama2,yang2024qwen2,qin2024large}, MLLMs, a new class of VLMs that introduces a LLM as both a textual module and a cross-modal module, have emerged and shown superior zero-shot performance on various downstream tasks~\cite{li2023blip,liu2024improved,li2024llava}.
Although most existing MLLMs only feed the last-layer visual representation from the visual encoder into the LLM for simplicity and efficiency, some of them have explored different ways to improve the visual representation to further improve performance, especially high-resolution scenarios, such as:
($i$) adopt high-resolution visual encoders~\cite{fuyu-8b,li2024mini,wang2024qwen2,liu2024oryx}, which require additional high-resolution training data;
($ii$) adopt the multi-grid algorithm to directly split the image into multiple image grids~\cite{lin2023sphinx,li2024monkey,liu2024llavanext,shi2025we}, which is a resource-efficient way but may bring semantic ambiguity~\cite{liu2024textmonkey,huang2024mini}.
Since both the manager and the multi-grid algorithm can be viewed as a plugin that improves the visual representation from two orthogonal perspectives (\textbf{depth} and \textbf{width}), we further explore the effectiveness of managers in MLLMs and multi-grid MLLMs and the underlying reasons for their \textbf{collaboration} to improve performance based on extensive experiments and detailed analyses.\looseness=-1
\section{Conclusion}
\label{sec:conclusion}
In this work, we propose Manager, a \textbf{lightweight}, \textbf{efficient} and \textbf{effective} plugin that helps better utilize multi-layer pre-trained unimodal representations for vision--language representation learning, and demonstrate its effectiveness in both \twotower{} VLM and MLLM architectures.
The manager can \textbf{adaptively} aggregate more required unimodal semantic knowledge to facilitate comprehensive vision--language alignment and fusion in each cross-modal layer.
We first propose \mtname{}, a novel \twotower{} VLM that aggregates \textbf{insights} from pre-trained unimodal experts at different levels via introduced managers in each cross-modal layer.
The feasibility of various designs of managers is well explored, and the effectiveness of \mtname{} on $4$ downstream tasks is well demonstrated.
Next, we further validate the effectiveness of managers in the latest MLLM architecture.
Managers can \textbf{significantly} improve the zero-shot performance of MLLMs and multi-grid MLLMs on $20$ downstream datasets across different categories of capabilities, images, and resolutions.
Both the manager and the multi-grid algorithm can be seen as a \textbf{plugin} that improves the visual representation from two orthogonal perspectives (\textbf{depth} and \textbf{width}).
Their synergy can capture and supplement more \textbf{diverse} visual details, to mitigate the semantic ambiguity caused by the multi-grid algorithm and further improve performance.

\clearpage

\hypersetup{linkcolor=darkblue}

\onecolumn
\renewcommand{\contentsname}{Contents}
\tableofcontents

\twocolumn

\clearpage

\hypersetup{linkcolor=linkcolor}

\appendices

\section{Discussions, Limitations and Future Work}
In this paper, we propose Manager, a \textbf{lightweight}, \textbf{efficient} and \textbf{effective} plugin that helps better utilize multi-layer pre-trained unimodal representations for vision--language representation learning.
We demonstrate its effectiveness in both \twotower{} VLM and MLLM architectures on $4$ and $20$ downstream datasets, respectively.

\subsection{The Spirit of Manager}
\label{appendix:spirit-of-manager}
In~\Sref{sec:manager-design},
under the \twotower{} VLM architecture, we introduce three types of managers, \ie{} SAM, SAUM, and AAUM.
We also provide an adaptation of SAUM to the latest MLLM architecture 
in~\Sref{sec:manager-design-mllm}.
All managers are designed to aggregate insights from different levels of pre-trained unimodal experts, \ie{} incorporate different levels of unimodal semantic knowledge contained in the multi-layer unimodal representations from the pre-trained unimodal encoders.
The detailed implementation of these managers obey the following design principles:
\begin{itemize}
	\item \textbf{Lightweight and Flexible}: The manager should be a lightweight and flexible plugin that can be easily integrated into any VLM architecture and work with any pre-trained unimodal encoders.
	\item \textbf{Effective and Efficient}: The manager should be effective in aggregating multi-layer unimodal representations and efficient in terms of computational budget.
	\item \textbf{Aggregation and Multiple Injection}: The manager should aggregate multi-layer unimodal representations and inject them into the different layers of the cross-modal encoder in a flexible and adaptive way to facilitate more comprehensive VL alignment and fusion.
	\item \textbf{Spatial Consistency}: The manager should maintain spatial consistency (or locality) when aggregate multi-layer unimodal representations, which is crucial for vision--language alignment and fusion~\cite{cha2024honeybee}.
\end{itemize}

It would be interesting to explore more types of managers, more efficient and effective designs of managers, and more flexible and adaptive ways to select which layers of unimodal representations to aggregate and how to inject them into different layers of the cross-modal encoder or the LLM in the future.
\begin{itemize}
	\item \textbf{Manager Design}: Although AAUM achieves better performance under the \twotower{} VLM architecture with the help of the cross-modal fused query,  it also slightly increases the computational budget, as we detailed discussed in~\Apref{appendix:computational_budget}. 
	More analysis and optimization are needed for AAUM and also for the other types of managers as shown in~\Apref{appendix:other-managers}.
	Especially for AAUM, in the MLLM architecture, it actually achieves similar or even slightly lower performance compared to SAUM in \ovmname{}.
	Hence, we directly use SAUM for better efficiency and effectiveness in our experiments.
	In our preliminary experiments, we try different ways to get a good visual query or cross-modal fused query for AAUM in the MLLM architecture, but the performance is still not as good as SAUM.
	We attribute this to the fact that, the visual query used by AAUM to generate the aggregation weights is quite different between the \twotower{} VLM and MLLM architectures.
	They are taken from bidirectional / unidirectional transformer encoder / decoder, respectively.
	The casual nature of the representation from the LLM in the MLLM architecture may not be suitable for the usage of visual query in AAUM.
	It contradicts the bidirectional nature of multi-layer visual representations from the visual encoder.
	Use bidirectional and casual attention for the visual and textual part of the LLM, respectively, may be a potential solution to this problem~\cite{dong2019unified,hao2022language,liu2024prismer}.
	\item \textbf{Visual Representation Selection}: 
	As shown in~\Fref{fig:number-of-unimodal-representations}, 
	in \twotower{} VLM architecture, the performance of \mtname{} first increases gradually with the number of unimodal representations, but then stops increasing and even decreases when the number of unimodal representations exceeds $6$.
	Similar trend is also observed in the MLLM architecture 
	in~\Fref{fig:vision-selection}.
	How to obtain better performance with acceptable computational budget by utilizing more/better insights of unimodal experts, especially when scaling the model or the MLLM architecture with deeper and wider modules, \eg{} $24$-layer CLIP-ViT L-224/16 and $24$-layer LLM Qwen2-0.5B-Instruct, is a question worth further exploration.
	For example, designing reasonable sparse activation functions for managers, instead of manually selection, simple top-$\mathrm{N}$ or top-$\mathrm{p}$ sampling (which did not work well in our preliminary experiments).
	\item \textbf{Manager Injection}: In the \twotower{} VLM architecture, we inject managers into each cross-modal layer.
	In the MLLM architecture, we uniformly inject managers into the LLM from the first layer at a fixed layer interval.
	How to inject managers into the LLM in a more flexible and adaptive way is also a question worth further exploration.
	For example, non-uniform injection and do not start from the first layer.
\end{itemize}

In addition, it would also be interesting to explore the effectiveness of managers in other VLM architectures~\cite{alayrac2022flamingo,wang2023cogvlm}, multimodal in-context learning~\cite{zhao2024mmicl,qin2024factors} and multimodal chain-of-thought reasoning~\cite{zhang2024multimodal,chen-etal-2024-m3cot}, and the collaboration with multi-grained multimodal data~\cite{zeng2021multi,xu2024exploring}.

There are also some recent works that follow or share the spirit of our manager to aggregate multi-layer unimodal representations in VLMs and MLLMs~\cite{dou2021meter,xu2023bridgetower,dou2022fiber,yao2024dense,li2024tokenpacker}.
Besides, some works~\cite{lin2023sphinx,shi2024eagle,tong2024cambrian} explore the utilization of different visual encoders (with different resolutions), \eg{}, DINOv2~\cite{oquab2024dinov} and SigLIP~\cite{zhai2023sigmoid}, to improve the visual representation.
They aggregate multiple visual representations from different visual encoders (experts), which is similar to our manager that aggregates multi-layer visual representations from different layers of the visual encoder (we treat each layer of the visual encoder as an expert).

\begin{figure*}[!ht]
    \centering
    \includegraphics[width=0.98\textwidth]{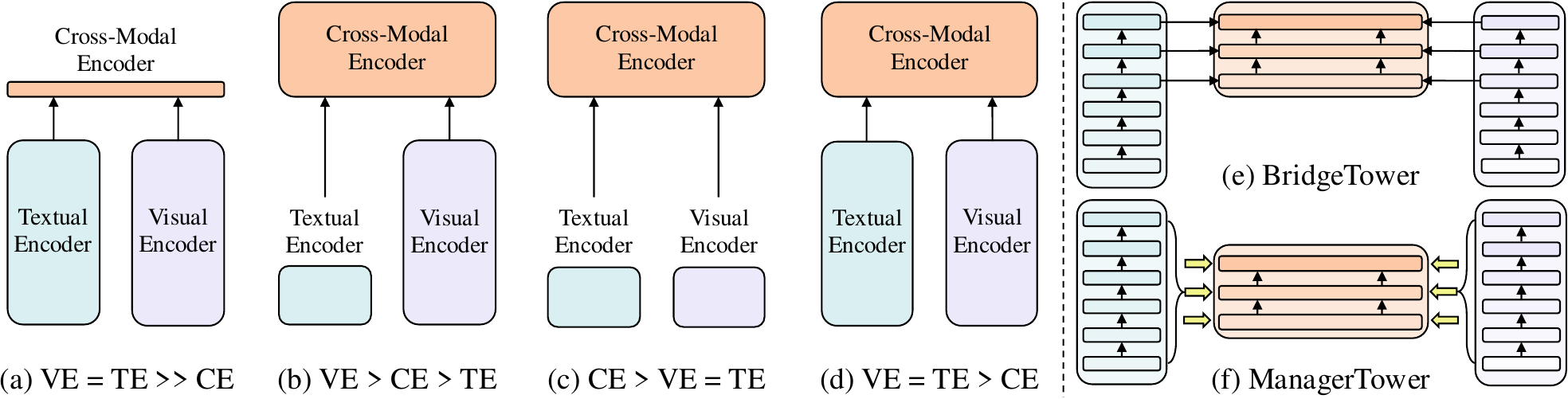}
    \caption{
        (a) -- (d) are four main categories of the \twotower{} VLM;~(e)\&(f) gives a brief illustration of \btname{} and \mtname{}. 
        VE, TE, and CE are short for the Visual Encoder, Textual Encoder, and Cross-modal Encoder, respectively.
        The height of each rectangle represents its relative computational cost. 
        $\text{VE}=\text{TE}$ indicates that the visual encoder and the textual encoder have the same or a similar number of parameters or computational costs. 
		Hollow arrows indicate the transmission of multi-layer visual representations aggregated by managers to the LLM at intervals.
        Illustration inspired by ViLT.
    }
    \label{fig:rebuttal_related_work_framework}
\end{figure*}

\subsection{Semantic Ambiguity in the Multi-Grid Algorithm}
In~\Sref{sec:detailed-analysis-mllm}, we provide a detailed analysis and case study of the semantic ambiguity caused by the multi-grid algorithm in the MLLM architecture.
In fact, the multi-grid algorithm is a resource-efficient way to directly split the image into multiple image grids, which can effectively improve the performance of MLLMs in high-resolution scenarios.
It is widely used in both academic research~\cite{lin2023sphinx,li2024monkey,liu2024llavanext,shi2025we} and industrial applications~\cite{dai2024nvlm,chen2024expanding,wu2024deepseekvl2}.
It also called dynamic image tiling or image cropping in some works.

However, the multi-grid algorithm may bring semantic ambiguity since the grid line may cut off objects and connected regions in the image~\cite{liu2024textmonkey,huang2024hires,huang2024mini}.
Current works try to mitigate the semantic ambiguity by introducing new modules to enhance the interaction between different image grids~\cite{liu2024llavanext,huang2024hires}, or by simultaneously adopting two different grid partitioning schemes~\cite{huang2024mini}.
They all bring additional training and computational costs.
Furthermore, our manager provides an orthogonal perspective to alleviate the semantic ambiguity by incorporating different levels of unimodal semantic knowledge contained in the multi-layer unimodal representations from the visual encoder, which can capture and supplement more diverse visual details.
The diversity of attention weights and attention heads in the multi-grid MLLM will be increased by our manager, which also demonstrates the guidance and collaboration between the manager and the multi-grid algorithm.

In fact, no matter the image patch in the ViTs or the feature map from a convolutional kernel in the CNNs, they all suffer from the semantic ambiguity caused by partitioning the image into small regions.
The self-attention mechanism in the ViTs and the convolutional operation in the CNNs can help to alleviate the semantic ambiguity by interacting with different regions of the image.
Directly training a high-resolution visual encoder is a more straightforward and effective way to improve the supported image resolution of the visual encoder, but need precious and scarce high-resolution training data for continual training of the visual encoder~\cite{fuyu-8b,li2024mini,wang2024qwen2,liu2024oryx}.
From the perspective of improving the supported image resolution, the multi-grid algorithm can be seen as an ensemble of the visual encoder but with different image grids.
The separate encodings of different image grids bring more diverse visual details, but also further intensify the semantic ambiguity.
Appending the representation of the resized base image before the image grids alleviates the semantic ambiguity, but not enough based on our experiments.
The collaboration between our manager (deeper representation) and the multi-grid algorithm (wider representation) with different interaction modules for different image grids is a promising direction to further improve the performance of multi-grid MLLMs.

\subsection{Multimodal Fusion from the Perspective of Architecture}
\label{appendix:related-work-multimodal-fusion}

Following the taxonomy proposed by ViLT~\cite{kim2021vilt}, most VLMs can be unified into the \twotower{} architecture shown in \Fref{fig:rebuttal_related_work_framework}~(a) -- (d). 
They feed last-layer representations of pre-trained unimodal encoders into the top cross-modal encoder and can be differentiated by the depth of the textual, visual, and cross-modal encoders\footnote{
    A cross-modal decoder can be placed on top of the cross-modal encoder or directly replace the cross-modal encoder, \eg{} the latest MLLMs.
}.

CLIP~\cite{radford2021learning} and ALIGN~\cite{jia2021scaling} are representative models that directly perform a shallow fusion (\eg{} dot product) of last-layer representations of equally expressive pre-trained unimodal encoders in the cross-modal encoder, as illustrated in \Fref{fig:rebuttal_related_work_framework}~(a).
The remaining models perform deep fusion in the multi-layer transformer-based cross-modal encoder but choose pre-trained unimodal encoders with varying levels of expressiveness. 
Numerous works~\cite{li2019visualbert,su2019vl,li2020unicoder,chen2020uniter,li2020oscar,zhou2020unified,zhang2021vinvl,pmlr-v139-cho21a,huang2020pixel,huang2021seeing,shen2021much,liu2021kd,li2021unimo,xia2021xgpt,ni2021m3p,chen2022pali,wang2022git,alayrac2022flamingo}
fall in the category of \Fref{fig:rebuttal_related_work_framework}~(b) as they adopt various types of deep vision models (\eg{} Faster R-CNN~\cite{ren2015faster}, ResNet~\cite{he2016deep} or ViT~\cite{dosovitskiy2020image}) as their visual encoder to obtain region, grid, or patch features, and concatenate them with word embedding to feed into their top cross-modal encoder.
The third category of models~\cite{kim2021vilt, wang2021simvlm, wang2021distilled, wang2022OFA}, illustrated in \Fref{fig:rebuttal_related_work_framework}~(c), utilizes lightweight visual and lightweight textual encoders and handles both modalities in a single transformer-based cross-modal encoder. 
In contrast, models~\cite{lu2019vilbert, tan2019lxmert, kamath2021mdetr, li2021align, zeng2021multi, dou2021meter, nagrani2021attention, wang2021vlmo, li2022blip, li2022unimo, wang2022image, yu2022coca, li2022mplug}, which belong to \Fref{fig:rebuttal_related_work_framework}~(d) category, use expressive deep pre-trained unimodal encoders and feed their last-layer representation into the top multi-layer cross-modal encoder.

Regardless of the visual, textual, or cross-modal encoders they utilize, 
from the perspective of architecture, most current models ignore the various levels of semantic information at the different layers of pre-trained unimodal encoders, and simply utilize the last-layer unimodal representations for multimodal fusion.
While the models belonging to \Fref{fig:rebuttal_related_work_framework}~(c) appear to retain the possibility of utilizing different levels of unimodal semantic information, it could be challenging for them to learn intra- and cross-modal interactions concurrently without modality-specific parameters. Their unconstrained cross-modal interaction could impede intra-modal interaction~\cite{dou2021meter, du2022survey}.

Unlike current models, \btname{} and \mtname{}, as shown in \Fref{fig:rebuttal_related_work_framework}~(e)\&(f), proposes to incorporate bridge layers or managers into the cross-modal encoder.
This allows the model to aggregate the insights from the top layers of unimodal encoders to improve multimodal fusion.
This does not affect intra-modal interaction in the pre-trained unimodal encoders, and enables different semantic levels of visual and textual representations to interact adaptively and thoroughly at each layer of the cross-modal encoder.

Furthermore, in short, from the perspective of mechanism, \cite{hendricks2021decoupling} perform analysis on different types of attention mechanisms used in \twotower{} VLMs and demonstrate that the \coattention{} mechanism~\cite{lu2019vilbert} performs best. 
This mechanism uses a different set of parameters for each modality. 
For example, for the visual part of the cross-modal encoder, the queries of each multi-head cross-attention (MCA) block are from the visual modality, but the keys and values are from the other modality (\ie{} the textual modality).
However, as LLMs become more powerful, the \mergedattention{} mechanism has gradually become the de-facto standard for MLLMs, which simply concatenate the visual and textual input tokens, and consider keys and values from both modalities.
This mechanism can maximize the potential of LLMs by mapping the representations of the visual encoder to the representation space of LLMs, and then directly relying on the multi-head self-attention (MSA) block for multimodal fusion.

\subsection{Appropriate Parameter Selection of Manager}
As a lightweight plugin, the reasonable introduction of managers can bring significant performance improvement to \twotower{} VLMs and MLLMs with acceptable computational cost.
Take MLLMs as an instance, the parameter fine-tuning of managers mainly focuses on the selection of visual representations and the injection times of managers.
The purpose of parameter fine-tuning on managers, or we can say the ablation study of managers, is to achieve better performance while controlling the computational cost.

\subsubsection{Visual Representation Selection}
\label{appendix:visual-representation-selection}

The first question in parameter selection is what kind of visual representation should the manager manage.
As shown in~\Fref{fig:vision-selection}, overall, no matter what visual representations are selected, managers \textbf{consistently} improve the performance of \basename{}.
Similar to the observations in both \btname{} and \mtname{}, visual representations from the top half of the visual encoder bring the best performance, and using visual representations from all layers leads to the lowest performance improvement.
We attribute this to the fact that the average attention distance of the visual encoder increases with the layer depth, especially in the top half of the visual encoder, where most attention heads attend \textbf{widely} across tokens~\cite{dosovitskiy2020image} and capture global visual features.

Hence, we recommend selecting visual representations from the top half of the visual encoder for optimal performance.
This can not only reduce the training time and GPU memory usage, but also improve the performance of managers.

\subsubsection{Manager Injection Times}
\label{appendix:manager-injection-times}

The second question in parameter selection is how many times should managers be injected into the LLM.
In this work, we uniformly inject managers into the LLM from the first layer at a fixed layer interval.
Specifically, for the LLM with $L_\mathrm{C}\!=\!24$, we can inject $6$ managers with the interval of $4$, or inject $24$ manager with the interval of $1$, \ie{} inject managers into each LLM layer.

As shown in~\Fref{fig:manager-injection}, the injection times of managers will affect the performance, and the overall trend is that performance improves with increasing injection frequency, but with some fluctuations.
\basemname{} can achieve \textbf{better} performance than \basename{} most of the time.
Compared to the injection times of $6$, although injecting managers into each LLM layer slightly increases the average performance from $50.96\%$ to $51.08\%$, it also \textbf{increases} the computational cost by about $7\%$ in both training and inference.

Hence, we recommend injecting managers into the LLM $6$ times with the interval of $4$.
This can not only achieve nearly the best performance, but also reduce the computational cost in both training and inference, which is more efficient than injecting managers into each LLM layer.

\subsubsection{Potential Limitations}
\label{appendix:potential-limitations}

As we discussed in~\Apref{appendix:spirit-of-manager}, it would be interesting to explore more flexible and adaptive ways to select which layers of unimodal representations to aggregate, and how to inject them into different layers of the cross-modal encoder or the LLM in the future.
\begin{itemize}
	\item \textbf{Visual Representation Selection}: 
	As shown in~\Fref{fig:number-of-unimodal-representations}, in \twotower{} VLM architecture, the performance of \mtname{} first increases gradually with the number of unimodal representations, but then stops increasing and even decreases when the number of unimodal representations exceeds $6$.
	Similar trend is also observed in the MLLM architecture in~\Fref{fig:vision-selection}.
	How to obtain better performance with acceptable computational budget by utilizing more/better insights of unimodal experts, especially when \textbf{scaling} the model or the MLLM architecture with \textbf{deeper} and \textbf{wider} modules, \eg{} $40$-layer OpenCLIP-ViT G-224/14 and $80$-layer LLM Qwen2.5-72B-Instruct, is a question worth further exploration.
	The search space of visual representation selection is larger and the search cost is significantly increased.
    For example, designing reasonable sparse activation functions for managers, instead of manually selection, simple top-$\mathrm{N}$ or top-$\mathrm{p}$ sampling (which did not work well in our preliminary experiments).
	\item \textbf{Manager Injection}: 
    In the \twotower{} VLM architecture, we inject managers into each cross-modal layer.
	In the MLLM architecture, we uniformly inject managers into the LLM from the first layer at a fixed layer interval.
	How to inject managers into the LLM in a more flexible and adaptive way is also a question worth further exploration.
    This not only affects the performance, but also the computational cost.
	For example, non-uniform injection and do not start from the first layer.
\end{itemize}

\subsection{Comparison with MoE and Deep Layer Aggregation}
\label{sec:moe-comparison}

From a conceptual perspective, our proposed manager is more like a method that combines the spirit of mixture-of-experts (MoE) and conducts deep unimodal expert aggregation in the multimodal representation learning of \twotower{} VLMs and MLLMs.

\begin{itemize}[parsep=2pt,itemsep=2pt,topsep=2pt]
    \item \textbf{Comparison with MoE}: 
    Standard MoEs, \eg{} Switch Transformer~\cite{fedus2022switch}, typically take each Feed-Forward Network (FFN) layer as an expert. 
    They adopt a gating network, \eg{} top-k router, to sparsely select a subset of experts for each input sample or token.
    In contrast, our proposed manager takes each layer of the unimodal encoder as an expert, and design a softmax-activated gating network to aggregate the outputs of different levels of experts.
    Instead of sparsely activated experts by routers, our managers can directly reuse the pre-trained unimodal encoders as experts, and aggregate their outputs.
    Furthermore, as shown in~\Sref{sec:visualization} and \Fref{fig:aggregation-weights-pre-trained-aaum} in our manuscript, the aggregation weight distributions from managers are \textbf{completely different} from the one-hot distributions manually specified in \btname{} or some sparse distributions in MoE.
    Our manager learns \textbf{diverse} distributions in different cross-modal layers.
    \item \textbf{Comparison with Deep Layer Aggregation}:
    Deep layer aggregation, \eg{} the linear combination of layers method~\cite{wang-etal-2019-learning-deep}, is a simple yet effective way that aggregates the representations of previous layers using learned weights in each encoder layer.
    They adopt different methods to aggregate the outputs of previous layers, to improve the performance of deep networks.
    As we discussed in~\Sref{sec:manager-design}, the direct adaptation of deep layer aggregation methods to the \twotower{} VLMs does not work well.
    Our analysis shows it may bring redundant representation when aggregating the outputs of both all unimodal layers and previous cross-modal layers.
    In contrast, our proposed manager is designed to adaptively aggregate the outputs of different levels of unimodal experts based on \textbf{only} the representation from the previous cross-modal layer.
    This can not only reduce the computational cost, but also improve the performance of managers.
    Our manager is an effective and efficient adaptation of deep layer aggregation methods to the \twotower{} VLMs and MLLMs.
\end{itemize}

\section{Exploration on \twotower{} VLM}

\subsection{Simplified Equations}
\label{appendix:managers-equations}

\begin{figure}[!h]
	\centering
	\includegraphics[width=0.4\textwidth]{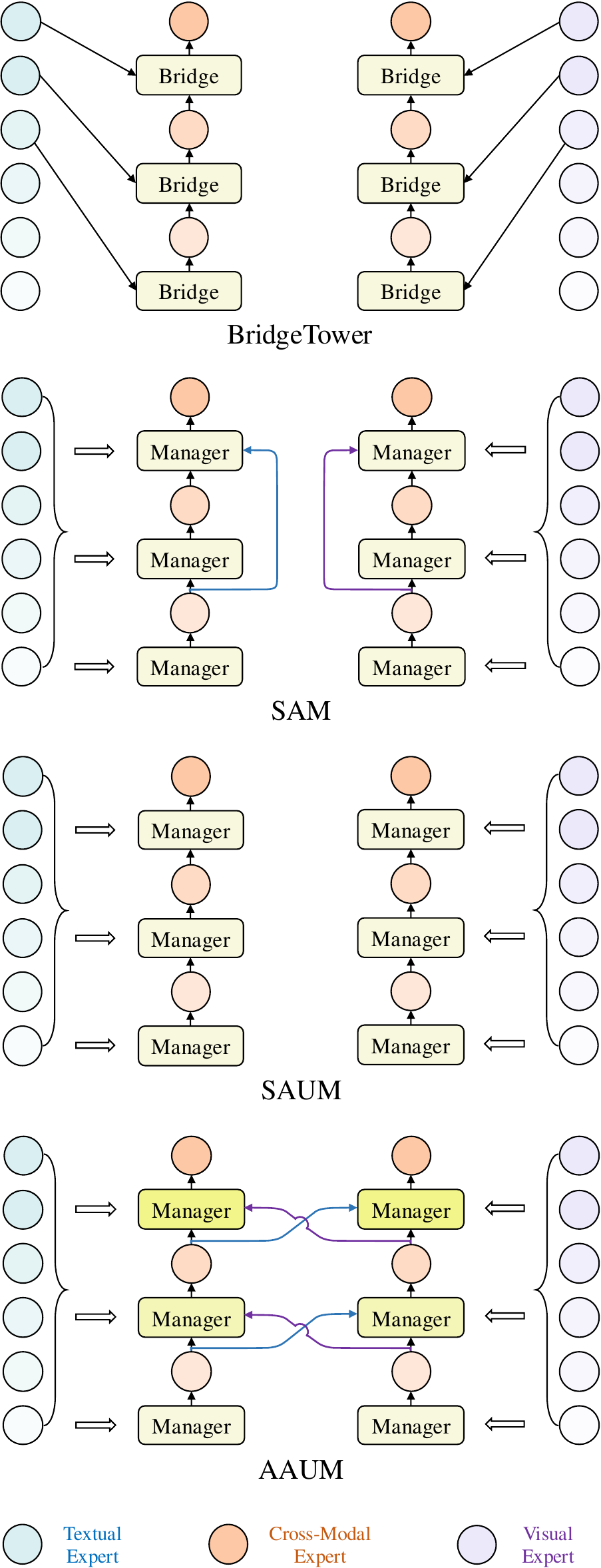}
	\caption{
		Brief illustrations of \btname{} and our \mtname{} with SAM, SAUM and AAUMs. 
		Hollow arrows indicate the transmission of multi-layer unimodal representations in \mtname{} instead of layer-by-layer transmission in \btname{}. 
		Each unimodal or cross-modal layer is seen as an unimodal or cross-modal expert.
		The arrow between the cross-modal expert of the previous layer and the manager of the current layer is to get the cross-modal fused query.
	}
	\label{fig:managers}
\end{figure}

We first provide simplified equations in~\Eref{eq:explain_bt}, \eqref{eq:explain_mt_sam}, \eqref{eq:explain_mt_saum}~\&~\eqref{eq:explain_mt_aaum} to better explain their distinctions and integration.
Take the left side of~\Fref{fig:managers} as an example, assume that we have a 12-layer textual encoder and the output representations of its top 6 layers will be the unimodal part of input for a manager.

For BridgeTower, only the output representations of a specified layer in the textual encoder will be used as the unimodal part of input for a bridge.

\begin{equation}
    \label{eq:explain_bt}
    \text{BridgeTower} 
    \left\{
        \begin{aligned}
            & \text{UniModal Part:} & L \times D \\
            & \text{Cross-Modal Part:} & L \times D
        \end{aligned}
    \right.
    \xrightarrow{\text{Bridge}} 
    L \times D
\end{equation}

\begin{equation}
    \label{eq:explain_mt_sam}
    \text{SAM} 
    \left\{
        \begin{aligned}
            & \text{UniModal Part:} & 6 \times L \times D \\
            & \text{Cross-Modal Part:} & (\ell - 1) \times L \times D \\
            & \text{Learned Weights:} & (6 + \ell - 1) \times D
        \end{aligned}
    \right.
    \xrightarrow{\text{Manager}} 
    L \times D
\end{equation}

\begin{equation}
    \label{eq:explain_mt_saum}
    \text{SAUM} 
    \left\{
        \begin{aligned}
            & \text{UniModal Part:} & 6 \times L \times D \\
            & \text{Cross-Modal Part:} & L \times D \\
            & \text{Learned Weights:} & 7 \times D
        \end{aligned}
    \right.
    \xrightarrow{\text{Manager}} 
    L \times D
\end{equation}

\begin{equation}
    \label{eq:explain_mt_aaum}
    \text{AAUM} 
    \left\{
        \begin{aligned}
            & \text{UniModal Part:} & 6 \times L \times D \\
            & \text{Cross-Modal Part:} & L \times D \\
            & \text{Generated Weights:} & 7 \times L
        \end{aligned}
    \right.
    \xrightarrow{\text{Manager}} 
    L \times D
\end{equation}

For different type of managers, they will accept the output representations of all 6 layers, and then learn or generate aggregation weights for both unimodal and cross-modal part of input.
Except SAM, other modules only take the output representations from the previous cross-modal layer.
In SAM, $\ell \in [1,2,3,4,5,6]$ is the layer index of the cross-modal encoder.
SAM will accept the output representations from all previous cross-modal layers.

\begin{figure*}[!h]
	\centering
	\includegraphics[width=\textwidth]{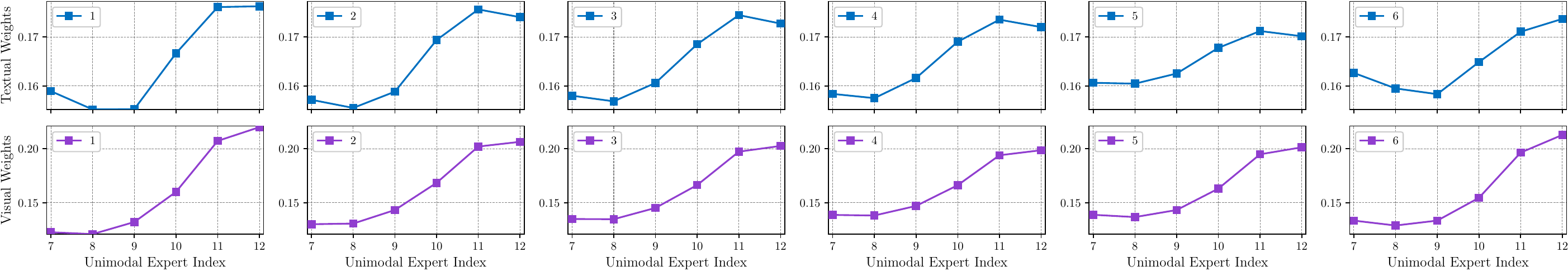}
	\caption{
		A visualization of aggregation weights of textual and visual SAMs in each cross-modal layer.
		The X-axis is the index of the unimodal expert, and the legend shows the index of the cross-modal layer.\looseness=-1
	}
	\label{fig:aggregation-weights-sam}
\end{figure*}

\begin{figure*}[!h]
	\centering
	\includegraphics[width=\textwidth]{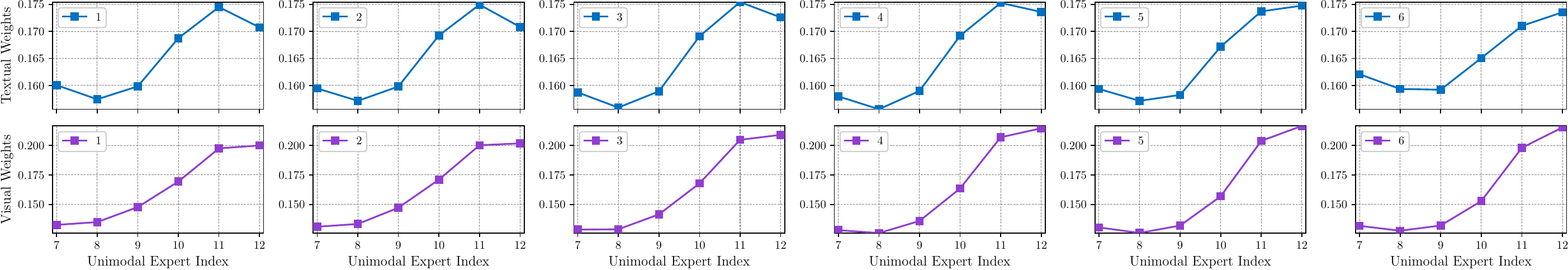}
	\caption{
		A visualization of aggregation weights of textual and visual SAUMs in each cross-modal layer.
		The X-axis is the index of the unimodal expert, and the legend shows the index of the cross-modal layer.\looseness=-1
	}
	\label{fig:aggregation-weights-saum}
\end{figure*}

\begin{figure*}[!h]
	\centering
	\includegraphics[width=\textwidth]{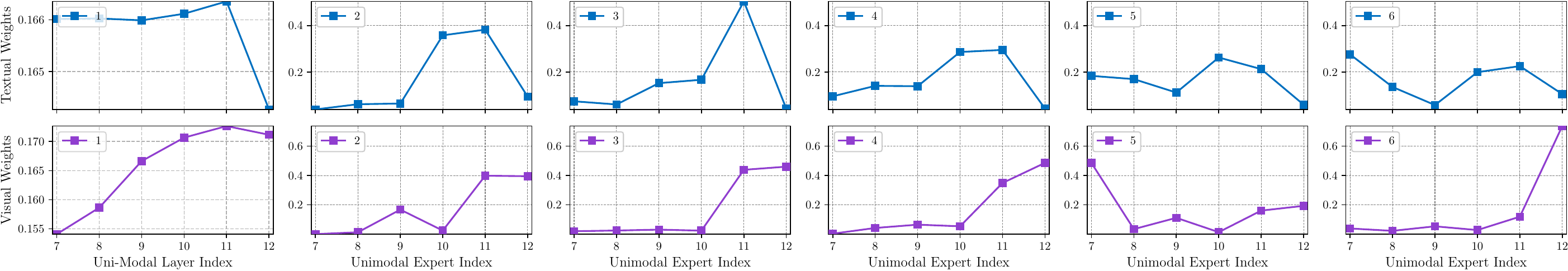}
	\caption{
		A visualization of aggregation weights of textual and visual AAUMs in each cross-modal layer.
		The X-axis is the index of the unimodal expert, and the legend shows the index of the cross-modal layer.\looseness=-1
	}
	\label{fig:aggregation-weights-aaum}
\end{figure*}

\subsection{Intuitive Comparison Between BT\&MT}
\label{appendix:managers}
We intuitively compare \btname{} (BT) and \mtname{} (MT) with different type of managers in~\fref{fig:managers}.

\subsubsection{BT~\vs{}~MT with SAUM}
In~\Tref{tab:number-of-cross-modal-layers}~\&~\ref{tab:different_backbones}, 
we provide the detailed performance comparison between \btname{} and \mtname{}.\footnote{
	The re-implemented BridgeTower obtained higher experimental results than the original paper due to the better fine-tuning settings we used for all experiments 
	in~\Sref{sec:investigation-and-analysis}.
}
In fact, \btname{} can be seen as an approximate special case of \mtname{} with SAUMs if we replace the learned weights $\mathbf{W}$ in each manager with layer-by-layer one-hot distributions\footnote{
	For each cross-modal layer, only one unimodal expert is activated at a time in the bottom-up direction.
} used in \btname{}.
However, as shown in~\fref{fig:aggregation-weights-saum}, the aggregation weight of textual and visual SAUMs share a similar progressive trend across cross-modal layers, which is completely different from the one-hot distributions in \btname{}.
This allows \mtname{} with SAUMs to achieve significant performance gains compared to \btname{} (\eg{}~$75.91\%$ \vs{} $76.55\%$ on \vqa{} test-dev in~\Tref{tab:type-and-query}).
Besides, the similar trend of aggregation weight distributions between different managers is consistent with the observations 
in~\fref{fig:cosine_similarity_cross_uni}, 
that is, the cosine similarity of aggregated unimodal representations between managers is always similar to $1$.

\subsubsection{SAM~\vs{}~SAUM~\vs{}~AAUM}
We provide the visualizations of aggregation weights of SAM, SAUM and AAUM without VLP in~\fref{fig:aggregation-weights-sam}~\&~\ref{fig:aggregation-weights-saum}~\&~\ref{fig:aggregation-weights-aaum}.
Comparing the visualization of three types of managers without VLP, we can find that: 
($i$) the learned aggregation weights of SAM and SAUM share a similar progressive trend across cross-modal layers;
($ii$) for each AAUM, its generated aggregation weights vary significantly across $6$ unimodal experts. Comparing different AAUMs, the aggregation weight distributions generated by AAUMs is also very different.
($iii$) when we compare~\fref{fig:aggregation-weights-saum}~\&~\ref{fig:aggregation-weights-aaum}, their respective aggregation weight distributions are completely different.
This further demonstrates that compared with SAUMs, AAUMs can adaptively \textbf{generates} different aggregation weights for different tokens in different samples.
Interestingly, the first column of two figures both comes from SAUMs, but the distributions are still \textbf{clearly} different.
We presume that high-layer AAUMs may help low-layer SAUMs \textbf{rectify} their management of experts.

\begin{table*}[t]
	\tablestyle{5pt}{1.1}
	\caption{
		Performance of \btname{} and \mtname{} with different visual and textual backbones.
		B, N and M in ``ViT B-N/M'' denote the model size, image resolution and patch size, respectively.
	}
	\label{tab:different_backbones}
	\adjustbox{width=0.85\linewidth}{
		\begin{tabular}{rc|cc|cc}
			{Visual}    & {Textual} & \multicolumn{2}{c|}{VQAv2 Test-Dev (\%)} & \multicolumn{2}{c}{Flickr30K \rmean{} (\%)}                                            \\
			Backbone    & Backbone  & \btname{}                                  & \multicolumn{1}{c|}{\mtname{}}              & \btname{}     & \multicolumn{1}{c}{\mtname{}}         \\
			\shline
			DeiT B-224/16 & RoBERTa   & 71.22                               & 72.20 ($\uparrow$\,0.98)               & 87.63 & 88.72($\uparrow$\,1.09)          \\
			ViT B-224/16  & RoBERTa   & 72.82                               & 73.67 ($\uparrow$\,0.85)               & 90.48 & 90.92($\uparrow$\,0.44)          \\
			ViT B-384/16  & RoBERTa   & 72.94                               & 73.80 ($\uparrow$\,0.86)               & 90.51 & 90.96($\uparrow$\,0.45)          \\
			CLIP-ViT B-224/32 & RoBERTa   & 73.73                               & 74.79 ($\uparrow$\,1.06)               & 91.33 & 91.76($\uparrow$\,0.43)          \\
			CLIP-ViT B-224/16 & BERT      & 75.74                               & 76.36 ($\uparrow$\,0.62)               & 92.84 & 93.42($\uparrow$\,0.58)          \\
			CLIP-ViT B-224/16 & RoBERTa   & 75.91                               & \textbf{76.65} ($\uparrow$\,0.74)      & 93.33 & \textbf{93.97}($\uparrow$\,0.64) \\
		\end{tabular}
	}
\end{table*}
\begin{table*}[t]
	\tablestyle{5pt}{1.1}
	\caption{
		Computational budget and downstream task performance without VLP for \btname{} and \mtname{}. 
		* denotes our re-implementation.
	}
	\label{tab:computational_budget}
	\adjustbox{width=\linewidth}{
		\begin{tabular}{l|cc|c|l|l|l|l}
			\multirow{2}{*}{Model}  & {Manager}                & {Manager}                                                        & {\#~Params}             & \multicolumn{1}{c|}{\#~FLOPs}              & {Inference Time}            & \multicolumn{1}{c|}{VQAv2}    & \multicolumn{1}{c}{Flickr30K} \\
			                        & \multicolumn{1}{c}{Type} & \multicolumn{1}{c|}{Visual Query}                                & \multicolumn{1}{c|}{(M)} & \multicolumn{1}{c|}{(G)} & \multicolumn{1}{c|}{(ms/sample)} & \multicolumn{1}{c|}{Test-Dev (\%)} & \multicolumn{1}{c}{\rmean{} (\%)}  \\
			\shline
			\btname{}\modelbase{} * & -                        & -                                                                & 326.58                  & 101.25                  &       39.43       & 75.91                         & 93.33                         \\ 
			\hdashline
			\mtname{}\modelbase{}   & SAUM                     & -                                                                & 326.77                  & 101.34 ($\times$1.00)                  & 41.12 ($\times$1.04)              & 76.55 ($\uparrow$\,0.64)      & 93.73 ($\uparrow$\,0.40)      \\ 
			\mtname{}\modelbase{}   & AAUM                     & $\mathbf{C}^\mathrm{V}_{\ell-1}$                                 & 326.77                  & 101.35 ($\times$1.00)                  & 41.80 ($\times$1.06)              & 76.52 ($\uparrow$\,0.61)      & 93.84 ($\uparrow$\,0.51)      \\ 
			\mtname{}\modelbase{}   & AAUM                     & $\mathbf{C}^\mathrm{V}_{\ell-1}, \mathbf{C}^\mathrm{T}_{\ell-1}$ & 338.64                  & 105.52 ($\times$1.04)                  & 43.20 ($\times$1.10)             & \textbf{76.65} ($\uparrow$\,0.74)      & \textbf{93.97} ($\uparrow$\,0.64)      \\ 
		\end{tabular}
	}
\end{table*}

\subsection{Switch Visual and Textual Backbones}

We experiment with different pre-trained visual and textual backbones as unimodal encoders to further investigate the impact on performance of the managers of \mtname{} compared to the bridges of \btname{}.
As shown in~\Tref{tab:different_backbones}, regardless of the visual and textual backbones we apply, \mtname{} significantly and consistently outperforms \btname{} on both datasets.
This further proves the effectiveness and generalization of our proposed \mtname{} architecture and managers, which can provide adaptive and effective aggregation of multi-layer unimodal representations for VL representation learning.

\begin{table*}[t]
	\tablestyle{5pt}{1.1} 
	\caption{
		Comparisons with previous models on $4$ downstream datasets.
		The best score is bolded. 
		B, N and M in ``ViT B-N/M'' denote the model size, image resolution and patch size, respectively.
		$\ast$ indicates that the model also uses VG-QA data to fine-tune on VQAv2.
		$\star$ denotes the model is trained from scratch. 
		``\# Pre-train Images'' denotes the number of unique images used in VLP.
	}
	\label{tab:detailed-main-results}
	\adjustbox{width=\linewidth}{
		\begin{tabular}{lrc|cc|cc|cc|cc}
			\multirow{2}{*}{Model}                              & {\#~Pre-train} & {Visual}          & \multicolumn{2}{c|}{VQAv2 (\%)} & \multicolumn{2}{c|}{SNLI-VE (\%)} & \multicolumn{2}{c|}{NLVR$^2$ (\%)} & \multicolumn{2}{c}{Flickr30K (\%)}                                                     \\
			                                                    & Images~        & Backbone          & Test-Dev                   & Test-Std                     & Dev                           & Test                          & Dev        & Test-P     & IR@1       & TR@1       \\
			\shline
			\multicolumn{11}{l}{ { \it{Base-size models pre-trained on 4M public data} } }                                                                                                                                                                                           \\
			\hline
			ViLT\modelbase{}~\cite{kim2021vilt}                & 4M             & ViT B-384/32      & 71.26                      & -                            & -                             & -                             & 75.70      & 76.13      & 64.4       & 83.5       \\
			UNITER\modelbase{}~\cite{chen2020uniter}\,$\ast$   & 4M             & Faster R-CNN      & 72.70                      & 72.91                        & 78.59                         & 78.28                         & 77.18      & 77.85      & 72.52      & 85.90      \\
			VILLA\modelbase{}~\cite{gan2020large}\,$\ast$      & 4M             & Faster R-CNN      & 73.59                      & 73.67                        & 79.47                         & 79.03                         & 78.39      & 79.30      & 74.74      & 86.60      \\
			UNIMO\modelbase{}~\cite{li2021unimo}               & 4M             & Faster R-CNN      & 73.79                      & 74.02                        & 80.00                         & 79.10                         & -          & -          & 74.66      & 89.70      \\
			ALBEF\modelbase{}~\cite{li2021align}\,$\ast$       & 4M             & DeiT B-224/16     & 74.54                      & 74.70                        & 80.14                         & 80.30                         & 80.24      & 80.50      & 82.80       & 94.30       \\
			VinVL\modelbase{}~\cite{zhang2021vinvl}            & 5.7M           & ResNeXt-152       & 75.95                      & 76.12                        & -                             & -                             & 82.05      & 83.08      & -          & -          \\
			\metername{}-Swin\modelbase{}~\cite{dou2021meter}  & 4M             & Swin B-384/32     & 76.43                      & 76.42                        & 80.61                         & 80.45                         & 82.23      & 82.47      & 79.02      & 92.40      \\
			\vlmoname{}\modelbase{}~\cite{wang2021vlmo}        & 4M             & BEiT B-224/16     & 76.64                      & 76.89                        & -                             & -                             & 82.77      & 83.34      & 79.30       & 92.30       \\
			\metername{}-CLIP\modelbase{}~\cite{dou2021meter}  & 4M             & CLIP-ViT B-224/16 & 77.68                      & 77.64                        & 80.86                         & 81.19                         & 82.33      & 83.05      & 82.22      & 94.30      \\
			\btname{}\modelbase{}~\cite{xu2023bridgetower}          & 4M             & CLIP-ViT B-224/16 & 78.66                      & 78.73                        & 81.11                         & 81.19                         & 81.85      & 83.09      & 85.83      & 94.73      \\
			\mtname{}\modelbase{}~(\bf{Ours})                   & 4M             & CLIP-ViT B-224/16 & \bf{79.39}                 & \bf{79.15}                   & \bf{81.26}                    & \bf{81.44}                    & \bf{82.81} & \bf{83.34} & \bf{86.56} & \bf{95.64} \\
			\hline
			\multicolumn{11}{l}{ { \it{Models pre-trained on more data and/or with larger size}}}                                                                                                                                                                                    \\
			\hline
			UNITER\modellarge{}~\cite{chen2020uniter}\,$\ast$  & 4M             & Faster R-CNN      & 73.82                      & 74.02                        & 79.39                         & 79.38                         & 79.12      & 79.98      & 75.56      & 87.30      \\
			VILLA\modellarge{}~\cite{gan2020large}\,$\ast$     & 4M             & Faster R-CNN      & 74.69                      & 74.87                        & 80.18                         & 80.02                         & 79.76      & 81.47      & 76.26      & 87.90      \\
			UNIMO\modellarge{}~\cite{li2021unimo}              & 4M             & Faster R-CNN      & 75.06                      & 75.27                        & 81.11                         & 80.63                         & -          & -          & 78.04      & 89.40      \\
			ALBEF\modelbase{}~\cite{li2021align}\,$\ast$       & 14M            & DeiT B-224/16     & 75.84                      & 76.04                        & 80.80                         & 80.91                         & 82.55      & 83.14      & 85.60       & 95.90       \\
			VinVL\modellarge{}~\cite{zhang2021vinvl}           & 5.7M           & ResNeXt-152       & 76.52                      & 76.63                        & -                             & -                             & 82.67      & 83.98      & -          & -          \\
			BLIP\modelbase{}~\cite{li2022blip}\,$\ast$         & 14M            & DeiT B-224/16     & 77.54                      & 77.62                        & -                             & -                             & 82.67      & 82.30      & 87.20       & 96.60       \\
			SimVLM\modelbase{}~\cite{wang2021simvlm}\,$\star$  & 1.8B           & ResNet-101        & 77.87                      & 78.14                        & 84.20                         & 84.15                         & 81.72      & 81.77      & -          & -          \\
			BLIP\modelbase{}~\cite{li2022blip}\,$\ast$         & 129M           & DeiT B-224/16     & 78.24                      & 78.17                        & -                             & -                             & 82.48      & 83.08      & 87.30       & 97.30       \\
			SimVLM\modellarge{}~\cite{wang2021simvlm}\,$\star$ & 1.8B           & ResNet-152        & 79.32                      & 79.56                        & 85.68                         & 85.62                         & 84.13      & 84.84      & -          & -          \\
			\vlmoname{}\modellarge{}~\cite{wang2021vlmo}       & 4M             & BEiT L-224/16     & 79.94                      & 79.98                        & -                             & -                             & 85.64      & 86.86      & 84.50       & 95.30       \\
			SimVLM\modelhuge{}~\cite{wang2021simvlm}\,$\star$  & 1.8B           & Larger ResNet-152 & 80.03                      & 80.34                        & 86.21                         & 86.32                         & 84.53      & 85.15      & -          & -
		\end{tabular}
	}
\end{table*}

\subsection{Computational Budget}
\label{appendix:computational_budget}

Table~\ref{tab:computational_budget} shows the computational budget and downstream task performance without VLP for \btname{} and \mtname{}, including the number of parameters, the number of FLoating-Point operations (FLOPs),\footnote{
	We use Facebook Research's \href{https://github.com/facebookresearch/fvcore}{fvcore}, to calculate FLOPs.
} and the average inference time per instance.

We measure the average inference time of processing $1$ VQA instance over $10$K runs on $1$ NVIDIA TITAN V GPU. The sequence length is $50$, and the image resolution is $384 \times 384$.
Compared with \btname{} ($1^{\text{st}}$ row), \mtname{} ($4^{\text{th}}$ row) uses an acceptable additional computational budget 
($3.7\%$ parameters, $4.2\%$ FLOPs, and $3.8$ms inference time) 
and achieves significant absolute performance improvements of $0.74\%$ and $0.64\%$ on VQAv2 and Flickr30K, respectively.
We further analyse other well-performed variants of \mtname{} in the $2^{\text{nd}}$ and $3^{\text{rd}}$ rows.
It is worth noting that the two variants share a similar computational budget as \btname{}, but achieve better performance.
This not only demonstrates the efficiency and effectiveness of our \mtname{} architecture, but also reminds us that the cross-modal fused query via the cross-attention mechanism is the main reason for the additional computational budget of \mtname{} ($4^{\text{th}}$ row), as it is the only difference between the $3^{\text{rd}}$ and $4^{\text{th}}$ row models.
This inspires us to explore a more efficient method to fuse $\mathbf{C}^\mathrm{V}_{\ell-1}$ and $\mathbf{C}^\mathrm{T}_{\ell-1}$ to get the cross-modal fused query in the future.

\subsection{Cross-Attention and Concat-Attention Managers}
\label{appendix:other-managers}

\subsubsection{Cross-Attention Managers}
We implement the standard cross-attention mechanism~\cite{vaswani2017attention} and reduce the linear projection layer for value to save computational budget.\footnote{The calculation of cross-modal fused query also uses this simplified version of the cross-attention mechanism.}
Take the visual manager for example, it takes $\mathbf{C}^\mathrm{V}_{\ell-1} \in \mathbb{R}^{\mathrm{L} \times \mathrm{D}}$ as the query, and the first token of multi-layer unimodal representations, \ie{} $\mathbf{V}[:, 0] \in \mathbb{R}^{\mathrm{N} \times \mathrm{D}}$, as the key. Hence, the shape of generated aggregation weights is $\mathrm{N} \times \mathrm{L}$, which can be broadcast to the aggregation weights $\mathbf{W}_\mathrm{A}\!\in\!\mathbb{R}^{\mathrm{N} \times \mathrm{L} \times \mathrm{D}}$. 
The following calculation is the same as AAUMs 
in~\Fref{fig:details}. 
The results 
in~\Tref{tab:type-and-query} 
show a significant decrease compared to other managers on Flickr30K. 
We attribute this to the fact that the cross-attention manager will break the locality (or spatial consistency) of unimodal representations~\cite{cha2024honeybee,tong2024cambrian}, which is crucial for the downstream tasks like image--text retrieval.
We leave the detailed analysis of this phenomenon to the future work.

\subsubsection{Concat-Attention Managers}
Take the visual manager as an example, it broadcasts $\mathbf{C}^\mathrm{V}_{\ell-1}\!\in\!\mathbb{R}^{\mathrm{L} \times \mathrm{D}}$ to $\mathbb{R}^{\mathrm{N} \times \mathrm{L} \times \mathrm{D}}$, and concatenates it with $\mathbf{V}\!\in\!\mathbb{R}^{\mathrm{N} \times \mathrm{L} \times \mathrm{D}}$ along the last dimension as the concatenated query.
It then directly projects the query to $\mathbf{W}_\mathrm{A}\!\in\!\mathbb{R}^{\mathrm{N} \times \mathrm{L} \times \mathrm{D}}$. 
The following calculation is the same as AAUMs 
in~\fref{fig:details}. 
In fact, this type of manager is different from all other managers from the perspectives of the generated aggregation weights.
Although its aggregation weights delve into the feature dimension of $\mathbf{C}^\mathrm{V}_{\ell-1}$ and $\mathbf{V}$, the substantially increased number of parameters and computational cost do not result in a significant performance gain, making it impractical and inefficient.
More efficient variants of this type of manager should be investigated in the future.\looseness=-1

\subsection{Detailed Comparison with Previous Arts}

Due to the space limitations, we omit some baselines and details 
in~\Tref{tab:main-results}. 
Here we provide more details on the comparison with previous arts in~\Tref{tab:detailed-main-results}.

\subsection{Implementation Details}
\label{appendix:implementation_details}

\subsubsection{Vision--Language Pre-training}
Following \metername{}~\cite{dou2021meter}, we use two common VLP objectives for \twotower{} VLMs.

\paragraph{Masked Language Modeling (MLM)}
For MLM, we follow the conditional masking approach used in UNITER~\cite{chen2020uniter} that randomly masks $15\%$ of the tokens in the text token sequence while keeping the image patch sequence unchanged.
The model is then trained to predict the original masked tokens given the incomplete text sequence and the complete image patch sequence.
The masking strategy and MLM task head we use are the same as RoBERTa. The top-layer representation of the textual part of the cross-modal encoder is used as input for the MLM task head.

\paragraph{Image--Text Matching (ITM)}
For ITM, both matched and mismatched image--text pairs are fed into the model with equal probability. 
The model is trained to predict whether a given image--text pair is a matched (positive) or a mismatched (negative) pair.
The top-layer representations of $\texttt{[class]}$ and $\texttt{[<s>]}$ tokens are activated by the non-linear function $\texttt{Tanh}$.
Then the concatenation of the above representations is fed into a linear classifier with cross-entropy loss for classification.

\begin{table}[t]
	\tablestyle{6pt}{1.1}
	\caption{
		Statistics of the pre-train datasets. We remove duplicate image--caption pairs in VG~\cite{kim2021vilt, dou2021meter} and only 2.9M image--caption pairs can be downloaded in CC.
	}
	\label{tab:statistics_pretrain}
	\adjustbox{width=0.8\linewidth}{
		\begin{tabular}{l|cccc}
			            & COCO & VG   & CC   & SBU  \\
			\shline
			\#~Images   & 113K & 108K & 2.9M & 860K \\
			\#~Captions & 567K & 4.8M & 2.9M & 860K \\
		\end{tabular}
	}
\end{table}

\subsubsection{Pre-training Settings}

\Tref{tab:statistics_pretrain} shows the statistics of the pre-train datasets.
Following previous work~\cite{kim2021vilt,chen2020uniter,li2021align,dou2021meter}, we adopt four public image--caption datasets for pre-training, including Conceptual Captions (CC)~\cite{sharma-etal-2018-conceptual}, SBU Captions (SBU)~\cite{NIPS2011_5dd9db5e}, MSCOCO Captions (COCO)~\cite{chen2015microsoft}, and Visual Genome (VG)~\cite{krishna2017visual}.
The total numbers of the unique images and image--caption pairs in the combined training data are $4$M and $9$M.
\Tref{tab:hyperparam_pretrain} describes the hyperparameters for pre-training the \mtname{}.
The learning rate of the cross-modal encoder is five times higher than the base learning rate~\cite{dou2021meter}, \eg{} used by unimodal encoders.

\subsubsection{Fine-Tuning Settings}
\paragraph{Dataset Setting}
Standard settings and splits are used for all datasets.
Noted that,
for Flickr30K~\cite{young-etal-2014-image}, we follow the standard Karpathy Split~\cite{karpathy2015deep}; 
for VQAv2~\cite{balanced_vqa_v2}, we follow the common practice~\cite{balanced_vqa_v2, teney2018tips}: convert VQAv2 to a classification task with $3,129$ answer classes; train the model with training data and validation data, and evaluate the model on the Test-Dev and Test-Std data.

\begin{figure*}[t]
	\centering
	\includegraphics[width=\textwidth]{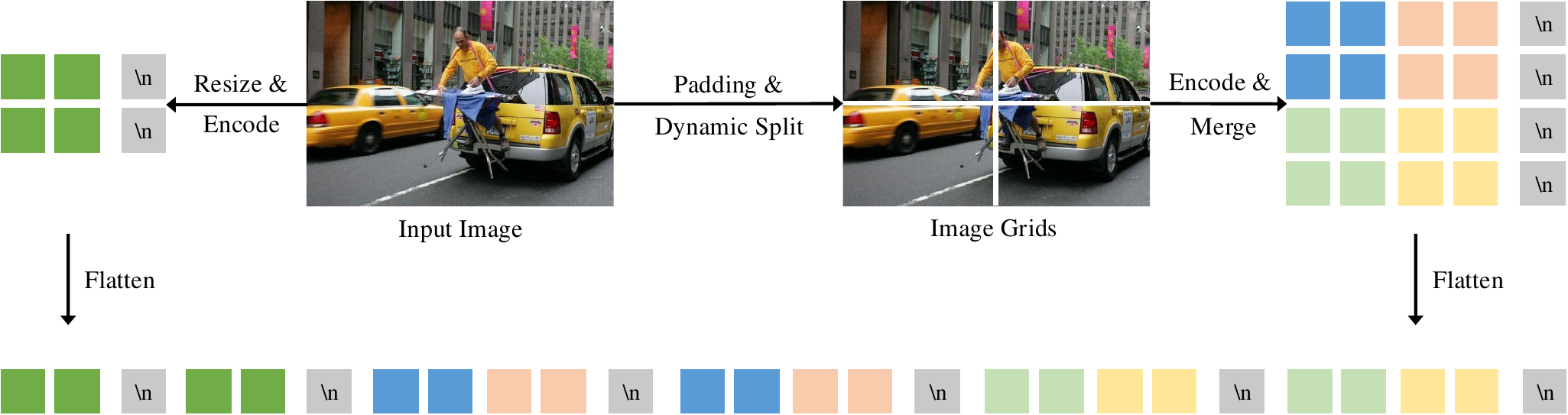}
	\caption{
		A brief illustrations of the multi-grid algorithm used in \ovname{}.
		The bilinear interpolation operation is omitted for simplicity, which is used to reduce \# input visual tokens.
		``$\backslash{}$n'' denotes the special token to indicate the end of a row and the shape of the input image. Illustration inspired by DeepSeek-VL2~\cite{wu2024deepseekvl2}.\looseness=-1
	}
	\label{fig:multi-grid}
\end{figure*}

\paragraph{Image Augmentation}
We follow previous works~\cite{li2021align,li2022blip} to use the combination of RandomResizedCrop, RandomHorizontalFlip, and RandAugment~\cite{cubuk2020randaugment} during training. 

\subsubsection{Fine-Tuning Strategies}
For visual question answering (VQAv2~\cite{balanced_vqa_v2}), visual entailment (SNLI-VE~\cite{xie2019visual}) and visual reasoning (NLVR$^2$~\cite{suhr2019corpus}), the fine-tuning strategy is similar to the strategy we used in the ITM objective. 
We pass the final representation of $\texttt{[class]}$ token and $\texttt{[<s>]}$ token to the non-linear layer activated by $\texttt{Tanh}$, and feed the concatenation of the output into a classifier. 

For image--text retrieval (Flickr30K~\cite{young-etal-2014-image}), we follow the approach used in ALBEF~\cite{li2021align} to optimize our model with both image--text contrastive (ITC) and ITM objectives.
In the training phase, we first add two linear projections on top of the unimodal encoders and calculate the contrastive similarity of unimodal representations of image--text pairs by dot product to compute the ITC loss.
Formerly, negative image--text pairs in ITM loss are sampled randomly.
However, after computing the ITC loss, we can use contrastive similarity distribution to sample one hard in-batch negative text (image) for each image (text) in a mini-batch.
In the inference phase, we first compute the contrastive similarity for all images and texts, and then select the top-k candidates based on their contrastive similarity. We then calculate their ITM scores for these candidates to determine the final ranking.

\Tref{tab:hyperparam_finetune} describes the hyperparameters for fine-tuning on $4$ downstream datasets.
Following previous work~\cite{kim2021vilt,dou2021meter}, we apply cross-entropy loss for SNLI-VE, NLVR$^2$ and Flickr30K and binary cross-entropy loss for VQAv2.

\section{Exploration on MLLM}

\subsection{A Brief Illustration of the Multi-Grid Algorithm}
\label{appendix:multi-grid}

The multi-grid algorithm~\cite{lin2023sphinx,li2024monkey,liu2024llavanext,shi2025we} is a widely used technique in MLLMs to help the visual encoder, \eg{} SigLIP, efficiently process images with varying image resolutions and aspect ratios.
We show a brief illustration of the multi-grid algorithm used in \ovname{} in~\fref{fig:multi-grid}, and recommend readers to refer to the original paper~\cite{li2024llava} for more details.

\begin{figure}[t]
	\centering
	\includegraphics[width=0.48\textwidth]{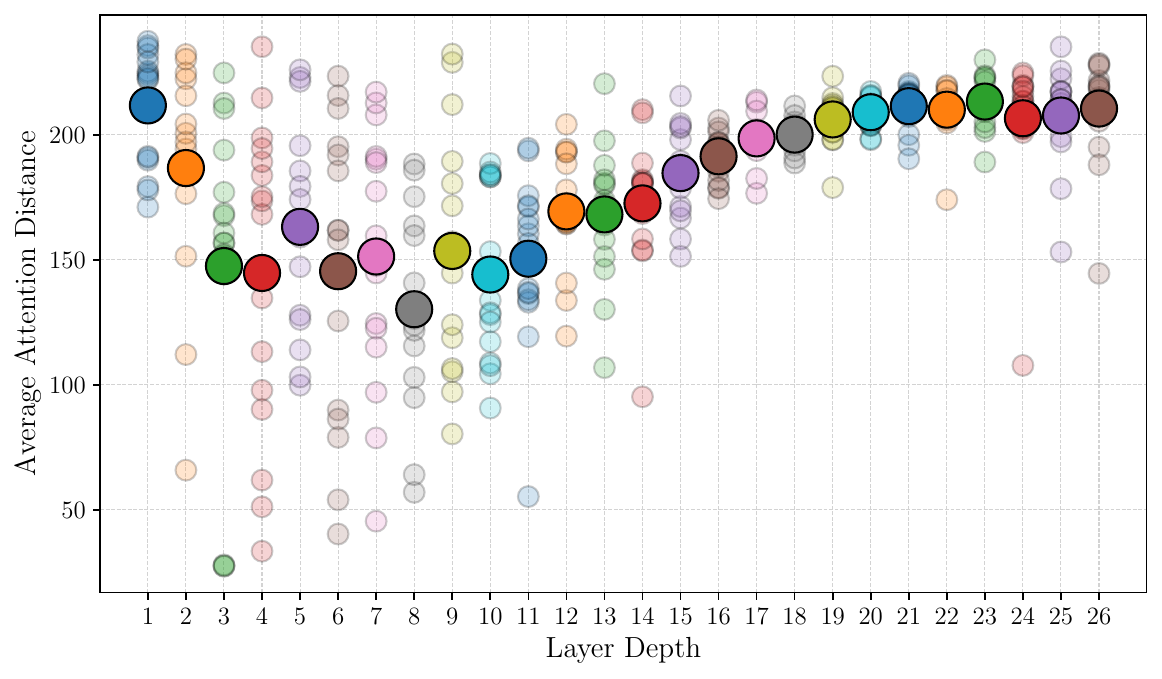}
	\caption{
		Size of attended area by head and layer depth. 
		Each small/large dot shows the mean attention distance of each head or all heads in each layer of the visual encoder.
		Image width is $384$ pixels and patch size is $14 \times 14$ pixels.\looseness=-1
	}
	\label{fig:siglip_attention_distance}
\end{figure}

\subsection{Average Attention Distance of SigLIP}
\label{appendix:attention_distance}
As stated in~\Sref{sec:visual-representation-selection},
similar to the observations in both \btname{} and \mtname{}, visual representations from the top half of the visual encoder bring the best performance, and using visual representations from all layers leads to the lowest performance improvement.

We visualize the average attention distance of SigLIP in~\fref{fig:siglip_attention_distance}.
It is computed by averaging the 2D euclidean distance between the query patch (token) and all other patches (tokens), weighted by the attention weight of visual self-attention.
Similar to the observations in ViT~\cite{dosovitskiy2020image}, in \ovname{},
the average attention distance vary significantly across heads in the bottom half of SigLIP, with some heads focusing on the most of the image while others focusing on the query location or a small region nearby.
As depth increases, especially in the top half of the visual encoder, the average attention distance of all heads increases, where most heads attend \textbf{widely} across tokens and capture global visual features.
Above observations provide a possible explanation for the performance difference between using visual representations from the top half and all layers of SigLIP in \ovname{}.

\subsection{Detailed Analysis on ScienceQA and OK-VQA}
\label{appendix:detailed-analysis-mllm}
As a supplement for~\Sref{sec:detailed-analysis-mllm}, 
we further provide a detailed analysis ScienceQA and OK-VQA, to intuitively analyse the effectiveness of managers.

\begin{figure}[t]
	\centering
	\includegraphics[width=0.48\textwidth]{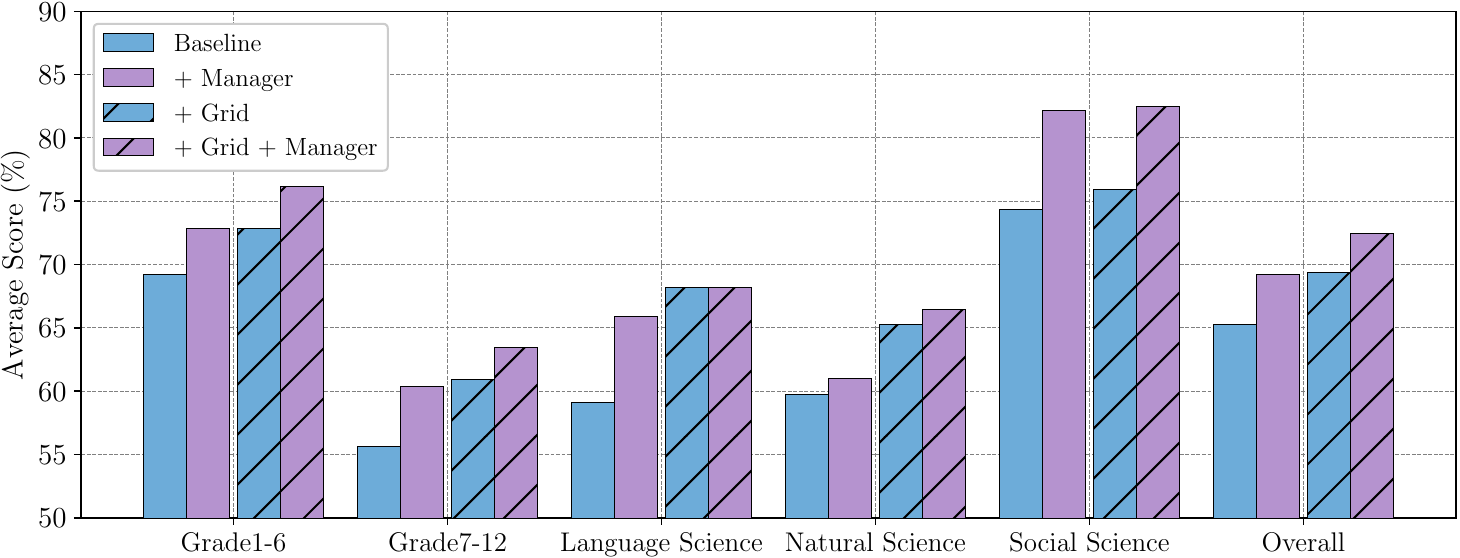}
	\caption{
        Zero-shot performance of four baselines on ScienceQA-IMG test set.
        The overall average score and the average score of each dimension are shown.\looseness=-1
	}
	\label{fig:category_scienceqa}
\end{figure}

\begin{figure}[t]
	\centering
	\includegraphics[width=0.45\textwidth]{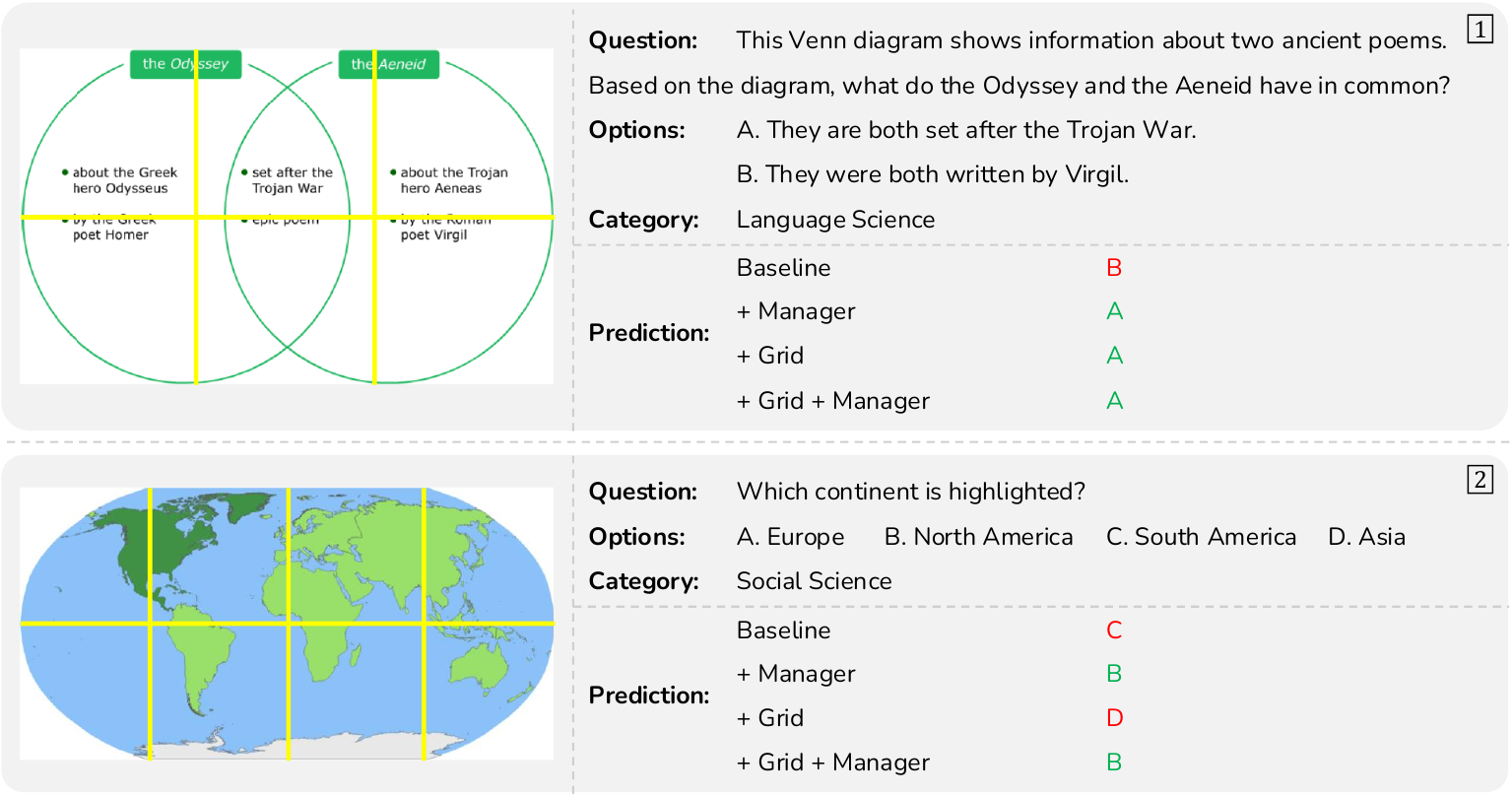}
	\caption{
        Case studies of four baselines on ScienceQA-IMG test set.
        Red and green fonts represent incorrect and correct predictions, respectively.
        Yellow lines indicate the boundaries of the image grids.
	}
	\label{fig:case_scienceqa}
\end{figure}

\subsubsection{ScienceQA}
This is a knowledge-intensive dataset with low-resolution hybrid images.
As shown in~\fref{fig:category_scienceqa}, both the manager and the multi-grid algorithm can bring performance improvements on different dimensions.
For ``Language Science'', the manager brings significant improvements to \basename{}, and such improvement seems to overlap with that provided by the multi-grid algorithm.
For ``Natural Science, Social Science'', the manager brings stable improvements. 
Especially in ``Social Science'', we can notice that the multi-grid algorithm brings \textbf{slight} improvements to \basename{}, while our manager can bring \textbf{significant} improvements to both \basename{} and \basegname{}.
Furthermore, their synergy can further improve performance.
Take the case {\scriptsize $\boxed{1}$} in ~\fref{fig:case_scienceqa} as an example, in ``Language Science'', most samples require analysis based on text information in the image or the question.
\basename{} misses visual details in the image, and both the manager and the multi-grid algorithm can help \basename{} capture visual details ``set after the Trojan War''.
As for the case {\scriptsize $\boxed{2}$}, the improvement of Manager mainly comes from map related problems.
The multi-grid algorithm divides the map into multiple grids, which may make it more difficult to understand, \eg{} the highlighted North America is divided into two grids, and then bring semantic ambiguity.
By introducing different levels of semantic knowledge, our manager can not only help Baseline to capture the highlighted part, but also mitigate semantic ambiguity caused by the multi-grid algorithm.\looseness=-1

\subsubsection{OK-VQA}
This is a general dataset with low-resolution natural images.
As shown in~\fref{fig:category_okvqa}, 
surprisingly, the multi-grid algorithm does not improve the performance much and even leads to \textbf{significant} degradation on many dimensions:
\begin{figure}[t]
	\centering
	\includegraphics[width=0.48\textwidth]{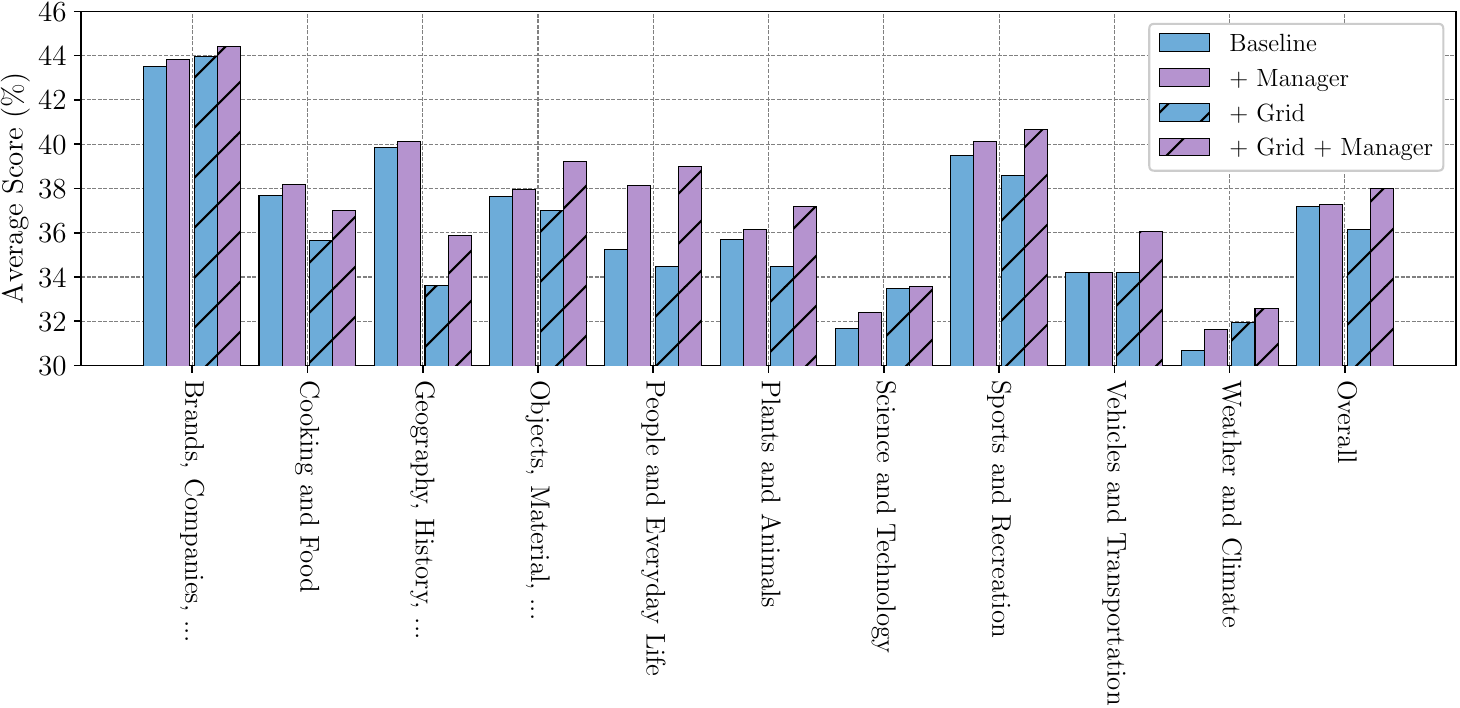}
	\caption{
        Zero-shot performance of four baselines on OK-VQA validation set.
	}
	\label{fig:category_okvqa}
\end{figure}

\begin{figure}[t]
	\centering
	\includegraphics[width=0.45\textwidth]{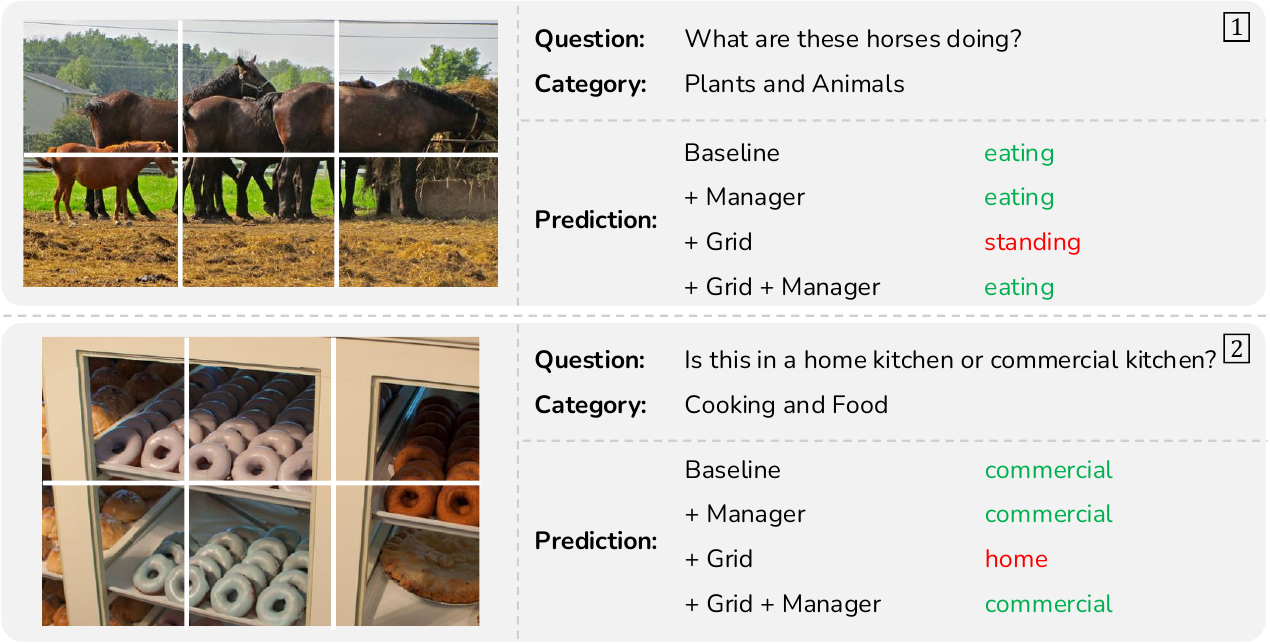}
	\caption{
        Case studies of four baselines on OK-VQA validation set.
        Red and green fonts represent incorrect and correct predictions, respectively.
        White lines indicate the boundaries of the image grids.
	}
	\label{fig:case_okvqa}
\end{figure}
\begin{itemize}
	\item Objects, Material, and Clothing; People and Everyday Life; Plants and Animals; Sports and Recreation: images about common objects, things, and scenes in daily life or in nature.
	\item Cooking and Food: close-ups of food or dining table.
	\item Geography, History, Language and Culture: 
	complex buildings (groups), event scenes containing different types of objects, \etc{}
\end{itemize}
Take cases in~\fref{fig:case_okvqa} as an example, the multi-grid algorithm may actually increase the understanding difficulty and bring semantic ambiguity.
Especially for (complex) scenes with many (different types of) objects and things, it cuts off objects and connected regions, which may hinder the perception of objects, things and scenes.
By aggregating insights from different levels of visual experts, our manager can not only improve \basename{}, but also partially compensate for the performance loss caused by the multi-grid algorithm, and may even further improve the performance on some dimensions.

\begin{figure}[t]
	\centering
	\includegraphics[width=0.48\textwidth]{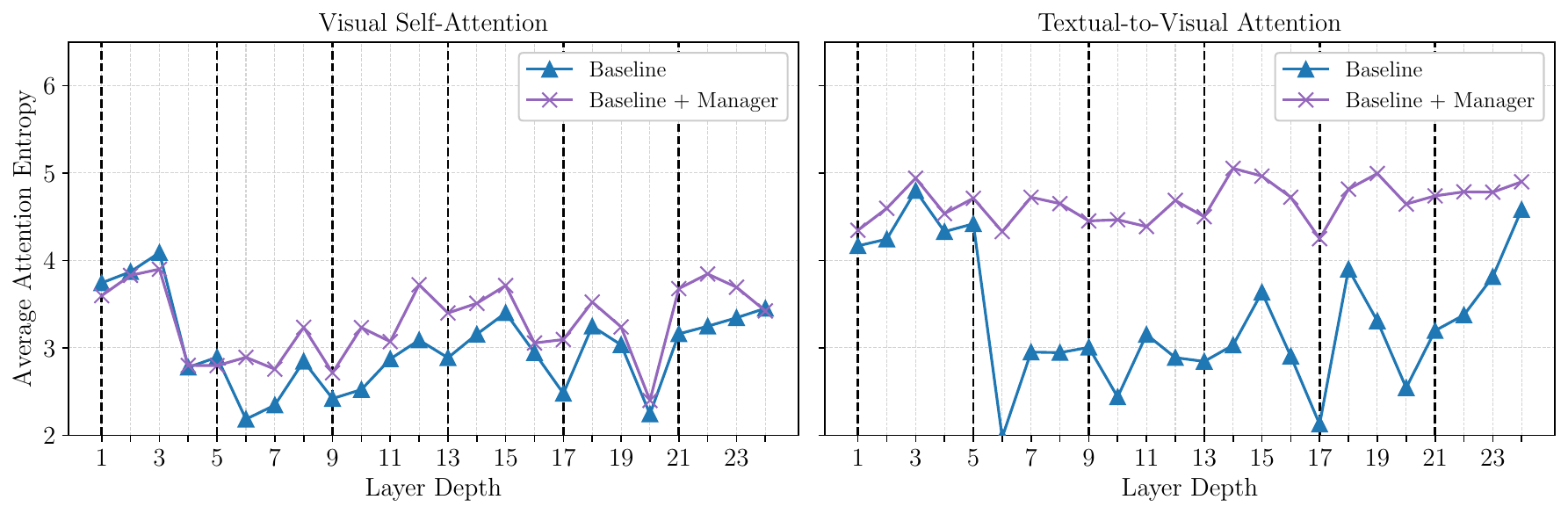}
	\caption{
        Average entropy of attention weight distributions in each layer of \basename{} and \basemname{}.
		The dotted vertical lines indicate the layers where managers are injected.
	}
	\label{fig:visualization_entropy_no_grid}
\end{figure}

\begin{figure}[t]
	\centering
	\includegraphics[width=0.48\textwidth]{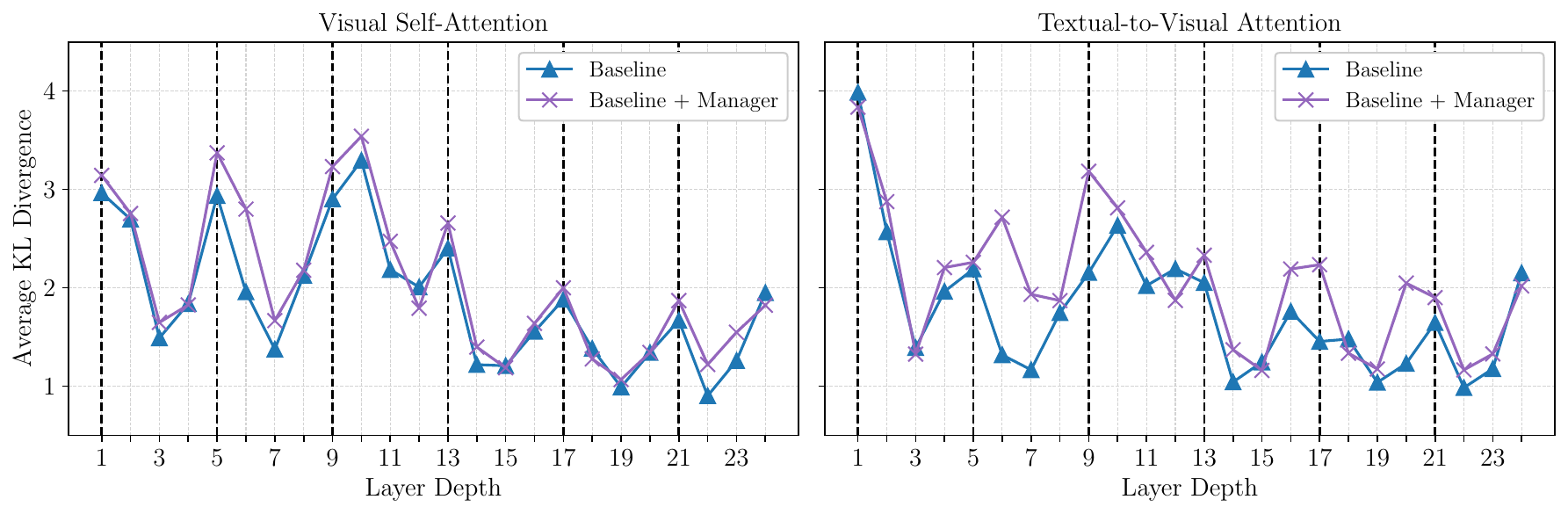}
	\caption{
        Average KL divergence between attention weight distributions of attention heads in each layer of \basename{} and \basemname{}.
		The dotted vertical lines indicate the layers where managers are injected.
	}
	\label{fig:visualization_kl_no_grid}
\end{figure}

\subsection{Attention Weight Distribution Analysis without Multi-Grid}
\label{appendix:attention-weight-distribution-analysis}
As a supplement for~\Sref{sec:attention-weight-distribution-analysis}
we further provide the average entropy and KL divergence of attention weight distributions in each layer of \basename{} and \basemname{}.
Similar trend can be observed in~\fref{fig:visualization_entropy_no_grid} and~\ref{fig:visualization_kl_no_grid}, 
benefiting from difference levels of semantic knowledge introduced by our manager, \basemname{} shows higher diversity of attention weight and attention heads in each layer. 
This helps the model capture richer and more diverse features, and further improve the performance on downstream tasks.

\subsection{Implementation Details}
\label{appendix:implementation_details_mllm}
In this paper, we take LLaVA-OneVision-0.5B-SI~\cite{li2024llava} as our baseline (\ovname{} for short), which is a widely used open-source multi-grid MLLM.
We follow the same training settings as the original \ovname{} and use about $8$M data samples for multi-stage training.
Different \twotower{} VLMs that trained with various pre-training objectives and fine-tuning strategies for different downstream tasks.
However, most of latest MLLMs are trained with the autoregressive objective, which teaches the model to maximize the likelihood of the next token given the previous tokens, including the input image, question and instruction tokens.
\ovname{} adopts a curriculum learning strategy that trains the model with $3$ different stages:
\begin{itemize}
	\item \textbf{Stage-1}: Language-Image Alignment. A small amount of data is used to initially align the visual representation from the visual encoder to the word embedding space of the LLM. $558$K image--caption pairs from LCS-558K~\cite{liu2024visual} are used in this stage, and the captions come from the Internet.
	\item \textbf{Stage-1.5}: High-Quality Knowledge Learning. About $4M$ data are used in this stage to learn help the model further align the visual encoder and the LLM, and also learn high-quality knowledge. 
	The data consists of re-captioned detailed description data and document/ocr data, and most of the data is synthetic.
	\item \textbf{Stage-2}: Visual Instruction Tuning. 
	About $3.2$M data are used in this stage to teach the model to solve diverse downstream tasks under the zero-shot setting. 
	The data consists of a wide range of multimodal datasets.
\end{itemize}
We recommend readers to refer to the original paper~\cite{li2024llava} for more details about the training settings and data.
The only two differences between our \basemname{} and the original \ovname{} are:
($i$) \textbf{Data}: partial training data are not open sourced;
($ii$) \textbf{Maximum length}: we use $16,384$ instead of $32,768$ as the maximum length of each input sample for better efficiency (more than $99$\% of the samples have a length less than $16,384$).
The above two differences bring slight performance differences between \basemname{} and the original \ovname{}, and we remove few some downstream datasets for better efficiency and robustness.

\begin{table}[t]
	\tablestyle{6pt}{1.1}
	\caption{
		Hyperparameters for pre-training. The first block is the hyperparameters for the cross-modal encoder.
	}
	\label{tab:hyperparam_pretrain}
	\adjustbox{width=0.8\linewidth}{
		\begin{tabular}{l|cc}
			Hyperparameters            & \mtname{}               \\
			\shline
			Number of Layers           & $6$                     \\
			Hidden size                & $768$                   \\
			FFN inner hidden size      & $3,072$                 \\
			Number of Attention heads  & $12$                    \\
			Dropout Ratio              & $0.1$                   \\
			Attention dropout          & $0.1$                   \\
			\hline
			Total Steps                & $100$k                  \\
			Batch Size                 & $4,096$                 \\
			Optimizer                  & $\operatorname{AdamW}$  \\
			Learning Rate              & $1e^{-5}$               \\
			Learning Rate Decay        & $\operatorname{Linear}$ \\
			Weight Decay               & $0.01$                  \\
			Warmup Steps               & $10$k                   \\
			Adam $\epsilon$            & $1e^{-8}$               \\
			Adam $\beta_1$             & $0.9$                   \\
			Adam $\beta_2$             & $0.98$                  \\
			\hline
			Center-Crop                & \cmark                  \\
			Random Resized Crop        & \xmark                  \\
			Random Augmentation        & \xmark                  \\
			Random Horizontal Flipping & \xmark                  \\
			\hline
			Textual Encoder            & RoBERTa\modelbase{}     \\
			Visual Encoder             & CLIP-ViT B-224/16              \\
			Patch Size                 & $16$                    \\
			Image Resolution for VLP   & $288$                   \\
		\end{tabular}
	}
\end{table}

\subsection{Evaluation Details}
\label{appendix:evaluation_details_mllm}

\begin{table*}[!h]
	\tablestyle{5pt}{1.1}
	\caption{
		Details of $20$ downstream datasets used in our exploration on MLLMs.
        For image resolution and aspect ratio, we provide the minimum, maximum, and average values in the form of $x / y / z$.
        The threshold for low and high resolution is $384 \times 2 = 768$ pixels.
        For MM-LiveBench, we use two different subsets of it (2407 and 2409).  
        $\star$ indicates the datasets are one of the $9$ datasets we used in the ablation study for efficiency and robustness.
	}
	\label{tab:evaluation_mllm}
	\adjustbox{width=\linewidth}{
        \begin{tabular}{c|l|l|l|c|c|rr|r}
        {Capability} & 
        \multirow{2}{*}{Dataset} &
        \multirow{2}{*}{Description} &
        \multirow{2}{*}{Answer Category} &
        \multicolumn{1}{c|}{{Image}} &
        \multicolumn{1}{c|}{{Resolution}} &
        \multicolumn{1}{c}{\multirow{2}{*}{Resolution}} &
        \multicolumn{1}{c|}{\multirow{2}{*}{Aspect Ratio}} &
        \multirow{2}{*}{\# Samples} \\
        {Category} & & & & \multicolumn{1}{c|}{{Category}} & {Category}  & & & \\
        \shline
        \multirow{6}{*}{General} &
        VQAv2~\cite{balanced_vqa_v2} $\star$ &
        Scene understanding VQA &
        Open form (Short) &
        Natural &
        Low &
        120 / \ \,640 / \ \,523 &
        0.17 / 1.00 / 0.72 &
        214.00K \\
        &
        OKVQA~\cite{okvqa} &
        External knowledge VQA &
        Open form (Short) &
        Natural &
        Low &
        208 / \ \,640 / \ \,524 &
        0.26 / 1.00 / 0.72 &
        5.05K \\
        &
        GQA~\cite{hudson2019gqa} $\star$ &
        Compositional VQA &
        Open form (Short) &
        Natural &
        Low &
        314 / \ \,640 / \ \,523 &
        0.36 / 1.00 / 0.71 &
        12.58K \\
        &
        MMVet~\cite{yu2024mmvet} &
        Multi-discipline &
        Open form (Long) &
        Hybrid &
        High &
        181 / 4180 / \ \,909 &
        0.27 / 1.00 / 0.74 &
        0.22K \\
        &
        SEED-Bench~\cite{li2023seed} $\star$ &
        Multi-discipline; Large-scale &
        Multi-choice &
        Natural &
        High &
        240 / 9906 / \ \,791 &
        0.50 / 1.00 / 0.66 &
        18.00K \\
        &
        RealWorldQA~\cite{xai2024realworldqa} $\star$ &
        Real-world VQA &
        Multi-choice; Open form (Short) &
        Natural &
        High &
        512 / 1469 / 1147 &
        0.45 / 1.00 / 0.67 &
        0.77K \\
        \hline
        \multirow{5}{*}{Text-Rich} &
        TextVQA~\cite{singh2019towards} $\star$ &
        Scene Text Recognition &
        Open form (Short) &
        Natural &
        High &
        512 / 3991 / \ \,872 &
        0.25 / 1.00 / 0.73 &
        5.00K \\
        &
        ChartQA~\cite{masry2022chartqa} &
        Chart Understanding &
        Open form (Short) &
        Abstract &
        Low &
        245 / 1199 / \ \,665 &
        0.39 / 0.99 / 0.71 &
        2.50K \\
        &
        DocVQA~\cite{mathew2021docvqa} $\star$ &
        Document Understanding &
        Extractive &
        Abstract &
        High &
        418 / 6209 / 1918 &
        0.33 / 0.99 / 0.75 &
        5.19K \\
        &
        InfoVQA~\cite{mathew2022infographicvqa} &
        Infographic Understanding &
        Extractive; Numerical &
        Abstract &
        High &
        452 / 7655 / 1724 &
        0.09 / 1.00 / 0.46 &
        3.29K \\
        &
        OCRBench~\cite{liu2024ocrbenchhiddenmysteryocr} &
        Text Recognition &
        Open form (Short) &
        Hybrid &
        Low &
        20 / 5900 / \ \,627 &
        0.08 / 1.00 / 0.57 &
        1.00K \\
        \hline
        \multirow{4}{*}{Knowledge} &
        AI2D~\cite{kembhavi2016diagram} $\star$ &
        Science Diagrams &
        Multi-choice &
        Abstract &
        Low &
        152 / 1500 / \ \,547 &
        0.18 / 1.00 / 0.72 &
        3.09K \\
        &
        ScienceQA~\cite{lu2022learn} $\star$ &
        High-school Science &
        Multi-choice &
        Hybrid &
        Low &
        114 / \ \,685 / \ \,378 &
        0.10 / 1.00 / 0.66 &
        2.02K \\
        &
        MMMU~\cite{yue2023mmmu} $\star$ &
        College-level Multi-discipline &
        Multi-choice; Open form (Short) &
        Hybrid &
        Low &
        185 / 2217 / \ \,580 &
        0.11 / 1.00 / 0.60 &
        0.90K \\
        &
        MathVista~\cite{lu2024mathvista} &
        General Math Understanding &
        Multi-choice; Open form (Short) &
        Hybrid &
        Low &
        58 / 4275 / \ \,499 &
        0.10 / 1.00 / 0.72 &
        1.00K \\
        \hline
        \multirow{4}{*}{Real-World} &
        ImageDC~\cite{li2024llavanext-ablations} &
        Image Detail Description &
        Open form (Long) &
        Hybrid &
        Low &
        470 / \ \,512 / \ \,499 &
        0.60 / 1.00 / 0.76 &
        0.10K \\
        &
        MM-LiveBench~\cite{zhang2024lmms} &
        Internet Content Understanding &
        Open form (Long) &
        Hybrid &
        High &
        980 / 2051 / 1788 &
        0.51 / 0.92 / 0.61 &
        0.25K \\
        &
        LLaVA-Wild~\cite{liu2024visual} &
        Real-world Chat &
        Open form (Long) &
        Hybrid &
        High &
        465 / 3921 / 1277 &
        0.51 / 1.00 / 0.80 &
        0.06K \\
        &
        LLaVA-Wilder~\cite{liu2024improved} &
        Real-world Chat &
        Open form (Long) &
        Hybrid &
        High &
        224 / 3492 / \ \,870 &
        0.37 / 1.00 / 0.72 &
        0.21K
        \end{tabular}
    }
\end{table*}

As stated in~\Sref{sec:evaluation-mllm},
we follow the same evaluation settings as the original \ovname{}~\cite{li2024llava} for the zero-shot evaluation on $20$ downstream datasets.
We further divide these datasets into different groups from three perspectives, \ie{} category of capabilities~\cite{tong2024cambrian,zhang2024mm1}, images, and resolutions, to better analyse the effectiveness of managers in MLLMs and multi-grid MLLMs.
Details about the evaluation datasets we used are shown in~\Tref{tab:evaluation_mllm}.

\subsection{Detailed Results}
\label{appendix:detailed-results-mllm}

\begin{table*}[!h]
	\tablestyle{3pt}{1.2}
	\caption{
        Zero-shot performance of four baselines on $20$ datasets.
        The overall average score and the average score of each capability category are shown.
	}
	\label{tab:category_results_mllm}
	\adjustbox{width=\linewidth}{
        \begin{tabular}{l|c|cccc|ccc|cc}
            \multirow{2}{*}{Model} & {Overall} & \multicolumn{4}{c|}{Capability Category} & \multicolumn{3}{c|}{Image Category} & \multicolumn{2}{c}{Resolution Category} \\
            &    (\%)    & General (\%) & Text-Rich (\%) & Knowledge (\%) & Real-World (\%) & Natural (\%) & Abstract (\%) & Hybrid (\%) & Low (\%)   & High (\%)  \\
            \shline
            Baseline         & 50.54 & 48.72   & 47.52     & 46.05     & 57.26      & 56.49   & 46.50    & 48.60  & 54.34 & 46.74 \\
            \hdashline
            + Manager        & 51.50 & 49.59   & 48.53     & 47.63     & 57.74      & 57.44   & 47.48    & 49.53  & 55.31 & 47.68 \\
            + Grid           & 53.87 & 49.07   & 57.26     & 47.50     & 58.96      & 57.92   & 55.06    & 50.98  & 56.25 & 51.50 \\
            + Grid + Manager & \textbf{55.21} & \textbf{50.35} & \textbf{59.22} & \textbf{48.65} & \textbf{59.69} & \textbf{59.22} & \textbf{56.63} & \textbf{52.24} & \textbf{57.48} & \textbf{52.95}
        \end{tabular}
    }
\end{table*}
\begin{table*}[!h]
	\tablestyle{3pt}{1.2}
	\caption{
        Zero-shot performance of four baselines on $11$ datasets of ``General'' and ``Text-Rich'' capability categories.
        The score of each dataset and the average score of each capability category are shown.
        The score of OCRBench will be normalized by dividing by $10$ for the average calculation.
	}
	\label{tab:general_and_text_rich_results_mllm}
	\adjustbox{width=\linewidth}{
        \begin{tabular}{l|ccccccc|cccccc}
            \multirow{3}{*}{Model} & \multicolumn{7}{c|}{General} & \multicolumn{6}{c}{Text-Rich} \\
            \cline{2-14}
            & VQAv2 & OKVQA & GQA & MMVet & SEED-Bench & RealWorldQA & {Average} & TextVQA & ChartQA & DocVQA & InfoVQA & OCRBench & {Average} \\
            & val (\%) & val (\%) & testdev (\%) & test (\%) & image (\%) & test (\%) &  (\%) & val (\%) & test (\%) & test (\%) & test (\%) & test (\%) &  (\%) \\
            \shline
           Baseline & 73.90 & 37.18 & 57.66 & 21.40 & 61.50 & 50.98 & 48.71 & 57.56 & 54.32 & 51.40 & 27.30 & 470.00 & 47.52 \\
           \hdashline
           + Manager & 74.18 & 37.28 & 57.89 & 22.40 & \textbf{65.33} & 51.11 & 49.61 & 59.00 & 56.24 & 52.49 & 27.31 & 476.00 & 48.53 \\
           + Grid & 74.65 & 36.13 & 57.95 & 24.00 & 61.08 & 52.55 & 49.07 & 65.14 & 59.32 & 68.55 & 38.61 & 547.00 & 57.26 \\
           + Grid + Manager & \textbf{74.90} & \textbf{37.99} & \textbf{58.66} & \textbf{25.60} & {65.14} & \textbf{52.81} & \textbf{50.35} & \textbf{65.82} & \textbf{60.96} & \textbf{70.90} & \textbf{40.70} & \textbf{577.00} & \textbf{59.22}
        \end{tabular}
    }
\end{table*}
\begin{table*}[!h]
	\tablestyle{3pt}{1.2}
	\caption{
        Zero-shot performance of four baselines on $9$ datasets of ``Knowledge'' and ``Real-World'' capability categories.
        The score of each dataset and the average score of each capability category are shown.
	}
	\label{tab:knowledge_and_realwolrd_results_mllm}
	\adjustbox{width=\linewidth}{
        \begin{tabular}{l|ccccc|cccccc}
            \multirow{3}{*}{Model} & \multicolumn{5}{c|}{Knowledge} & \multicolumn{6}{c}{Real-World} \\
            \cline{2-12}
             & AI2D & ScienceQA & MMMU & MathVista & {Average} & ImageDC & \multicolumn{2}{c}{MM-LiveBench} & LLaVA-Wild & LLaVA-Wilder & {Average} \\
             & test (\%) & test (\%) & val (\%) & testmini (\%) &  (\%) & test (\%) & July (\%) & Sep (\%)  & test (\%) & test (\%) & (\%) \\
             \shline
            Baseline & 52.95 & 65.25 & 31.78 & 34.20 & 46.05 & 89.18 & 38.76 & 41.50 & 68.40 & 50.00 & 57.57 \\
            \hdashline
            + Manager & 53.89 & 69.21 & \textbf{33.11} & 34.30 & 47.63 & 89.44 & 39.74 & 43.27 & 69.00 & 50.70 & 58.43 \\
            + Grid & 53.76 & 69.36 & 30.78 & 36.10 & 47.50 & 89.73 & 41.28 & 43.80 & 68.80 & 51.20 & 58.96 \\
            + Grid + Manager & \textbf{53.95} & \textbf{72.43} & {31.22} & \textbf{37.00} & \textbf{48.65} & \textbf{89.95} & \textbf{41.89} & \textbf{45.13} & \textbf{69.30} & \textbf{52.20} & \textbf{59.69}
        \end{tabular}
    }
\end{table*}

We provide results of each category for our four baselines in~\Tref{tab:category_results_mllm}, across different categories of capabilities, images, and resolutions
Furthermore, we also provide detailed results of our four baselines on each dataset in~\Tref{tab:general_and_text_rich_results_mllm}, and~\ref{tab:knowledge_and_realwolrd_results_mllm}.

\begin{table*}[t]
	\tablestyle{5pt}{1.1}
	\caption{
		Hyperparameters for fine-tuning \mtname{} on $4$ downstream datasets. FT denotes fine-tuning. CE and BCE are short for cross-entropy loss and binary cross-entropy loss, respectively.
	}
	\label{tab:hyperparam_finetune}
	\adjustbox{width=\linewidth}{
		\begin{tabular}{l|cccc}
			Hyperparameters            & VQAv2                   & SNLI-VE                 & NLVR$^2$                & Flickr30K               \\
			\shline
			Total Epochs               & $10$                    & $4$                     & $5$                     & $20$                    \\
			Batch Size                 & $576$                   & $64$                    & $256$                   & $512$                   \\
			Optimizer                  & $\operatorname{AdamW}$  & $\operatorname{AdamW}$  & $\operatorname{AdamW}$  & $\operatorname{AdamW}$  \\
			Learning Rate              & $9e^{-6}$               & $3e^{-6}$               & $1.4e^{-5}$             & $6e^{-6}$               \\
			Learning Rate Decay        & $\operatorname{Linear}$ & $\operatorname{Linear}$ & $\operatorname{Linear}$ & $\operatorname{Linear}$ \\
			Weight Decay               & $0.06$                  & $0.01$                  & $0.01$                  & $0.01$                  \\
			Warmup Ratio               & $0.06$                  & $0.06$                  & $0.1$                   & $0.1$                   \\
			Adam $\epsilon$            & $1e^{-8}$               & $1e^{-8}$               & $1e^{-8}$               & $1e^{-8}$               \\
			Adam $\beta_1$             & $0.9$                   & $0.9$                   & $0.9$                   & $0.9$                   \\
			Adam $\beta_2$             & $0.98$                  & $0.98$                  & $0.98$                  & $0.98$                  \\
			\hline
			Center-Crop                & \xmark                  & \xmark                  & \xmark                  & \xmark                  \\
			Random Resized Crop        & \cmark                  & \cmark                  & \cmark                  & \cmark                  \\
			Random Augmentation        & \cmark                  & \cmark                  & \cmark                  & \cmark                  \\
			Random Horizontal Flipping & \xmark                  & \cmark                  & \cmark                  & \cmark                  \\
			\hline
			Textual Encoder            & RoBERTa\modelbase{}     & RoBERTa\modelbase{}     & RoBERTa\modelbase{}     & RoBERTa\modelbase{}     \\
			Visual Encoder             & CLIP-ViT B-224/16              & CLIP-ViT B-224/16              & CLIP-ViT B-224/16              & CLIP-ViT B-224/16              \\
			Patch Size                 & $16$                    & $16$                    & $16$                    & $16$                    \\
			Image Resolution for FT    & $576$                   & $384$                   & $384$                   & $384$                   \\
			Loss Function              & BCE                     & CE                      & CE                      & CE                      \\
		\end{tabular}
	}
\end{table*}

\clearpage

\clearpage

\bibliography{reference_short}
\bibliographystyle{IEEEtran}

\begin{IEEEbiography}[{\includegraphics[width=1in,height=1.25in,clip,keepaspectratio]{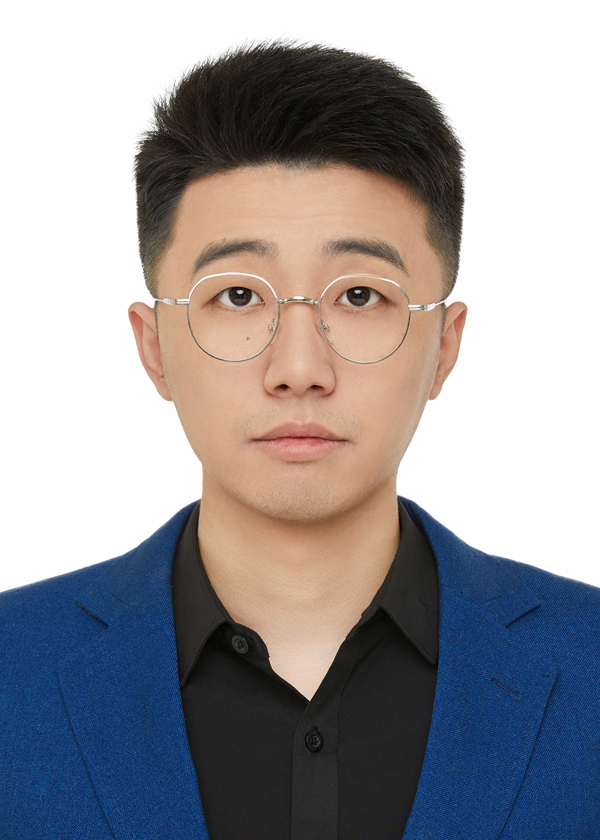}}]{Xiao Xu}
    received the B.S. degree from Northeastern University, Shenyang, China, in 2020. He is currently working toward the Ph.D. degree with the Harbin Institute of Technology, Harbin, China. He has published research papers at international NLP/AI conferences and journals, such as ACL, EMNLP, AAAI, and TASLP. His research interests include natural language processing, vision--language learning and multimodal large language models.
\end{IEEEbiography}

\begin{IEEEbiography}[{\includegraphics[width=1in,height=1.25in,clip,keepaspectratio]{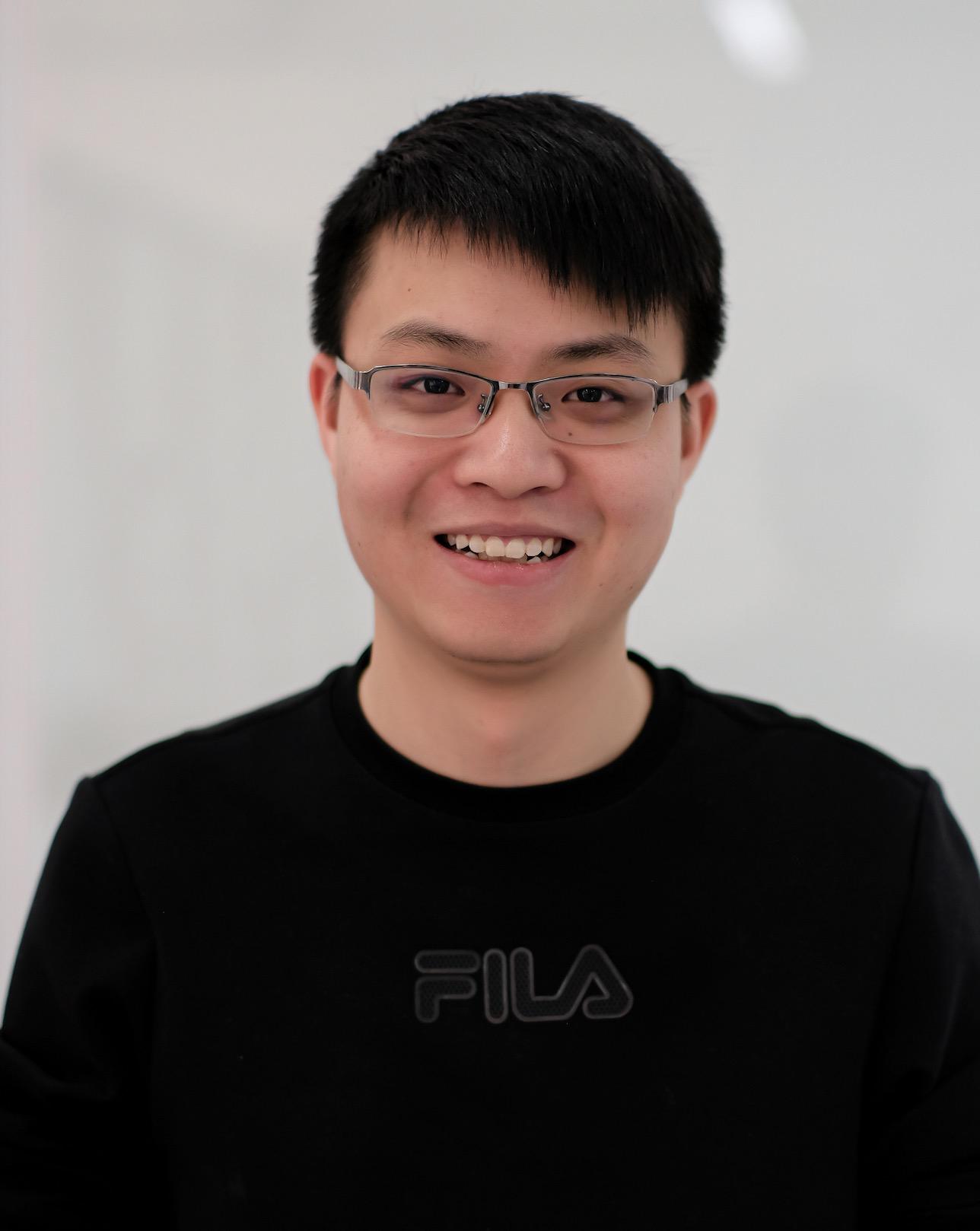}}]{Libo Qin}
    is a professor of School of Computer Science and Engineering, Central South University. He has published research papers at international NLP/AI conferences and journals, such as ACL, EMNLP, AAAI, and TASLP. His work has been selected as the Most Influential Paper by Paper Digest and won the Best Paper Award at the EMNLP2022 MMNLU Workshop. He has served as an Area Chair for EMNLP, NAACL, an Action Editor for ARR, and a Senior Program Committee Member for IJCAI. His research interests include natural language processing and large language models. 
\end{IEEEbiography}

\begin{IEEEbiography}[{\includegraphics[width=1in,height=1.25in,clip,keepaspectratio]{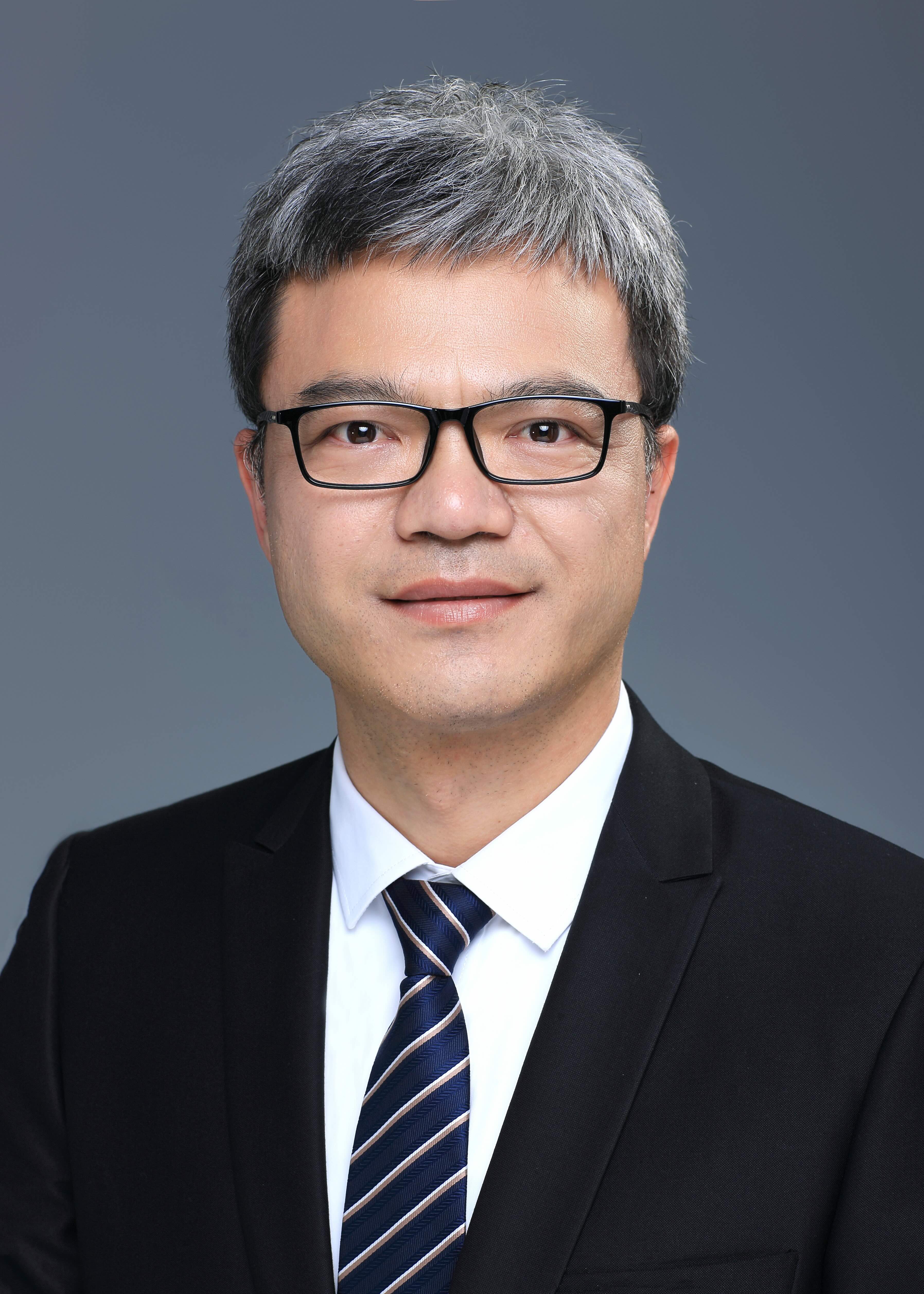}}]{Wanxiang Che}
    is a professor of School of Computer Science and Technology, Harbin Institute of Technology. He is the vice director of Research Center for Social Computing and Information Retrieval. He is a young scholar of ``Heilongjiang Scholar'' and a visiting scholar of Stanford University. He is currently the vice director and secretary-general of the Computational Linguistics Professional Committee of the Chinese Information Society of China; Officer and Secretary of AACL Executive Board; a senior member of the China Computer Federation (CCF). He received the AAAI 2013 Outstanding Paper Honorable Mention Award. His research interests include natural language processing and large language models.
\end{IEEEbiography}

\begin{IEEEbiography}[{\includegraphics[width=1in,height=1.25in,clip,keepaspectratio]{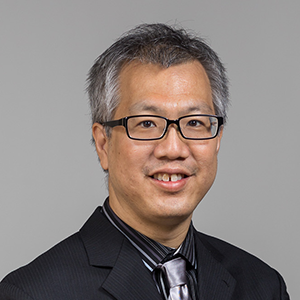}}]{Min-Yen Kan}
    is an Associate Professor and Vice Dean of Undergraduate Studies at the National University of Singapore. Min is an active member of the Association of Computational Linguistics (ACL), currently serving as a co-chair for the ACL Ethics Committee, and previously as the ACL Anthology Director (2008–2018). He is an associate editor for Information Retrieval and the survey editor for the Journal of AI Research (JAIR). His research interests include digital libraries, natural language processing and information retrieval. He was recognized as a distinguished speaker by the ACM for natural language processing and digital libraries research. Specific projects include work in the areas of scientific discourse analysis, fact verification, full-text literature mining, lexical semantics and large language models. He is a senior member of the IEEE.
\end{IEEEbiography}

\vfill

\end{document}